\documentclass{article}

\usepackage{arxiv}

\usepackage[utf8]{inputenc} % allow utf-8 input
\usepackage[T1]{fontenc}    % use 8-bit T1 fonts
\usepackage{hyperref}       % hyperlinks
\usepackage{url}            % simple URL typesetting
\usepackage{booktabs}       % professional-quality tables
\usepackage{amsmath, amsfonts, amsthm, amssymb, mathtools, xparse, stackengine}
\usepackage{nicefrac}       % compact symbols for 1/2, etc.
\usepackage{microtype}      % microtypography
\usepackage{lipsum}		% Can be removed after putting your text content
\usepackage{graphicx}
\usepackage{natbib}
\usepackage{doi}
\usepackage{mathtools, stackengine, threeparttable}
\usepackage{xcolor}
\usepackage{subfigure}
\usepackage{algorithm,algorithmic}
\usepackage{appendix}
\usepackage{footmisc}

\newtheorem{theorem}{Theorem}%  meant for continuous numbers
\newtheorem{remark}{Remark}%
\newtheorem{lemma}{Lemma}%
\newtheorem{corollary}{Corollary}

\definecolor{custom_blue}{HTML}{1F77B4}
\definecolor{custom_pink}{HTML}{E377C2}
\definecolor{custom_orange}{HTML}{FF7F0E}
\definecolor{custom_purple}{HTML}{9467BD}
\definecolor{custom_green}{HTML}{2CA02C}
\definecolor{custom_red}{HTML}{D62728}
\definecolor{custom_brown}{HTML}{8C564B}

\usepackage{tikz}
\usepackage{pgfplots}
\pgfplotsset{compat=1.17}

\newcommand{\blue}{\raisebox{2pt}{\tikz{\draw[custom_blue, solid, line width=2.3pt](0,0) -- (5mm,0);}}}

\newcommand{\orange}{\raisebox{2pt}{\tikz{\draw[custom_orange, solid, line width=2.3pt](0,0) -- (5mm,0);}}}
\newcommand{\purple}{\raisebox{2pt}{\tikz{\draw[custom_purple, solid, line width=2.3pt](0,0) -- (5mm,0);}}}
\newcommand{\green}{\raisebox{2pt}{\tikz{\draw[custom_green, solid, line width=2.3pt](0,0) -- (5mm,0);}}}
\newcommand{\red}{\raisebox{2pt}{\tikz{\draw[custom_red, solid, line width=2.3pt](0,0) -- (5mm,0);}}}
\newcommand{\brown}{\raisebox{2pt}{\tikz{\draw[custom_brown, solid, line width=2.3pt](0,0) -- (5mm,0);}}}

\newcommand*\rot{\rotatebox{90}}
\newcommand{\orcid}[1]{\href{https://orcid.org/#1}{\textcolor[HTML]{A6CE39}{\aiOrcid}}}

\title{Deep Intrinsically Motivated Exploration in Continuous Control}

\date{} 					% Or removing it

\author{Baturay Saglam, Suleyman S.~Kozat \\
	Department of Electrical and Electronics Engineering\\
	Bilkent University\\
	06800 Bilkent, Ankara, Turkey \\
	\texttt{\{baturay,kozat\}@ee.bilkent.edu.tr} \\
}

\renewcommand{\headeright}{}
\renewcommand{\undertitle}{}
\renewcommand{\shorttitle}{Deep Intrinsically Motivated Exploration in Continuous Control}

\hypersetup{
pdftitle={Deep Intrinsically Motivated Exploration in Continuous Control},
pdfsubject={arxiv},
pdfauthor={Baturay Saglam, Suleyman S.~Kozat},
pdfkeywords={deep reinforcement learning, exploration, intrinsic motivation, actor-critic algorithms, continuous control},
}

\begin{document}
\maketitle

\begin{abstract}
	In continuous control, exploration is often performed through undirected strategies in which parameters of the networks or selected actions are perturbed by random noise. Although the deep setting of undirected exploration has been shown to improve the performance of on-policy methods, they introduce an excessive computational complexity and are known to fail in the off-policy setting. The intrinsically motivated exploration is an effective alternative to the undirected strategies, but they are usually studied for discrete action domains. In this paper, we investigate how intrinsic motivation can effectively be combined with deep reinforcement learning in the control of continuous systems to obtain a directed exploratory behavior. We adapt the existing theories on animal motivational systems into the reinforcement learning paradigm and introduce a novel and scalable directed exploration strategy. The introduced approach, motivated by the maximization of the value function's error, can benefit from a collected set of experiences by extracting useful information and unify the intrinsic exploration motivations in the literature under a single exploration objective. An extensive set of empirical studies demonstrate that our framework extends to larger and more diverse state spaces, dramatically improves the baselines, and outperforms the undirected strategies significantly.
\end{abstract}

% keywords can be removed

\section{Introduction}
The balance between exploration and exploitation is a protracted and key challenge in modern deep reinforcement learning (RL) \citep{kearns_and_singh, sml_1}. The main aim of the exploration is to ensure that agents collect a set of diverse experiences not to adopt a premature behavior \citep{sml_3}. If the exploration is insufficient, action decisions that yield high rewards can be missed, and the policy of an agent can converge to a local optimum \citep{sml_3}. In contrast, with excessive exploration, agents may spend too much time and waste resources in trying many suboptimal actions without efficiently using the collected experiences \citep{sml_3}. Therefore, an optimal balance for the exploration and exploitation trade-off, also known as the \textit{exploration-exploitation dilemma}, must be well found through effective strategies \citep{thrun_1992, sml_4}. However, designing an effective and efficient exploration structure is not trivial as it cannot be inferred from the reward function of the underlying Markov Decision Process (MDP), and high-dimensional state and action spaces increase the required time and resources for the exploration \citep{thrun_1992}. 

Exploration in RL is usually studied in directed and undirected methods \citep{thrun_1992}. Undirected exploration techniques in high dimensional action spaces aim to perturb selected actions or weights of parameterized policies, where well-known examples are additive Gaussian noise \citep{williams_1992} or deep parameter space noise \citep{noisy_net, psne, deep_coherent_exploration}. Even though the deep strategies have advantages of being consistent, structured, and generalizable due to the exploration in parameter space conditioned on the states, they introduce excessive computational cost and heavily rely on randomness \citep{thrun_1992}. While undirected methods aim to inject randomness into the agents' action decisions, directed strategies rely on extracting useful information from the collected experiences in a rule-based manner \citep{thrun_1992}. Although these approaches require lower-level exploration to collect experiences and are shown to be effective, they are usually employed in hard exploration tasks where rewards are delayed, deceptive, or sparse \citep{deep_coherent_exploration}. 

As a counterpart to the directed exploration strategies in hard exploration tasks, intrinsic motivation has been effectively utilized from three aspects: prediction error, state novelty, and information gain \citep{barto_im_book}. The first motivation leads the agents towards the state spaces where the prediction is difficult. The second intuition is to add an intrinsic bonus when the agent goes into a state which it does not usually go, and the latter is an intrinsic reward depending on the reduction of uncertainty on the environment dynamics \citep{barto_im_book}. Although intrinsically motivated exploration methods are shown to perform well, they suffer from several limitations \citep{barto_im_book}. These limitations include the generalizability to different domains in that they only focus on low- or high-dimensional environments and are mostly studied for discrete action spaces. Furthermore, the usage of representation learning may prevent the long-term control from being included in the exploration \citep{barto_im_book}. Thus, exploration in continuous action spaces is restricted to undirected methods \citep{deep_coherent_exploration, barto_im_book}. However, they are known to suffer from off-policy learning due to the separation of the exploration from the control \citep{sml_11}. When the evaluation of the exploration strategy is not on par with the policy update, undirected methods may not be an informative exploratory guide to the agents \citep{sml_11}.

In this paper, we introduce a novel approach to the exploration-exploitation dilemma through the maximization of the temporal difference (TD) error that adapts the existing theories on animal motivational systems to the deep reinforcement learning for continuous control. We introduce a scalable deep directed high-frequency perturbation algorithm. This unified framework combines the motivations of prediction error, state novelty, and information gain by modeling an exploration policy that prevents agents from repeating actions through an additive learnable exploration direction. Our empirical studies on MuJoCo \citep{mujoco}, and Box2D \citep{box2d} continuous control tasks demonstrate that a learned internally motivated exploration policy outperforms the undirected exploration strategies in terms of stability, learning speed, and highest evaluation returns by expanding to useful state spaces. The main contributions of this paper are as follows:
\begin{itemize}
    \item We adapt the existing theories on animal motivational systems into the deep reinforcement learning paradigm to constitute an intrinsically motivated directed exploration strategy for continuous action spaces. We introduce a novel and scalable algorithm that directs agents to less-known state spaces with high value error through an additive explorer network.
    \item By maximizing the TD-error, our method forces the approximate value function to learn the values of unseen or less-known states or state-action pairs and prevent premature behavior before sufficiently learning the environment. This motivation also benefits from off-policy learning by extracting useful information from the collected experiences.
    \item Our technique is readily applicable to actor-critic methods that employ temporal difference learning \citep{sutton_1988}. We also show that the introduced framework comes with the cost of training a single network and optimizing a single hyper-parameter.
    \item We support our experimental results with theoretical analysis and prove that our algorithm has a faster convergence property than undirected strategies.
    \item Through an extensive set of experiments in challenging OpenAI Gym \citep{gym} continuous control tasks, we show that our algorithm improves the performance of the baseline algorithms and attains significantly higher evaluation returns in fewer time steps than the undirected strategies in the majority of the tasks tested.
    \item To ensure reproducibility, we report precise experimental and implementation details, and our code and results are available at the GitHub repository\footnote{\url{https://github.com/baturaysaglam/DISCOVER}\label{our_repo}}.
\end{itemize}

\section{Related Work}
% \subsection{Exploration}
\label{sec:related_work_exploration}
\citet{thrun_1992} classified exploration into directed and undirected strategies. Well-known methods for exploration in the control of continuous systems are random action selection with a certain probability \citep{sml_6}, additive Gaussian noise \citep{williams_1992} and Ornstein-Uhlenbeck noise process \citep{ou_noise}. However, they result in poor state space coverage due to the random walk behavior \citep{thrun_1992}. Alternative mechanisms include the employment of posterior distributions \citep{sml_7}, dimension reduction \citep{sml_10} or sampling the policy parameters from a search distribution \citep{sml_2}. Such parameter-space noise techniques can improve the exploratory behavior over the high-perturbation counterparts by reducing the variance when combined with actor-critic methods. Another alternative to the independent action perturbations is the correlation of the noises applied to the subsequent actions \citep{ou_noise, sml_5}. Although these methods are shown to be effective in parameter space exploration, their applicability is limited as they are only evaluated with shallow function approximators and operate in low-dimensional state and action spaces. 

To overcome the limited applicability of these approaches, the deep exploration techniques NoisyNet \citep{noisy_net}, Parameter Space Noise for Exploration (PSNE) \citep{psne} and Deep Coherent Exploration (Coherent) \citep{deep_coherent_exploration} were proposed. While the first two learn a parameter-space distribution over policies for trajectory-based exploration, the latter generalizes the step-based and trajectory-based exploration to perturb the last layer of the policy networks. However, several factors limit the performance of these methods. First, trajectory-based strategies can be ineffective as only a single exploration strategy is evaluated per trajectory, which may be stuck at the local optima. Secondly, Monte-Carlo integration introduces a large variance in the gradient estimates. In addition, they are computationally intensive and heavily rely on randomness, and known to fail in the off-policy setting since the exploration is separated from learning, and these methods evaluate the exploration strategy only when the exploration is performed. Hence, a deterministic or directed approach can save on time and computational resources \citep{thrun_1992}. As an alternative to the undirected methods in continuous control, \citet{meta_exploration} introduced the Meta-Policy Gradient that adapts the meta-learning to the exploration problem to obtain a learnable directed exploratory behavior through a student-teacher framework for the Deep Deterministic Policy Gradient (DDPG) algorithm \citep{ddpg}. Furthermore, the intrinsic motivation has been effectively utilized for exploration under the standard MDP setting yet for discrete action domains \citep{kearns_and_singh,sml_9}. Therefore, our comparative evaluations consider the directed exploration strategy of Meta Policy Gradient \citep{meta_exploration} in the experiments with DDPG \citep{ddpg}, and the undirected deep exploration techniques Deep Coherent Exploration \citep{deep_coherent_exploration}, NoisyNet \citep{noisy_net} and PSNE \citep{psne}.

\section{Background}
\subsection{Reinforcement Learning}
Reinforcement learning considers an agent that interacts with its environment to solve a sequential decision-making task. At every discrete time step $t$, the agent observes a state $s \in \mathcal{S}$, and selects an action $a \in \mathcal{A}$, where $\mathcal{S}$ and $\mathcal{A}$ are the state and action spaces, respectively. Depending on its action decision, the agent receives a reward $r$ from a reward function, and observes a next state $s^{\prime} \in \mathcal{S}$. In fully observable environments, the RL problem is represented by a finite Markov Decision Process, a tuple $(\mathcal{S}, \mathcal{A}, p, \gamma)$ where $p$ is the transition dynamics such that $s', r \sim p(s, a)$, and $\gamma  \in [0, 1]$ is the constant discount factor. The objective in reinforcement learning is to find an optimal policy $\pi$ that maximizes the \textit{value} defined as the expected cumulative discounted returns $R_{t} = \sum_{i = 0}^{\infty}\gamma^{i}r_{t + i + 1}$, where $\gamma$ prioritizes the short term rewards. Policy of an agent $\pi$ is stochastic if it maps states to action probabilities, $\pi: \mathcal{S} \rightarrow \mathcal{P}(\mathcal{A})$, or deterministic if it maps states to unique actions, $\pi: \mathcal{S} \rightarrow \mathcal{A}$.

In continuous action domains, the maximum of the value $R_{t}$ over possible actions is intractable. Therefore, an agent's policy is represented by a separate function that selects actions on the observed states. In deep reinforcement learning, such a function is approximated by actor networks. Algorithms that utilize actor networks are referred to as actor-critic methods, and they enable the control of continuous systems through the employed actor network. In actor-critic algorithms, actor networks $\pi_{\phi}$ parameterized by $\phi$, are optimized by gradient ascent over the policy gradient $\nabla_{\phi}J(\phi)$ computed by a policy gradient algorithm.

\subsection{On-Policy Temporal Difference Learning}
On-policy learning requires learning from experiences collected by the currently followed policy. For a given policy $\pi$, the state-value function estimates the expected return in state $s$ while following the current policy $\pi$:
\begin{equation}
    V^{\pi}(s_{t}) = \mathbb{E}_{s_{i} \sim \pi}[R_{t}\vert s_{t}], 
\end{equation}
In deep reinforcement learning, the state-value functions are modeled by differentiable function approximators $V_{\psi}(s)$ parameterized by $\psi$. The deep state-value network is learned by temporal difference learning \citep{sutton_1988}, an update rule based on the Bellman equation \citep{bellman} which represents a fundamental relationship used to learn the state-value function by bootstrapping from the value estimate of the current state $s$ to the subsequent state $s^{\prime}$:
\begin{equation}
    V_{\psi}(s) = r + \gamma\mathbb{E}_{s^{\prime}}[V_{\psi}(s^{\prime})].
\end{equation}

\subsection{Off-Policy Temporal Difference Learning}
In off-policy learning, agents collect transitions defined by the tuple $(s, a, r, s')$ and store them into the experience replay buffer \citep{experience_replay}. The collected experiences are then reused multiple times to perform gradient steps on the deep policies and value functions to improve data and sampling efficiency. There exists an action-value function $Q^{\pi}(s, a) = \mathbb{E}_{s_{i} \sim \pi, a_{i} \sim \pi}[R_{t}\vert s, a]$, associated with the policy $\pi$, also known as the critic or Q-function, which represents the expected return while following the policy after taking action $a$ in state $s$. 

% The optimal action-value function $Q^{*}(s, a) = \underset{a}{\mathrm{max}}Q^{\pi}(s, a)$ is obtained through the greedy actions of the corresponding policy, i.e., actions that yield highest action-value.

Similar to the state-value functions, the deep setting of off-policy reinforcement learning approximates action-value functions by deep neural networks $Q_{\theta}$ with parameters $\theta$. Then, the deep action-value functions is learned through TD-learning \citep{sutton_1988} by bootstrapping from the estimate of the current state-action pair $(s, a)$ to the subsequent state-action pair $(s^{\prime}, a^{\prime})$, also known as Q-learning \citep{watkins_1989}:
\begin{equation}
    Q_{\theta}(s, a) = r + \gamma\mathbb{E}_{s^{\prime}, a^{\prime}}[Q_{\theta^{\prime}}(s^{\prime}, a^{\prime})]; \quad a^{\prime} \sim \pi_{\phi^{\prime}}(s^{\prime}),
\end{equation}
where $\theta^{\prime}$ and $\phi^{\prime}$ are the parameters of the secondary frozen networks, $Q_{\theta^{\prime}}(s, a)$ and $\pi_{\phi^{\prime}}(s)$, to maintain stability and fixed objective over high-frequency updates. In practical applications of deep RL, the next action chosen by the behavioral or target policy network in the TD-learning \citep{sutton_1988} is perturbed by random action noise to realize the exploration in the next state, known as the target policy smoothing regularization \citep{td3}:
\begin{equation}
    Q_{\theta}(s, a) = r + \gamma\mathbb{E}_{s^{\prime}, a^{\prime}}[Q_{\theta^{\prime}}(s^{\prime}, \Tilde{a}^{\prime})]; \quad \Tilde{a}^{\prime} \sim a^{\prime} + \mathcal{N}(0, \sigma),
\end{equation}
where $\mathcal{N}(0, \sigma)$ is a zero-mean Gaussian random variable with standard deviation $\sigma$. The weights of the target networks can be updated through the soft or hard update. In the soft update, parameters are updated with a small proportion $\tau$ at each time step, $\theta^{\prime} \leftarrow \tau\theta + (1 - \tau)\theta^{\prime}$, while hard update considers periodically updating the weights to exactly match the weights of the current or behavioral network. 

\section{Deep Directed Exploration Motivated by the Temporal Difference Error}
\textit{Motivation} is a term that describes the processes which impact arousal, strength, and action direction \citep{ryan_and_deci}. To be motivated is to feel compelled to do an action \citep{ryan_and_deci}. Psychologists distinguish the motivation into two classes, \textit{extrinsic motivation}, taking action because of externally supplied rewards, and \textit{intrinsic motivation}, doing something for its own sake as it is inherently enjoyable or interesting \citep{ryan_and_deci}. Intrinsic motivation leads consciousness to engage in play, competence, exploration, and other curiosity-driven behavior in the absence or inadequacy of the externally supplied rewards \citep{ryan_and_deci}. 

The concept of motivation is not directly associated with machine learning \citep{simsek_im}. However, there is a conformity between the animal motivation and RL systems in that both have an objective of maximizing the external rewards \citep{simsek_im}. The remarkable resemblance between the critical components of RL algorithms and the activity of dopamine neurons \citep{dopamine_neurons} play an essential, though poorly understood, role in animal motivational systems \citep{simsek_im}. Although there is still a notable gap between animal motivational systems and computational RL algorithms \citep{simsek_im}, it can be argued that the gradient of an RL agent's value function similarly directs the behavior to ``incentive salience" \citep{incentive_salience}. This cognitive process directs an individual's behavior towards an object \citep{simsek_im}. Hence, in consistency with the existing theories on animal motivation, the gradient of an RL system's evaluation function motivates its behavior \citep{simsek_im}. 

The intrinsically motivated exploration problem requires agents to change their behavior to quickly adapt the skills in retrieving the intrinsically-rewarding event \citep{simsek_im}. This is what motivation entails: the agent must ``want" to bring about the event in interest, and this desire must be expressed by its actions \citep{simsek_im}. This is a concept that pseudo-reward functions cannot accomplish \citep{simsek_im}. A corollary to this is that intrinsic reward should gradually diminish as the action starts to be repeated, that is, the agents should ultimately become ``bored" and move on to develop and learn a new alternative \citep{simsek_im}. Therefore, for an intrinsically rewarding exploration, the change in the agent's evaluation function should be consistently kept maximum. 

Furthermore, we know for actor-critic methods that the state- or action-value function serves as the evaluation function of the agent, and the gradients of which are directly proportional to the change in the TD-error \citep{sutton_book}. Therefore, an RL agent in continuous action spaces can be motivated by a consistent maximization of the TD-error to obtain an exploratory behavior. The TD-error based intrinsic motivation for on- and off-policy learning are summarized in Remark \ref{rem:in_mot_on} and \ref{rem:in_mot_off}, respectively. 

\begin{remark}[Intrinsic motivation for exploration in on-policy learning]
\label{rem:in_mot_on}
    For on-policy learning in continuous action spaces, a reinforcement learning agent can be motivated by a consistent maximization of the prediction error by its state-value function $V(s)$. 
\end{remark}

\begin{remark}[Intrinsic motivation for exploration in off-policy learning]
\label{rem:in_mot_off}
    For off-policy learning in continuous action spaces, a reinforcement learning agent can be motivated by a consistent maximization of the prediction error by its action-value function or critic $Q(s, a)$. 
\end{remark}

\subsection{On-Policy Intrinsically Motivated Exploration}
To construct our directed and intrinsically motivated exploration framework, we first consider a separate, deep deterministic \textit{exploration policy}, $\xi_{\omega}: \mathcal{S} \rightarrow \mathcal{E}$, with parameters $\omega$, that maps states to unique exploratory directions. Moreover, we assume that the state-value function estimates the value of a state by also considering the exploration directions that are in the subspace of the state space $\mathcal{E} \in \mathcal{S}$. Similarly, the policy also considers the exploration directions while selecting actions. Thus, the exploration policy can influence the distribution of the states received by the state-value function and policy under the parameters $\omega$. Let $V^{*}(s, \eta)$ be the optimal state value for a given state $s$ and exploration direction $\eta$ under the optimal policy $\pi^{*}(\cdot \vert s, \eta)$, defined by the Bellman equation \citep{bellman}:
\begin{equation}
    V^{*}(s, \eta) = r + \gamma V^{*}(s^{\prime}, \eta^{\prime})\vert_{\eta^{\prime} = \lambda \cdot \xi_{\omega}(s^{\prime})},
\end{equation}
where $\eta^{\prime}$ is the exploration direction in the next step and $\lambda$ is a regularization term for the exploration policy not to detrimentally perturb the state distribution. By Remark \ref{rem:in_mot_on}, intrinsically motivated exploration can be considered as an adversarial game such that the state-value function tries to minimize the prediction error while the exploration policy aims to perturb the state distribution so that prediction error is maximized. Therefore, the objective for the joint optimization of the state-value and exploration networks is expressed by:
\begin{equation}
    \underset{\omega}{\mathrm{max}}\underset{\psi}{\mathrm{min}}\sum_{s \sim p_{\pi}}\|V^{*}(s, \eta) - V_{\psi}(s, \eta)\|^{2}_{\eta = \lambda \cdot \xi_{\omega}(s)}.
\end{equation}
However, the knowledge of the optimal state-value function beforehand is not possible in practical applications of deep reinforcement learning. A common approach in on-policy learning is to regard empirical reversed sum of rewards $\hat{R}$, i.e., rewards-to-go, as the fixed objective for the state-value function, defined by:
\begin{equation}
\label{eq:rewards_to_go_def}
    \hat{R}_{t} = \sum_{i = t}^{T}R(s_{i}, a_{i}, s_{i + 1})
\end{equation}
where $T$ is the length of the horizon for which rollout transitions are collected to train the agent. Using Equation (\ref{eq:rewards_to_go_def}) and considering the exploration directions, the objective for the state-value network is expressed by:
\begin{equation}
        J(\psi) = \|\hat{R}- V_{\psi}(s, \eta)\|^{2}_{\eta = \lambda \cdot \xi_{\omega}(s)}.
\end{equation}
Then, exploration policy's objective is in the inverse direction of the state-value function's objective:
\begin{align}
\label{eq:on_policy_obj}
    \begin{split}
        J(\omega) &= -J(\psi), \\
        &= -\|\hat{R} - V_{\phi}(s, \lambda \cdot \xi_{\omega}(s))\|^{2}, \\
        &\coloneqq -\Phi(s, \lambda \cdot \xi_{\omega}(s)).
    \end{split}
\end{align}
Ultimately, the objective of the exploration policy is to perturb the state distribution so that the prediction error by the state-value network is maximized. This leads agents to state spaces where state value prediction is difficult and hence, allows the correction of unknown or less visited states' values. Furthermore, the exploration network can be updated through the Deterministic Policy Gradient (DPG) Theorem \citep{dpg} using the constructed objective Equation (\ref{eq:on_policy_obj}):
\begin{equation}
    \nabla_{\omega} J(\omega) = \mathbb{E}_{s \sim p_{\pi}}[\nabla_{\eta}\Phi(s, \eta)\vert_{\eta = \lambda \cdot \xi_{\omega}(s)}\nabla_{\omega}\xi_{\omega}(s)].
\end{equation}
Note that the existence of the explorer network's gradient follows the Deterministic Policy Gradient Theorem \citep{dpg}. During the evaluation, however, no exploration is performed. Hence, the state-value and actor networks should not consider any direction that perturbs the observed states' distribution. This can be achieved by simply replacing the exploration directions with zeros. We refer to the resulting on-policy variant of our algorithm as \textbf{On-Policy} \textbf{D}eep \textbf{I}ntrin\textbf{SC}ally M\textbf{O}ti\textbf{V}ated \textbf{E}xplo\textbf{R}ation (On-Policy DISCOVER), and provide a pseudocode in Algorithm \ref{alg:discover_on}.

\begin{algorithm}[t]
    \caption{On-Policy Deep Directed Intrinsically Motivated Exploration (On-Policy DISCOVER)}
    \begin{algorithmic}[1]
        \STATE Initialize the explorer network $\xi_{\omega}$ with parameters $\omega$
        \FOR{each rollout time step}
            \STATE Observe state $s$
            \STATE Obtain the exploration direction: $\eta = \lambda \cdot \xi_{\omega}(s)$
            \STATE Select an action considering the exploration direction:
            $a = \pi_{\phi}(s, \eta)$
            \STATE Receive reward $r$ and observe next state $s^{\prime}$
            \STATE Store the transition tuple $(s, a, \eta, r, s^{\prime})$ into the rollout buffer
        \ENDFOR
        \FOR{each policy update step}
        \STATE Update $\omega$ by the deterministic policy gradient and mini-batch learning on the rollouts:\\$\nabla_{\omega} J(\omega) = N^{-1}\sum\nabla_{\eta}\Phi(s, \eta)\vert_{\eta = \lambda \cdot \xi_{\omega}(s)}\nabla_{\omega}\xi_{\omega}(s),$
        where $N$ is the mini-batch size
        \STATE Update policy and state-value networks by considering the sampled exploration directions $\eta$
        \ENDFOR
    \end{algorithmic}
    \label{alg:discover_on}
\end{algorithm}

\subsection{Off-Policy Intrinsically Motivated Exploration}
For off-policy learning, now consider that the exploration directions are in a subspace of the action space $\xi_{\omega}: \mathcal{S} \rightarrow \mathcal{E} \in \mathcal{A}$. We additionally initialize a secondary frozen target explorer network with parameters $\omega^{\prime}$, to obtain a fixed objective and stability in the explorer network updates, similar to the Deep Q-learning algorithm \citep{dqn}. The objective of the exploration policy is to maximize the TD-error of the Q-networks in consistency with the observed states and actions selected by the policy. First, let the actions chosen by the policy be perturbed by the explorer networks:
\begin{gather}
        \Tilde{a} = a + \lambda \cdot \xi_{\omega}(s),\label{eq:noisy_action_definition}  \\
        \Tilde{a}^{\prime} = \pi_{\phi^{\prime}}(s^{\prime}) + \lambda \cdot \xi_{\omega^{\prime}}(s^{\prime}),\label{eq:noisy_target_action_definition}
\end{gather}
where $\lambda$ now regularizes the additive exploration direction for the actions. Note that the action in the next state may also be chosen by the behavioral policy such as in the SAC algorithm \citep{sac}. Furthermore, to realize exploration in the next state, target policy smoothing regularization is obtained through the target explorer network in Equation (\ref{eq:noisy_target_action_definition}), which is shown to improve the performance of off-policy methods \citep{td3}.

As discussed in Remark \ref{rem:in_mot_off}, TD-error should be kept maximum by considering the selected actions. Thus, parameters of the exploration policy can be updated through gradient ascent over the error by the Q-network. This constitutes the loss function for the exploration policy in the form of an implicit adversarial game:
\begin{align}
    \begin{split}
      y &= r + \gamma Q_{\theta^{\prime}}(s^{\prime}, \Tilde{a}^{\prime}), \\
      J(\omega) &= -\|y - Q_{\theta}(s, \Tilde{a})\|^{2}, \\
      &\coloneqq -\Tilde{\delta}(s, a + \lambda \cdot \xi_{\omega}(s)).
    \end{split}
\end{align}
where $\Tilde{\delta}$ is the off-policy TD-error under the perturbed actions $\Tilde{a}$. Then, the exploration policy is again updated through the DPG algorithm \citep{dpg}:
\begin{equation}
        \nabla_{\omega} J(\omega) = \mathbb{E}_{s \sim p_{\pi}, a \sim \pi_{\phi}(\cdot \vert s)}[\nabla_{\eta}\Tilde{\delta}(s, a + \eta)\vert_{\eta = \lambda \cdot \xi_{\omega}(s)}\nabla_{\omega}\xi_{\omega}(s)].
\end{equation}
Moreover, to apply the target policy smoothing regularization \citep{td3} in the Q-network updates and maintain the exploration in the action-value estimates, we use the actions in the current and next state that are perturbed by the behavioral and target explorer networks, respectively, as expressed in Equation (\ref{eq:noisy_target_action_definition}). This allows the directed exploration to be maintained in the critic updates:
\begin{equation}\label{eq:perturbed_next_action}
    J(\theta) = \|y - Q_{\theta}(s, \Tilde{a})\|^{2}.
\end{equation}
This forms the off-policy variant of DISCOVER, which is summarized in Algorithm \ref{alg:discover_off}. In the next section, we explain how DISCOVER can be effectively combined with actor-critic algorithms to control continuous systems. Moreover, we perform a comprehensive theoretical comparison of the complexity of the introduced directed exploration scheme and undirected exploration strategies. 

\begin{algorithm}[t]
    \caption{Off-Policy Deep Directed Intrinsically Motivated Exploration (Off-Policy DISCOVER)}
    \begin{algorithmic}[1]
        \STATE Initialize the explorer network $\xi_{\omega}$ with parameters $\omega$
        \STATE Initialize the target explorer network $\omega^{\prime} \leftarrow \omega$
        \FOR{each exploration time step}
            \STATE Choose an action $a$
            \STATE Perturb the selected action $\Tilde{a} = a + \lambda \cdot \xi_{\omega}(s)$
            \STATE Observe reward $r$ and next state $s^{\prime}$
            \STATE Store transition tuple $(s, a, \Tilde{a}, r, s^{\prime})$ into the replay buffer
        \ENDFOR
        \FOR{each Q-network update step}
            \STATE Perturb the next action through the target explorer network to modify the fixed objective for Q-learning: $\Tilde{a}^{\prime} = a^{\prime} + \lambda \cdot \xi_{\omega^{\prime}}(s^{\prime})$
        \ENDFOR
        \FOR{each policy update step}
        \STATE Update $\omega$ by the deterministic policy gradient through the sampled batch of $N$ transitions:\\$\nabla_{\omega} J(\omega) = N^{-1}\sum\nabla_{\eta}\Tilde{\delta}(s, a + \eta)\vert_{\eta = \lambda \cdot \xi_{\omega}(s)} \nabla_{\omega}\xi_{\omega}(s)$
        \ENDFOR
        \FOR{each target policy update period}
            \STATE Update the target explorer network: $\omega^{\prime} \leftarrow \tau\omega + (1 - \tau)\omega^{\prime}$
        \ENDFOR
    \end{algorithmic}
    \label{alg:discover_off}
\end{algorithm}

\subsection{Actor-Critic with Deep Directed Intrinsically Motivated Exploration}
Although the introduced exploration strategy is based on the prediction error in the TD-learning \citep{sutton_1988}, the consistent maximization of the TD-error unifies the intrinsic motivations in the directed exploration as discussed in Remark \ref{rem:prediction_error}, \ref{rem:state_novelty} and \ref{rem:info_gain} for the prediction error, state novelty and information gain incentives, respectively. 

\begin{remark}[Prediction error]
\label{rem:prediction_error}
    The TD-error maximization objective leads the approximate state- and action-value functions to learn the transitions with high TD-error where predicting states or state-action tuples is difficult. This is performed in on-policy learning by perturbing the distribution under the observed states or actions chosen by the policy for off-policy learning.
\end{remark}

\begin{remark}[State novelty]
\label{rem:state_novelty}
    If the agent repeatedly visits the same state space in state novelty, the intrinsic reward corresponding to that subspace decreases. Hence, a state space may be repeatedly visited in undirected methods due to randomness \citep{thrun_1992}. However, in the deterministic TD-error maximization objective, the agent is constantly forced to visit the tuples with the highest TD-error. Thus, the agent always seeks novel state spaces without any possibility for a repetition in the action selection. 
\end{remark}

\begin{remark}[Information gain]
\label{rem:info_gain}
    High TD-error implies the uncertainty on environment dynamics as the expected sum of rewards in the corresponding state space, and the underlying MDP is not well-known by the value function \citep{sutton_book}. The explorer network aims to reduce such uncertainty by learning the value of uncertain states or state-action pairs through the additive exploration direction.
\end{remark}

For off-policy learning, the intrinsically motivated exploration can also consider the agent's history of learning process by training from the batch of transitions sampled from the experience replay buffer \citep{experience_replay}. However, as discussed, the separated updates of exploration strategy and policies suffer from off-policy learning \citep{sml_11}. Therefore, the exploration policy should be evaluated and updated whenever the policy of the underlying actor-critic algorithm is updated. This yields that the exploration framework is synchronized to the agent's policy update, with having the same update frequency and update rule of the target networks, i.e., soft or hard update. Furthermore, the exploration policy should be identically structured with the agent's policy when combined with both on- and off-policy algorithms. The motivation of this imitation is that the exploration policy should be fairly as capable as the agent's policy. If the agent's policy can learn to control the environment through a policy gradient algorithm, the exploration policy can also learn to maximize the TD-error with DPG \citep{dpg} under the same set of hyper-parameters and update structure, which we show through empirical studies in the next section. This requirement is highlighted in Remark \ref{rem:imitation}

\begin{remark}
\label{rem:imitation}
    When combined with actor-critic algorithms, the structure of the exploration policy should follow the exact architecture and hyper-parameter setting of the underlying algorithm's policy, e.g., learning rate, size and depth of the actor network, update period, and utilization of target networks.
\end{remark}

The large computational complexity remains as a detrimental drawback in the undirected methods \citep{thrun_1992}. We investigate the time complexity of the introduced exploration strategy in Theorem \ref{thm:discover_det_time_comp} under ergodic deterministic MDPs. Moreover, Lemma \ref{thm:exp_sto_time_comp} remarks the time complexity of any exploration strategy under ergodic non-deterministic MDP setting. We then perform a time complexity comparison of our approach versus undirected strategies in Corollary \ref{thm:convergence_rate_comp}, from which we derive that the time to find an optimal policy under the exploration motivated by maximum TD-error is always less than or equal to the undirected exploration. In the following section, we show through an extensive set of simulations that DISCOVER exhibits remarkable results in on- and off-policy settings, outperforming the state-of-the-art exploration techniques. Furthermore, we demonstrate that only a single hyper-parameter choice is necessary for a wide range of tasks when the explorer network is identically structured and synchronized to the agent's actor network, validating Remark \ref{rem:imitation}.

\begin{lemma}[Whitehead's theorem on undirected exploration complexity in deterministic MDPs]
\label{thm:undirected_det_time_comp}
    Under deterministic Markov Decision Process conditions, the expected time to find the goal state using undirected exploration is bounded below by an expression exponential in the state dimension $n$.
    
\begin{proof}
    See \citep{whitehead}.
    \end{proof}
\end{lemma}

\begin{lemma}[Time complexity of the exploration under ergodic stochastic MDPs]
\label{thm:exp_sto_time_comp}
    For each $n \in \{2, 3, \dots\}$, there is an ergodic non-deterministic Markov Decision Process, at which the time complexity for reaching the goal state under exploration is exponential in $n$, even if the optimal policy is given in advance. 
    
    \begin{proof}
    See \citep{thrun_1992}. 
    \end{proof}
\end{lemma}

\begin{theorem}[Time complexity of the exploration motivated by the maximum temporal difference error under ergodic deterministic MDPs]
\label{thm:discover_det_time_comp}
    With exploration directed by the maximum temporal difference error, the number of actions required to find an optimal policy in any finite ergodic deterministic Markov Decision Process with a goal state is bounded above by $\mathcal{O}((d^{2}m + m^{2})n^{2})$, given the number of unknown states or state-action pairs $m$, and time complexity for each iteration in the temporal difference learning $\mathcal{O}(d)$.
    
    \begin{proof}
    Follows from the extension of Theorem 3 in \citep{thrun_1992} to the exploration motivated by the maximum TD-error. The TD-learning \citep{sutton_1988} update rule at iteration $t$ following the current policy $\pi$ is expressed for on- and off-policy learning, respectively, as:
    \begin{align}
        V_{t + 1}^{\pi}(s) &= r + \gamma V^{\pi}_{t}(s^{\prime}),\label{eq:proof_policy_iter} \\
        Q_{t + 1}^{\pi}(s, a) &= r + \gamma \underset{a^{\prime}}{\mathrm{max}}Q_{t}^{\pi}(s^{\prime}, a^{\prime}).\label{eq:proof_q_learning}
    \end{align}
    First, the equations imply that $V^{\pi}(\cdot)$ and $Q^{\pi}(\cdot)$ are monotonically increasing in the learning time. Then, consider an optimal path $\langle s_{1} = s_{init}, s_{1}, s_{2}, \dots, s_{k} = s_{goal}\rangle$ from some initial state $s_{init}$ to a goal state $s_{goal}$. If $V^{\pi}(s_{i})$ is correct for some state $s_{i}$ on this path ($0 < i \leq k$), then applying the TD-learning \citep{sutton_1988} in state $s_{i - 1}$ will assign the correct state-value to $s_{i-1}$ \citep{thrun_1992}. This is also valid for Q-learning \citep{watkins_1989} denoted by Equation (\ref{eq:proof_q_learning}), specifically, if $Q^{\pi}(s_{i}, a_{i})$ is correct for some state-action tuple $(s_{i}, a_{i})$ on this path ($0 < i \leq k$), then applying the TD-learning \citep{sutton_1988} at state-action pair $(s_{i - 1}, a_{i - 1})$ will assign the correct action-value to $(s_{i-1}, a_{i-1})$ \citep{thrun_1992}. 
    
    Learning the environment will identify the correct value function and hence, an optimal policy if the goal state exists. The exploration strategy in interest will direct the agent to the unknown state space with the highest TD-error and apply the TD-learning \citep{sutton_1988} update rule. Therefore, the value of a unique state or state-action pair will be corrected in each exploration run since always the one with the highest TD-error is visited. If there are $m$ unknown entities, i.e., state or state-action pair, then after $m$ iterations, the optimal policy is found. However, learning the environment requires identifying the action model, i.e., the outcome of each action. This can be done with at most $\mathcal{O}(d^{2}mn^{2})$ actions \citep{thrun_1992}. Furthermore, by Theorem 1 shown by \citet{thrun_1992}, the complexity of each exploration run is bounded above by $\mathcal{O}(mn^{2})$. Thus, the resulting worst-case complexity for an exploration directed by the maximum TD-error under ergodic deterministic MDPs is then in $\mathcal{O}((d^{2}m + m^{2})n^{2})$.
\end{proof}

\end{theorem}

\begin{corollary}
\label{thm:convergence_rate_comp}
    In any finite ergodic Markov Decision Process, the expected time to find an optimal policy using exploration directed by the maximum temporal difference error is bounded above the expected time using undirected exploration. 
    
\begin{proof}
    The proof infers from Lemma \ref{thm:undirected_det_time_comp}, \ref{thm:exp_sto_time_comp}, and Theorem \ref{thm:discover_det_time_comp}. The polynomial or quadratic time is bounded above by the exponential time. Hence, by Lemma \ref{thm:undirected_det_time_comp} and Theorem \ref{thm:discover_det_time_comp}, the TD-error based exploration is bounded above by undirected exploration for ergodic deterministic MDPs. From Lemma \ref{thm:exp_sto_time_comp}, we know that both maximum TD-error based directed exploration and undirected exploration are bounded below by the time complexity exponential in the state space dimension for ergodic stochastic MDPs. Thus, exploration directed by maximum TD-error has a time complexity that is either the same or less than undirected exploration's time complexity for any MDP, in the worst-case scenario. 
\end{proof}
\end{corollary}

\section{Experiments}
\subsection{Experimental Details}
We perform experiments to demonstrate the effectiveness of our method, and compare it with other exploration strategies: NoisyNet \citep{noisy_net}, PSNE \citep{psne}, Deep Coherent Exploration \citep{deep_coherent_exploration}, standard Gaussian action noise \citep{williams_1992} and greedy action selection. We evaluate each method in combination with on-policy actor-critic algorithms, Advantage Actor-Critic (A2C) \citep{a2c}, Proximal Policy Optimization (PPO) \citep{ppo}, and off-policy actor-critic algorithms, Deep Deterministic Policy Gradient (DDPG) \citep{ddpg}, Soft Actor-Critic (SAC) \citep{sac} and Twin Delayed Deep Deterministic Policy Gradient (TD3) \citep{td3}. We also consider the directed strategy  Meta-Policy Gradient \citep{meta_exploration} in our experiments with DDPG \citep{ddpg}. We focus on challenging MuJoCo \citep{mujoco} and Box2D \citep{box2d} continuous control tasks interfaced by OpenAI Gym \citep{gym}. Algorithm-specific adaptations of DISCOVER can be found in our code\footref{our_repo}.

Our implementation of NoisyNet \citep{noisy_net} is based on the code from authors' GitHub repository\footnote{\url{https://github.com/Kaixhin/NoisyNet-A3C}\label{noisy_net_repo}}. We refer to the authors' implementation in OpenAI Baselines\footnote{\url{https://github.com/openai/baselines}\label{baselines_repo}} \citep{openai_baselines} for PSNE \citep{psne}, and original papers for Deep Coherent Exploration \citep{deep_coherent_exploration} and SAC \citep{sac}. In addition, authors' GitHub repository\footnote{\url{https://github.com/sfujim/TD3}\label{td3_repo}} is used to implement the fine-tuned TD3 \citep{td3} and baseline DDPG \citep{ddpg}. Finally, we implement the on-policy algorithms A2C \citep{a2c} and PPO \citep{ppo} through the well-known GitHub repository\footnote{\url{https://github.com/ikostrikov/pytorch-a2c-ppo-acktr-gail}\label{ppo_repo}} with tuned hyper-parameters for the continuous control tasks in OpenAI Gym \citep{gym}.

DDPG \citep{ddpg} and SAC \citep{sac} follow the tuned hyper-parameters provided by OpenAI Baselines3 Zoo\footnote{\url{https://github.com/DLR-RM/rl-baselines3-zoo}\label{zoo_repo}} \citep{openai_zoo}. For the action space noise, we use a fixed zero-mean Gaussian distribution with a standard deviation of 0.1. Competing algorithms closely follow the hyper-parameter settings outlined in the original papers. For the parameter-space noise algorithms, we initialize the parameter noise at 0.017 for on-policy and 0.034 for off-policy algorithms as they were found to produce the best results in \citep{deep_coherent_exploration}. In addition, we use $\beta = 0.01$ for all environments in the Deep Coherent Exploration algorithm \citep{deep_coherent_exploration}, and set the mean-squared error threshold in PSNE \citep{psne} to 0.1. 

As discussed in Remark \ref{rem:imitation}, we initialize our exploration framework to exactly match the policy network's structure and hyper-parameter setting for all baselines. This initialization includes the size and depth of the networks, optimizer, learning rate, frequency of the updates, and weight decay if it exists. In addition, we utilize target explorer networks in off-policy algorithms for which the underlying actor-critic methods' target network update rule, i.e., soft or hard update, and learning rate also apply to the target explorer network. This yields the delayed explorer updates in the TD3 algorithm \citep{td3} also apply to the explorer network. We consider six values for the exploration regularization term $\lambda = \{0.0, 0.1, 0.3, 0.6, 0.9, 1.0\}$, where we use $\lambda = 0.1$ for On-Policy DISCOVER and $\lambda = 0.3$ for Off-Policy DISCOVER. Exact experimental setup and hyper-parameter settings are found in Appendix \ref{app:imp_details}.

Each algorithm is run for 1 million time steps with evaluations every 1000 steps. Each evaluation reports the average cumulative return over 10 episodes in a distinct evaluation environment without exploration and updates. We report the results over 10 random seeds of the OpenAI Gym \citep{gym} simulator, network initialization, and dependencies. The environment dynamics, state and action spaces, and reward functions are not pre-processed and modified for easy reproducibility and fair evaluation procedure.

\subsection{Comparative Evaluation and Discussion}
\label{sec:comp_eval}

% Evaluation Results - A2C
\begin{figure*}[!hbt]
    \centering
    \begin{align*}
        &\text{{\blue} DISCOVER-A2C ($\lambda$ = 0.1)}  &&\text{{\orange} Coherent-A2C ($\beta$ = 0.01)} &&\text{{\purple} NoisyNet-A2C} \\
        &\text{{\green} PSNE-A2C} &&\text{{\red} A2C}
    \end{align*}
	\subfigure{
		\includegraphics[width=2.0in, keepaspectratio]{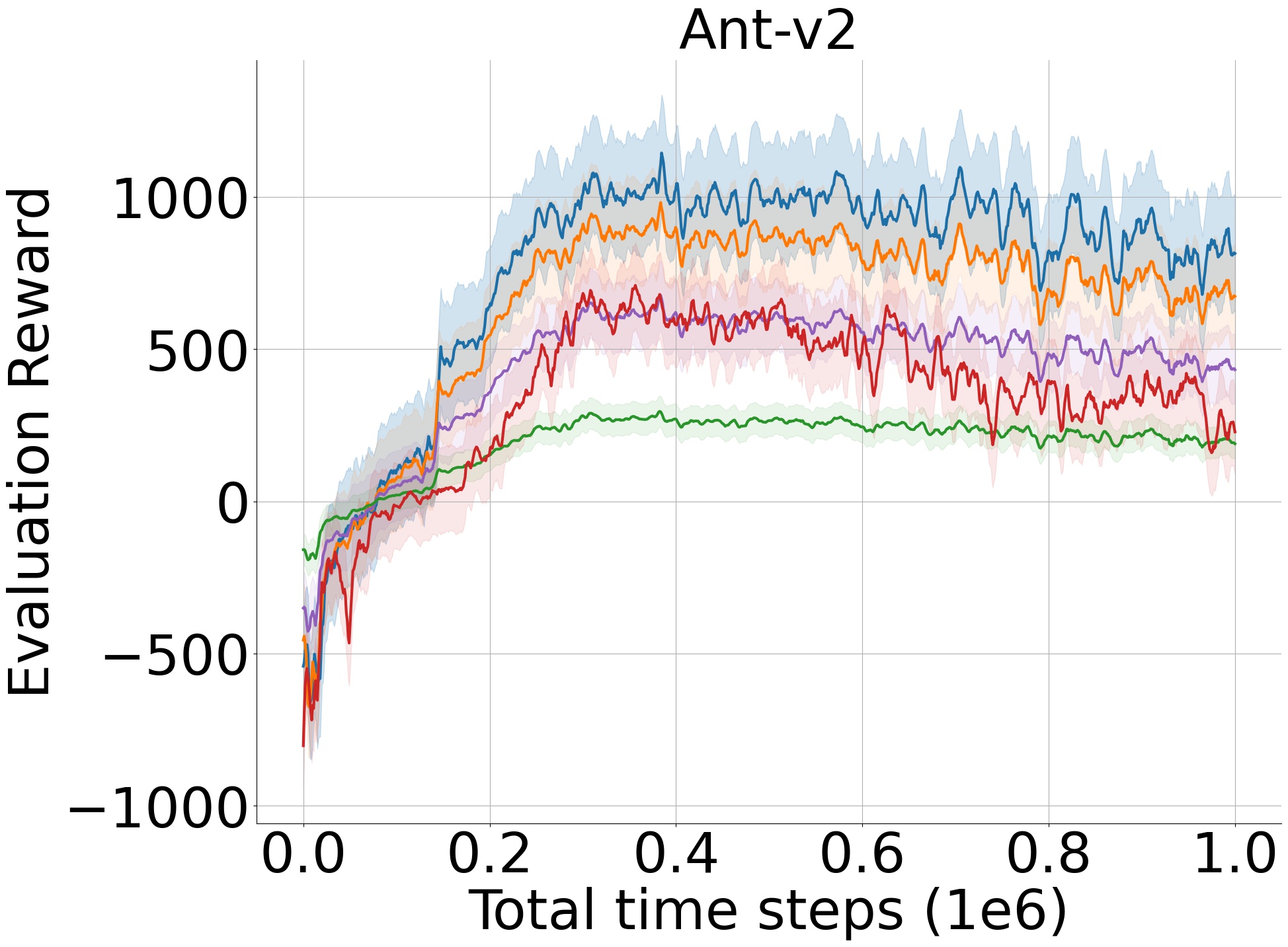}
		\includegraphics[width=2.0in, keepaspectratio]{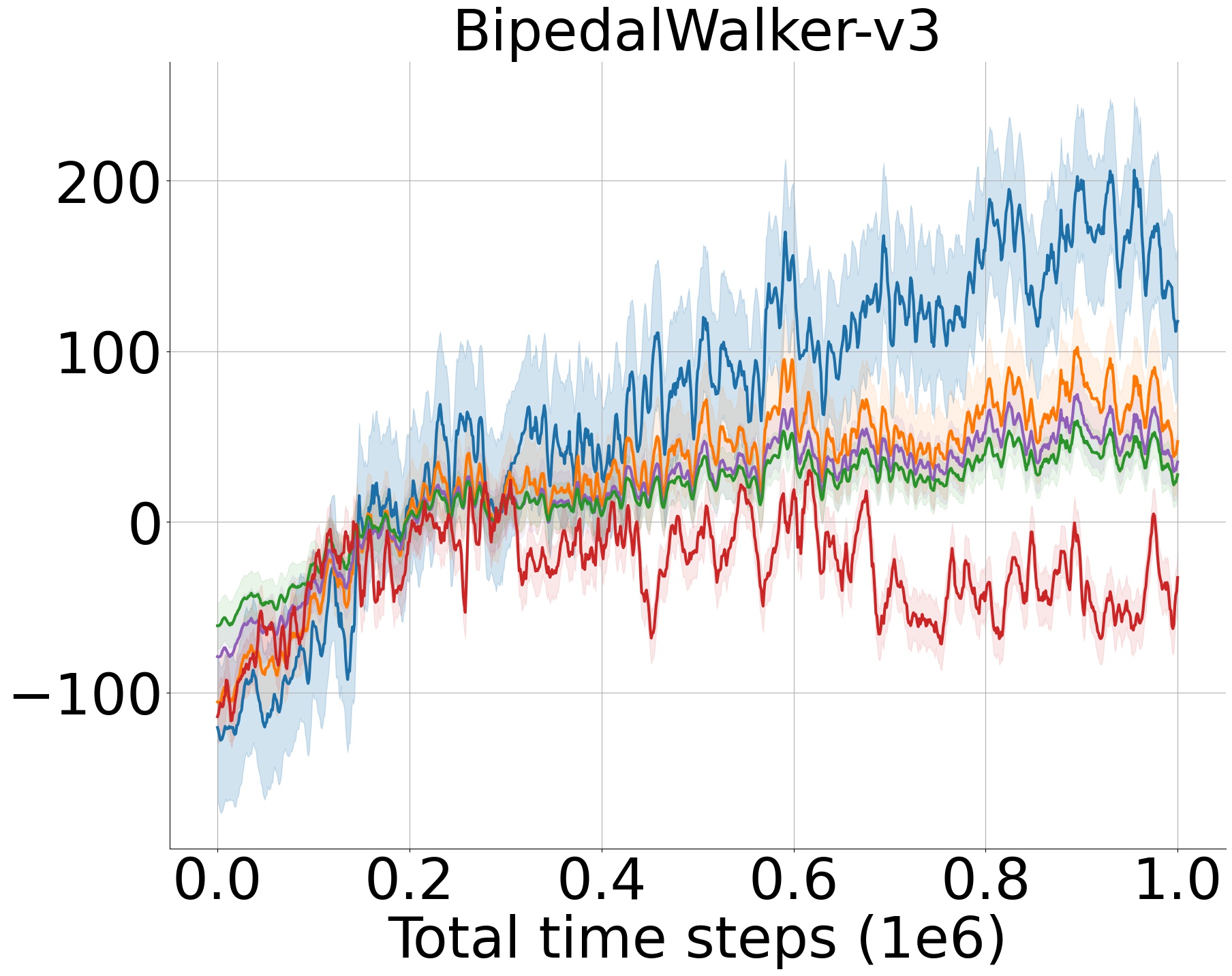}
		\includegraphics[width=2.0in, keepaspectratio]{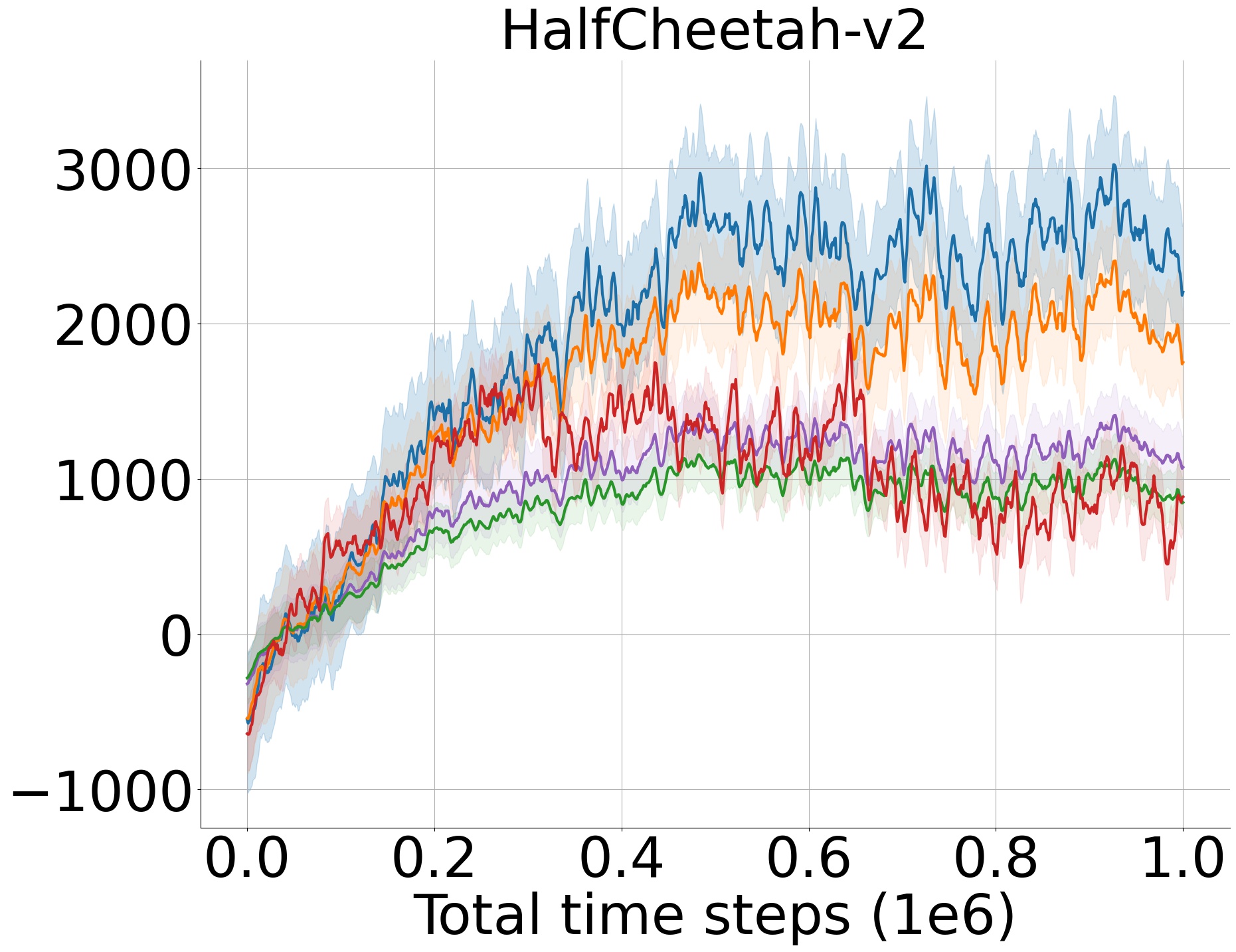}
	} \\
	\subfigure{
	    \includegraphics[width=2.0in, keepaspectratio]{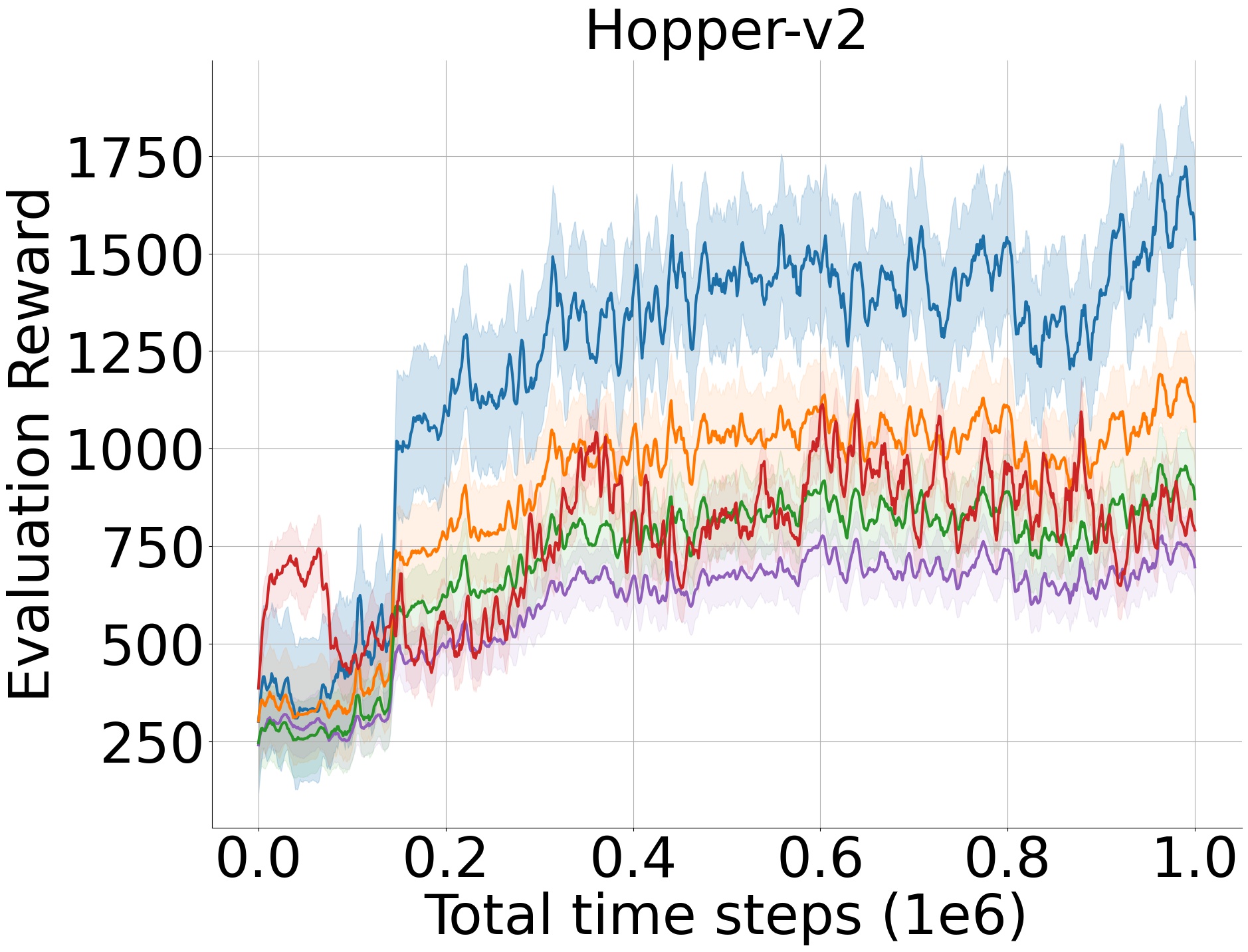}
		\includegraphics[width=2.0in, keepaspectratio]{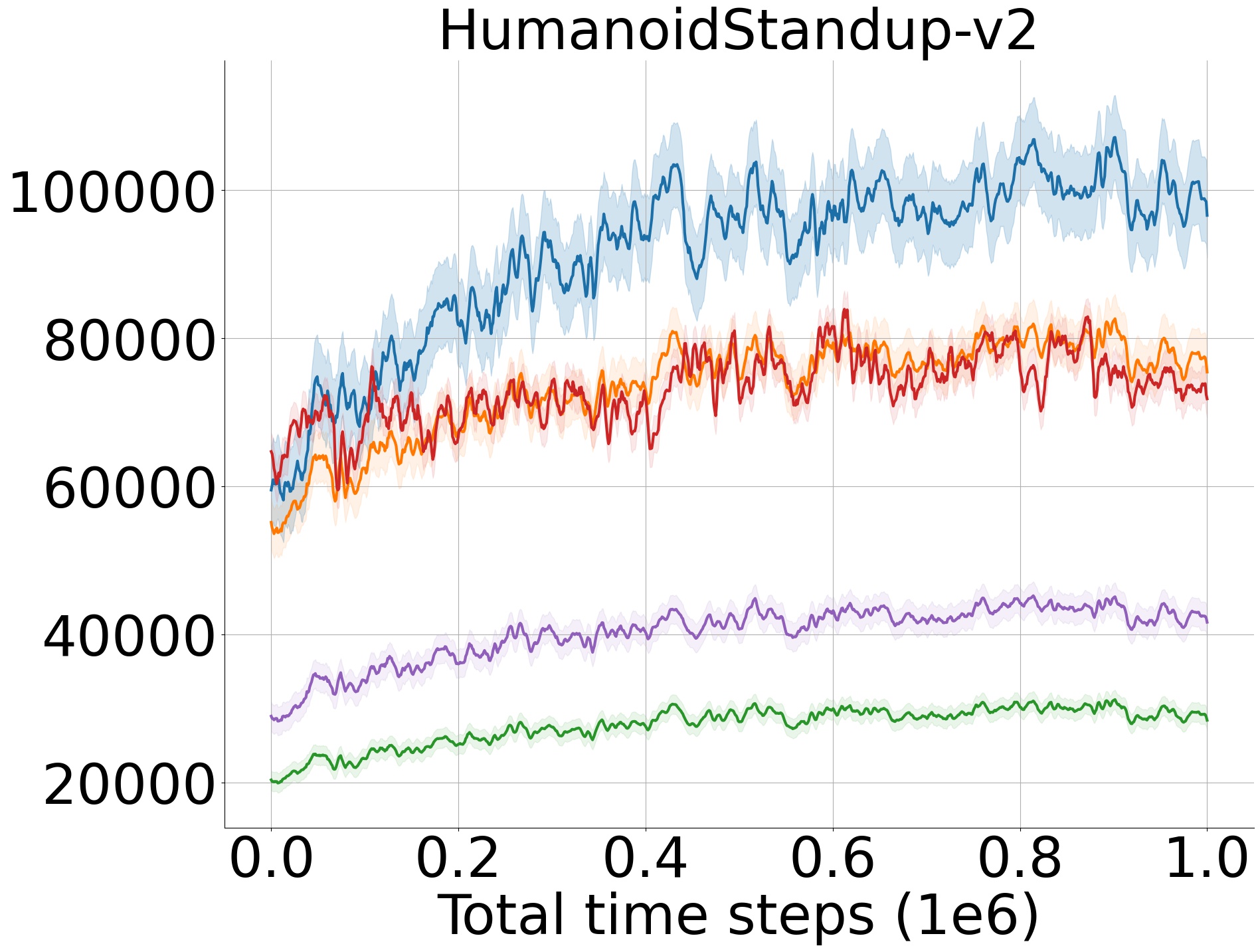}
		\includegraphics[width=2.0in, keepaspectratio]{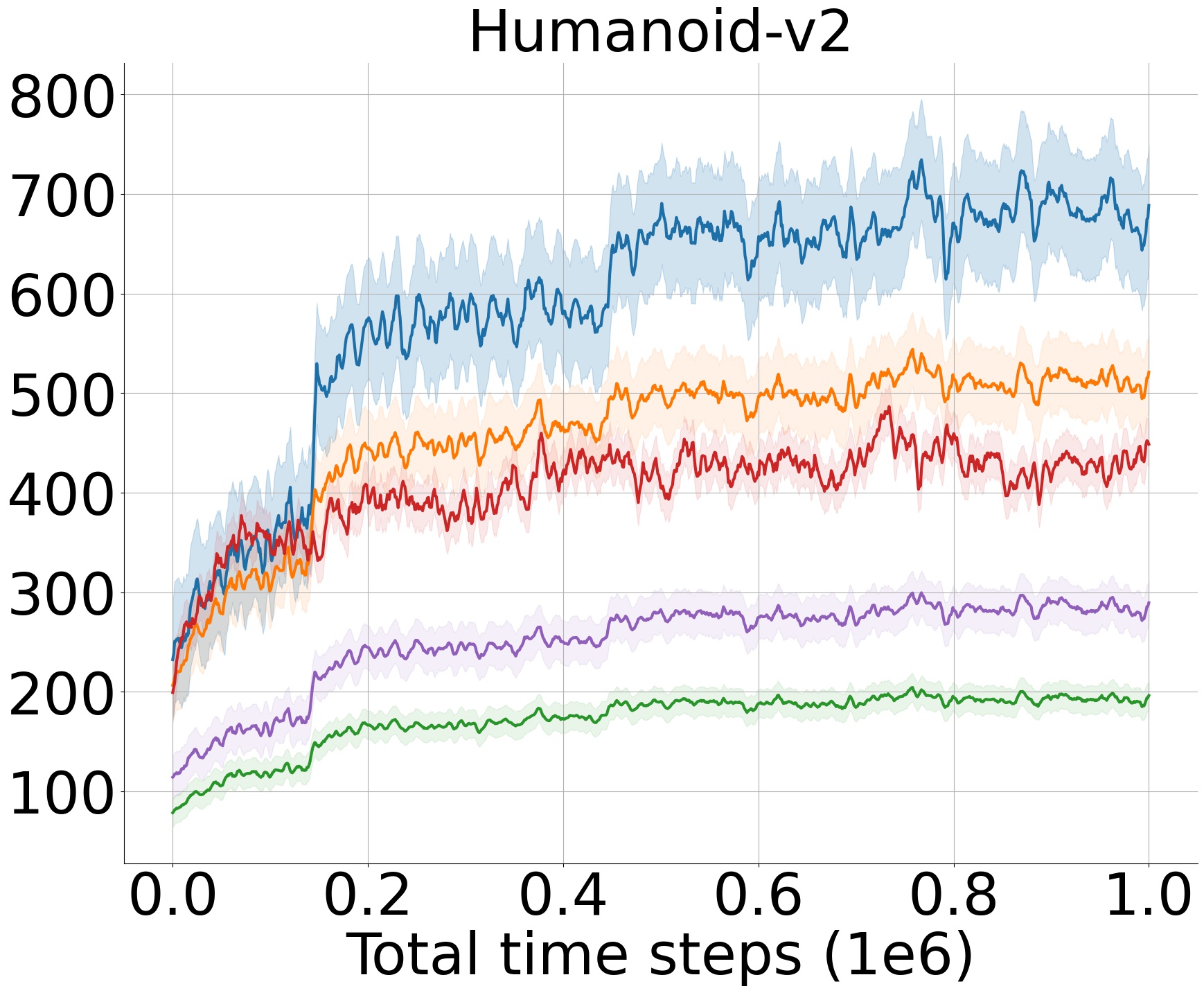}
    } \\
	\subfigure{
		\includegraphics[width=2.0in, keepaspectratio]{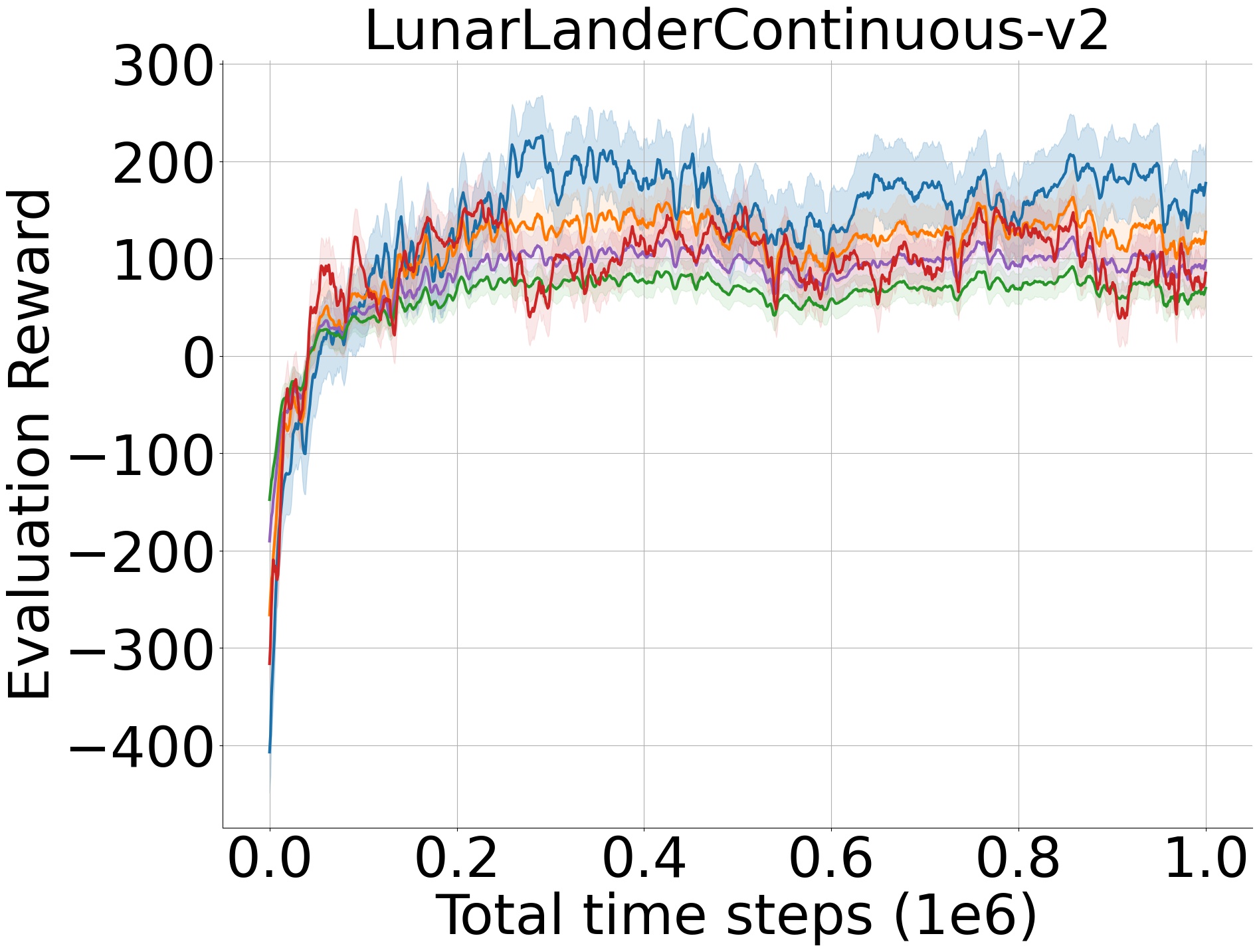}
		\includegraphics[width=2.0in, keepaspectratio]{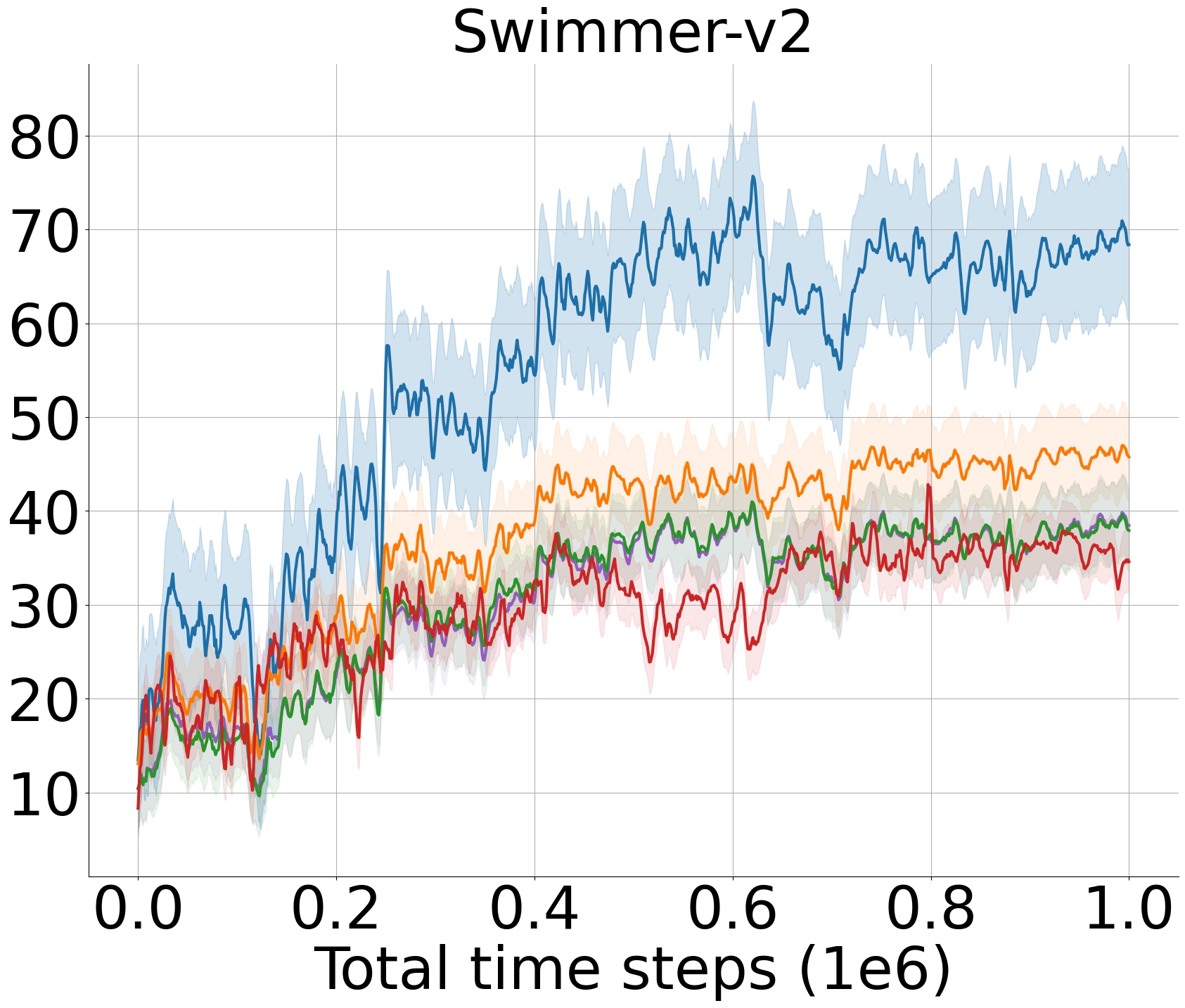}
		\includegraphics[width=2.0in, keepaspectratio]{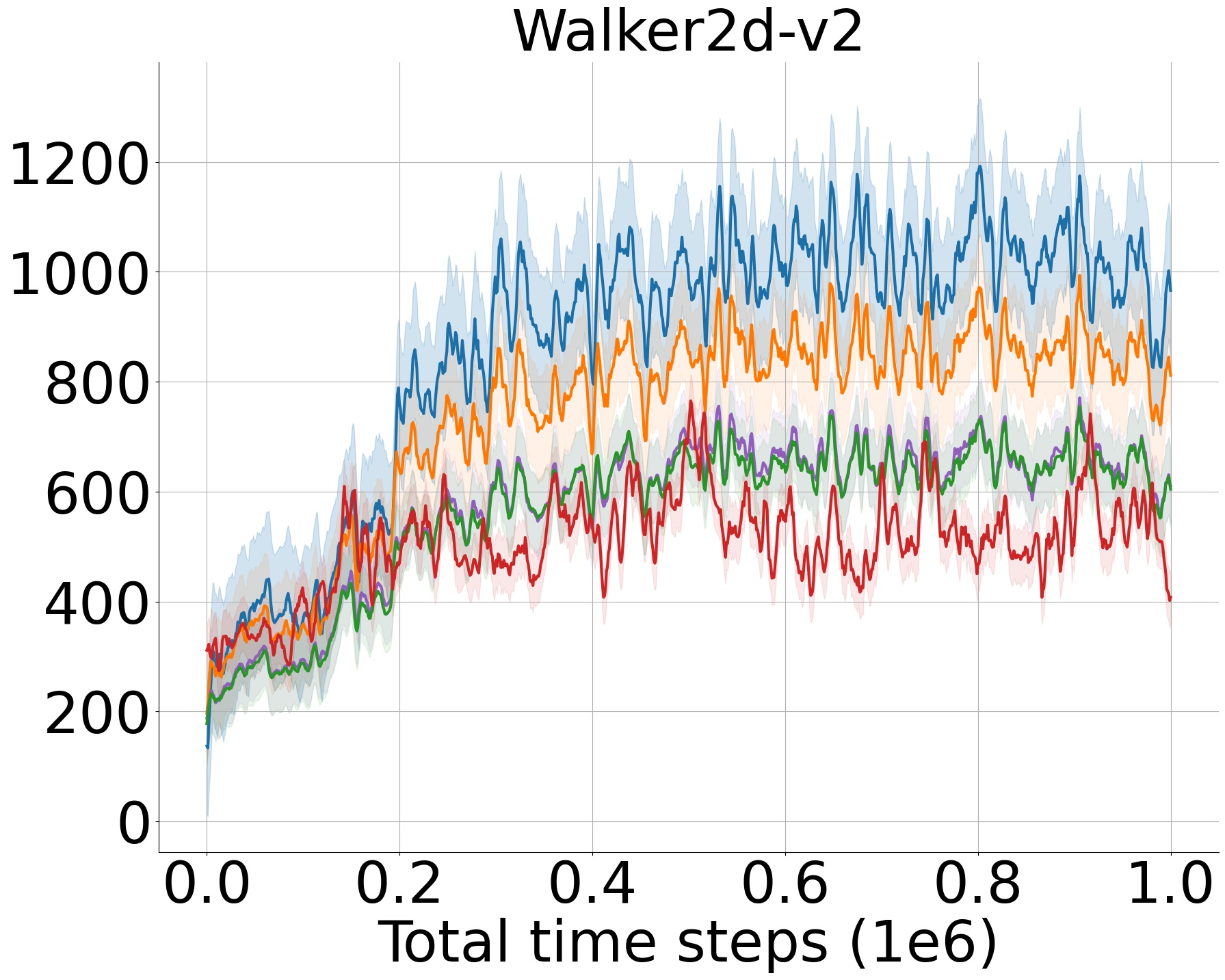}
    }
	\caption{Evaluation curves for the set of MuJoCo and Box2D continuous control tasks under the A2C algorithm. The shaded region represents half a standard deviation of the average evaluation return over 10 random seeds. A sliding window of size 5 smoothes curves for visual clarity.}
	\label{fig:eval_results_a2c}
\end{figure*}

% Evaluation Results - PPO
\begin{figure*}[!hbt]
    \centering
    \begin{align*}
        &\text{{\blue} DISCOVER-PPO ($\lambda$ = 0.1)}  &&\text{{\orange} Coherent-PPO ($\beta$ = 0.01)} &&\text{{\purple} NoisyNet-PPO} \\ &\text{{\green} PSNE-PPO} &&\text{{\red} PPO}
    \end{align*}
	\subfigure{
		\includegraphics[width=2.0in, keepaspectratio]{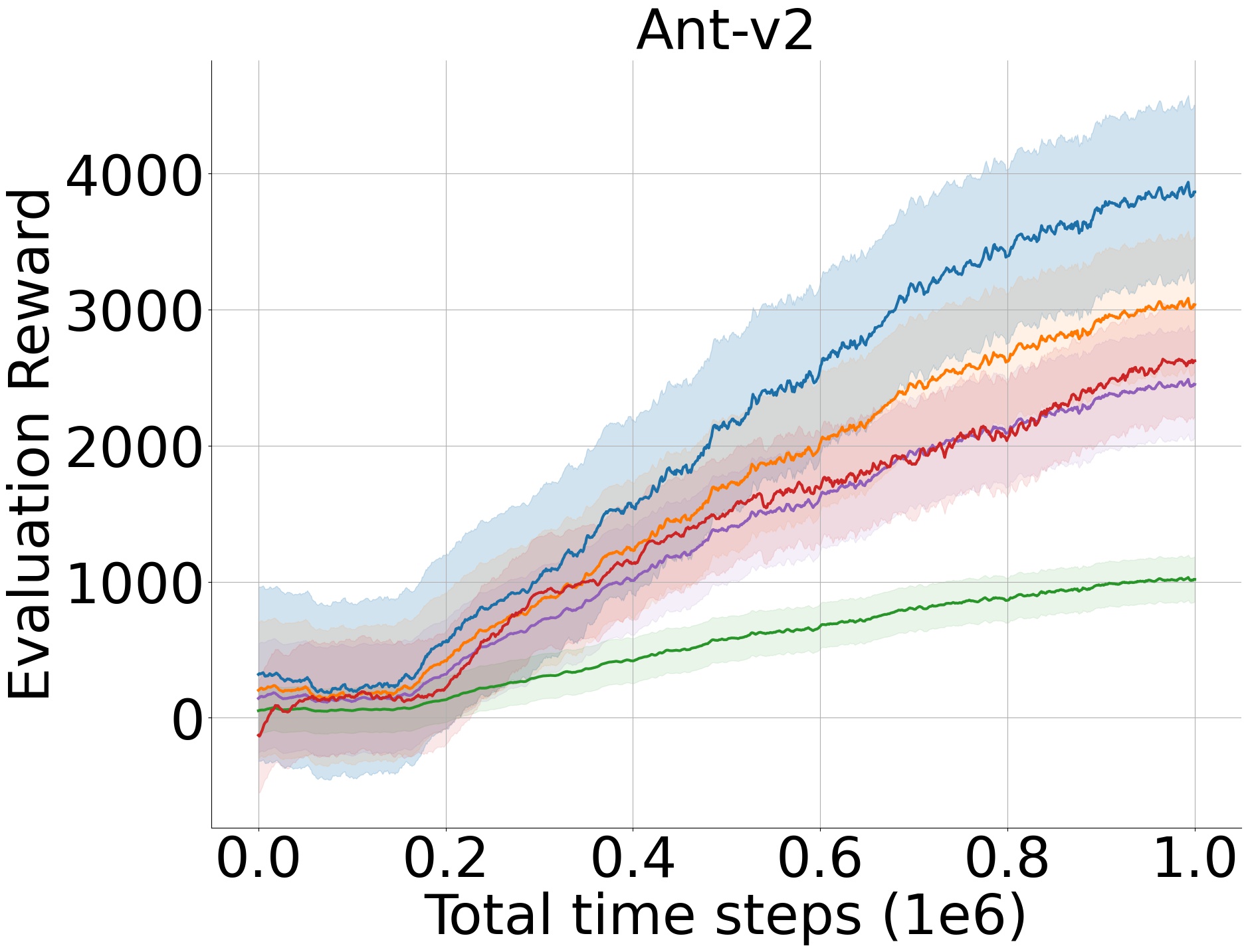}
		\includegraphics[width=2.0in, keepaspectratio]{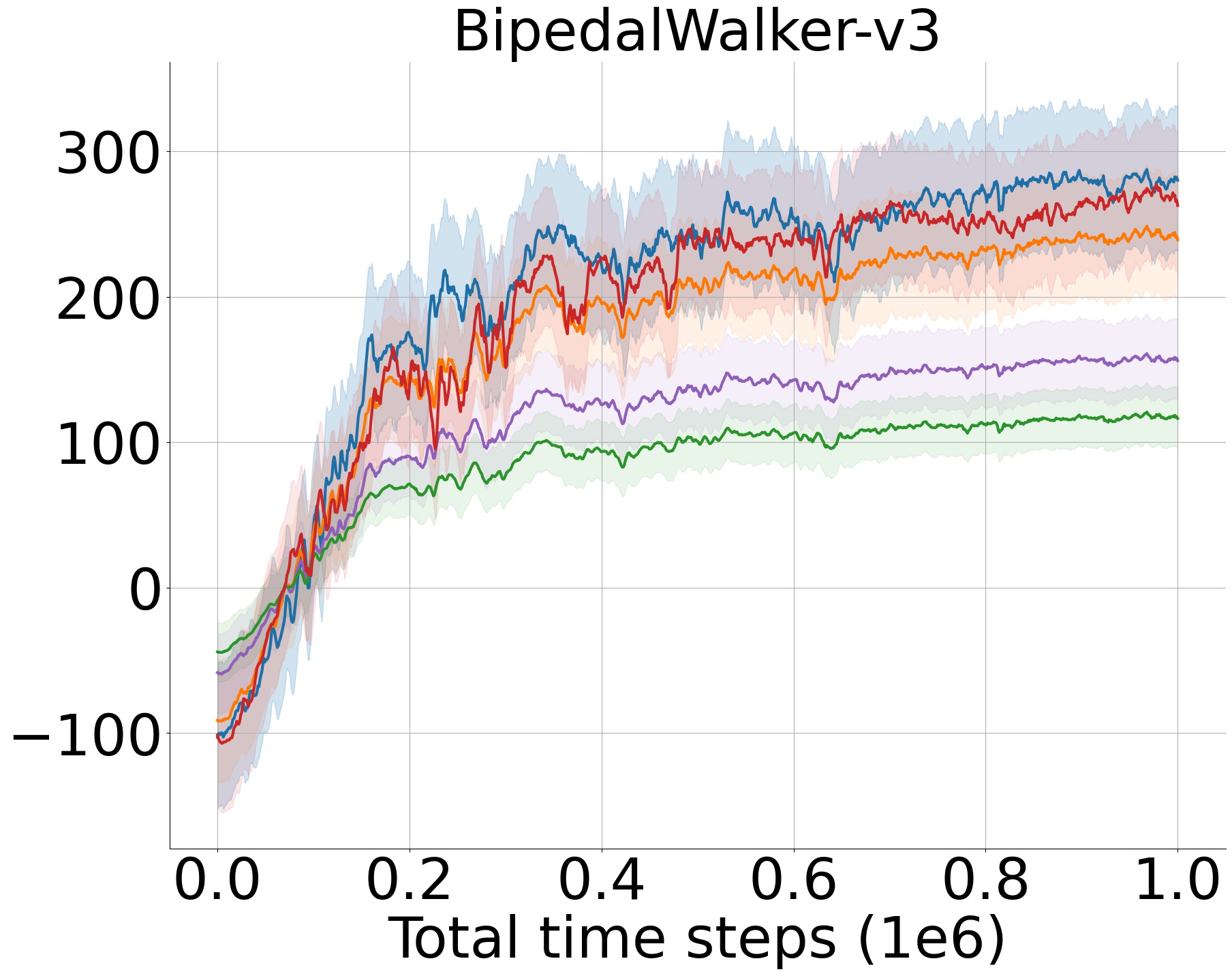}
		\includegraphics[width=2.0in, keepaspectratio]{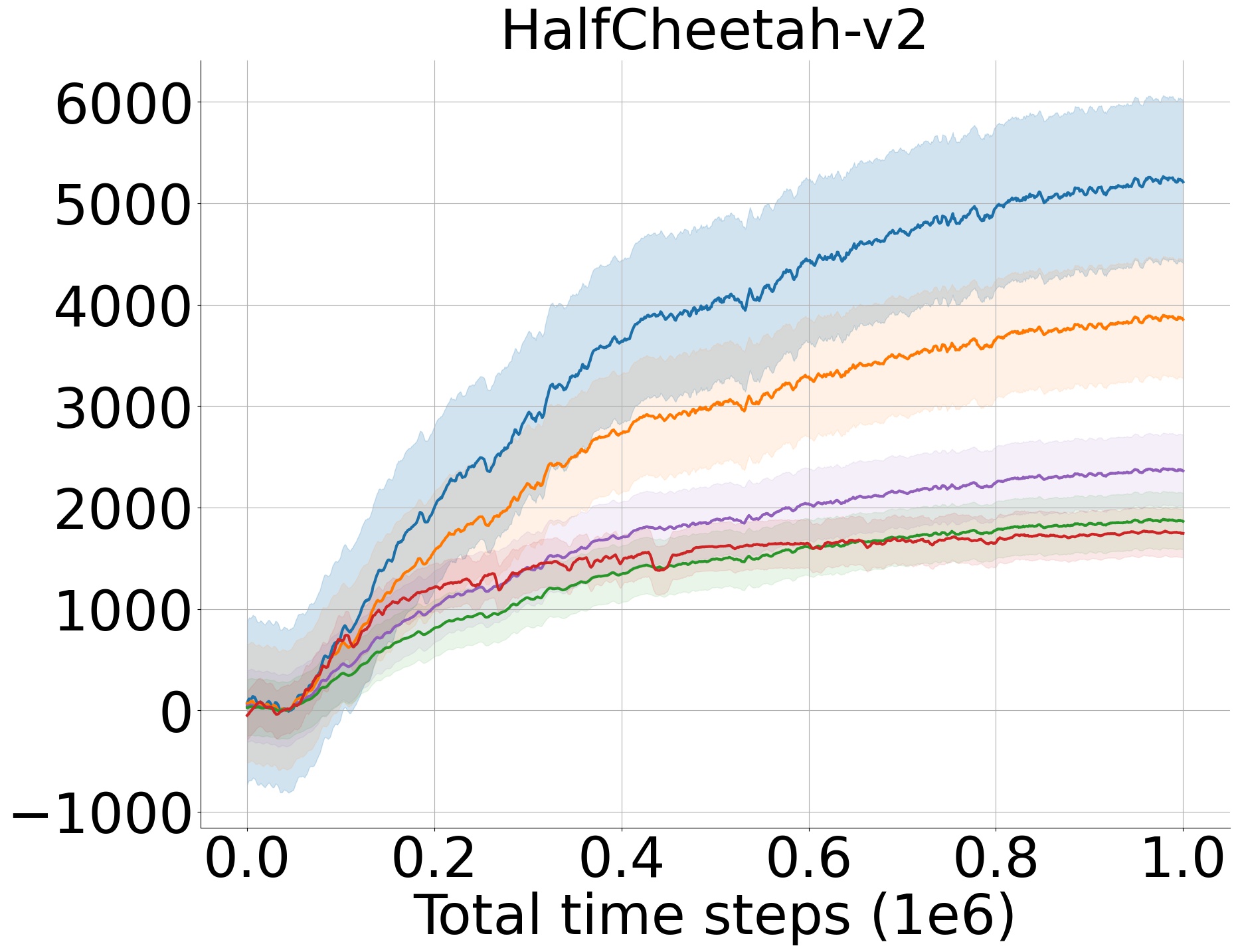}
	} \\
	\subfigure{
		\includegraphics[width=2.0in, keepaspectratio]{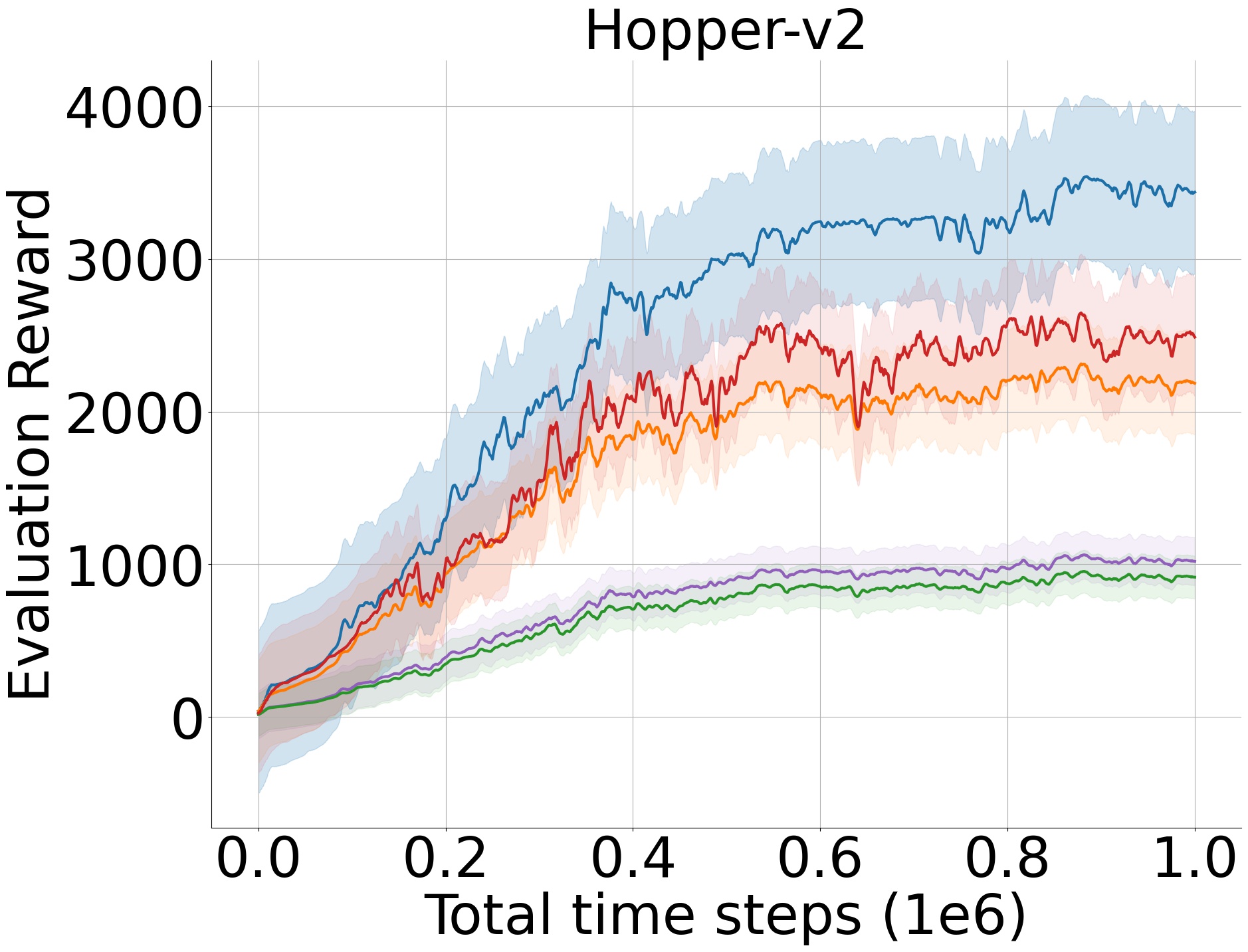}
		\includegraphics[width=2.0in, keepaspectratio]{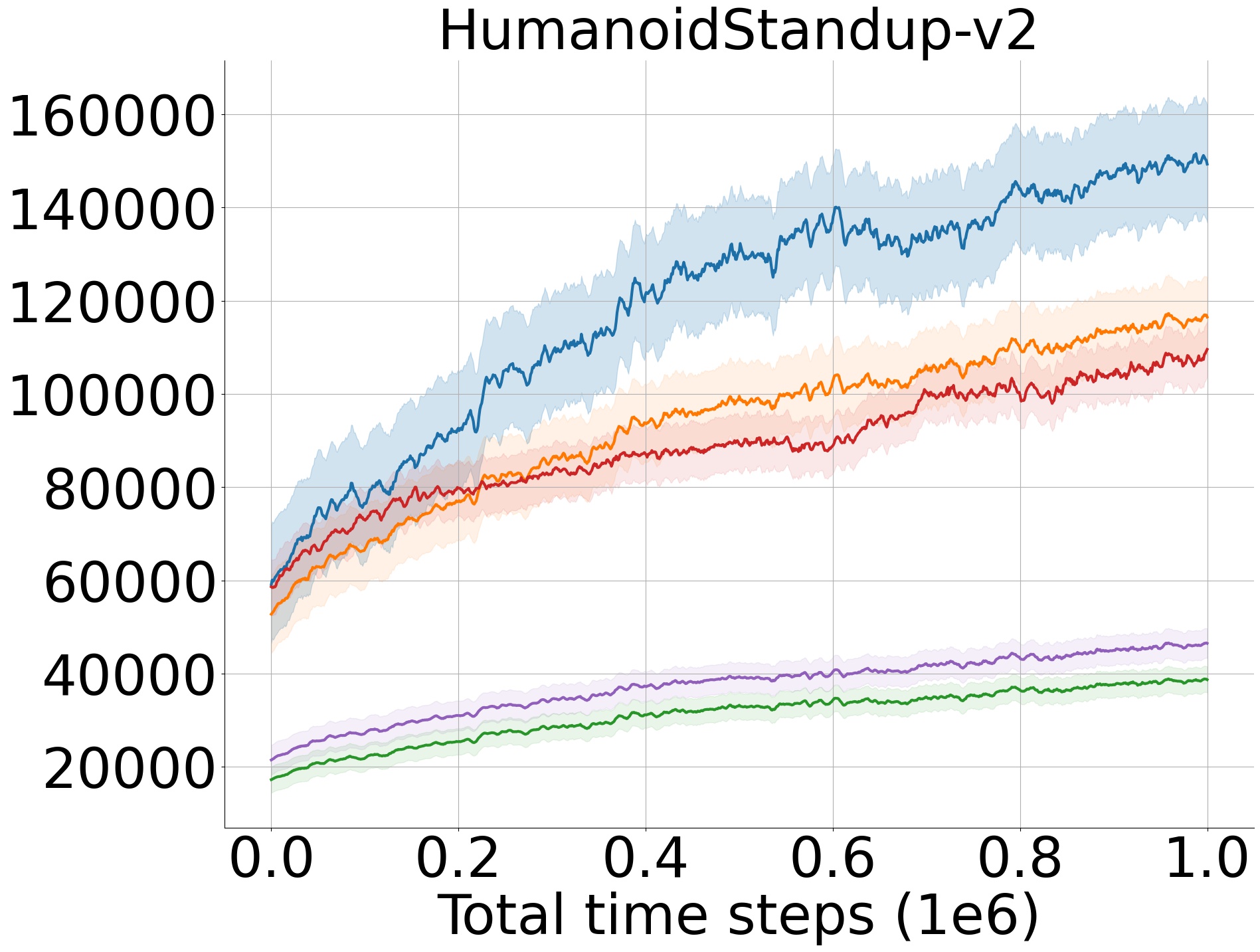}
		\includegraphics[width=2.0in, keepaspectratio]{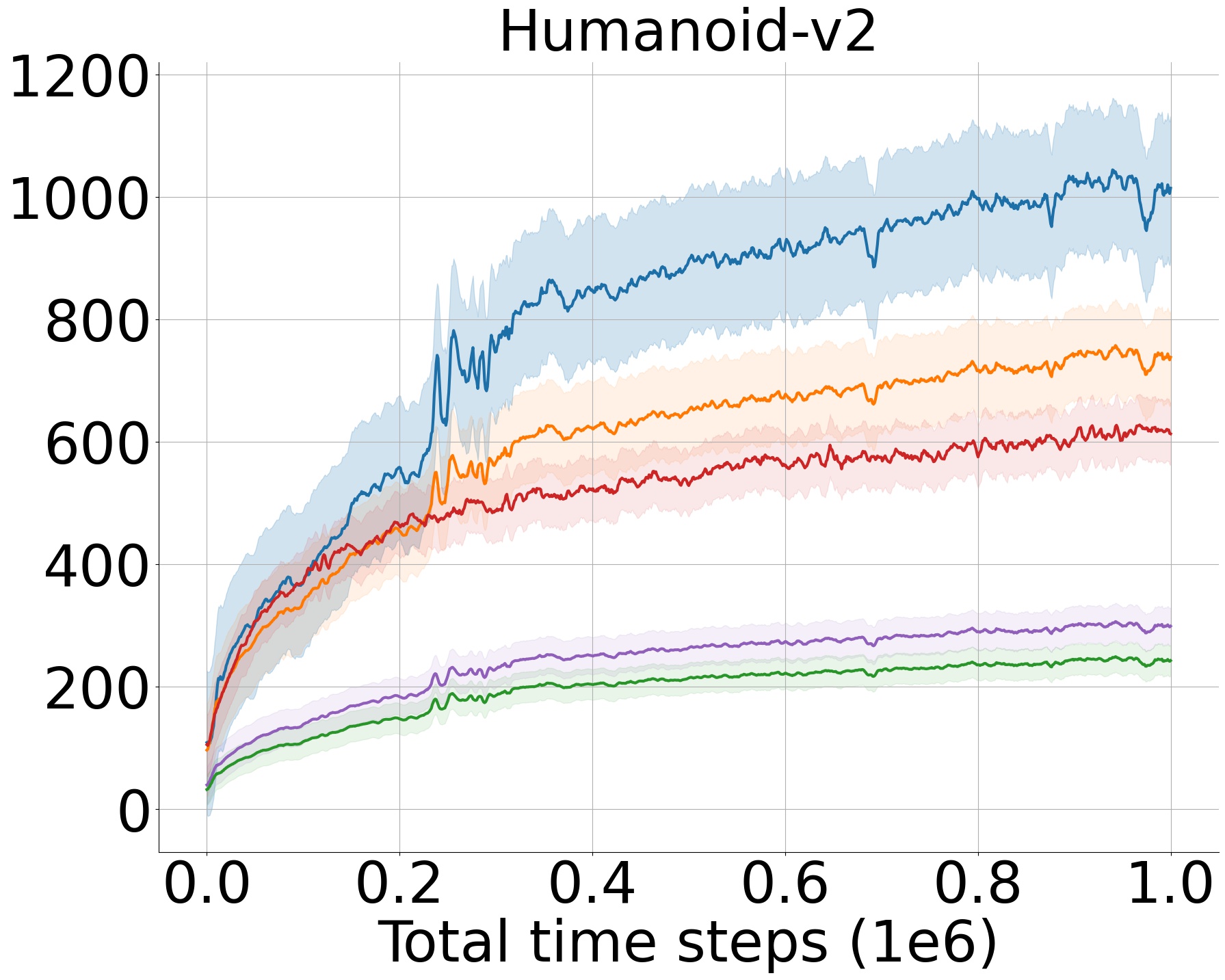}
    } \\
	\subfigure{
		\includegraphics[width=2.0in, keepaspectratio]{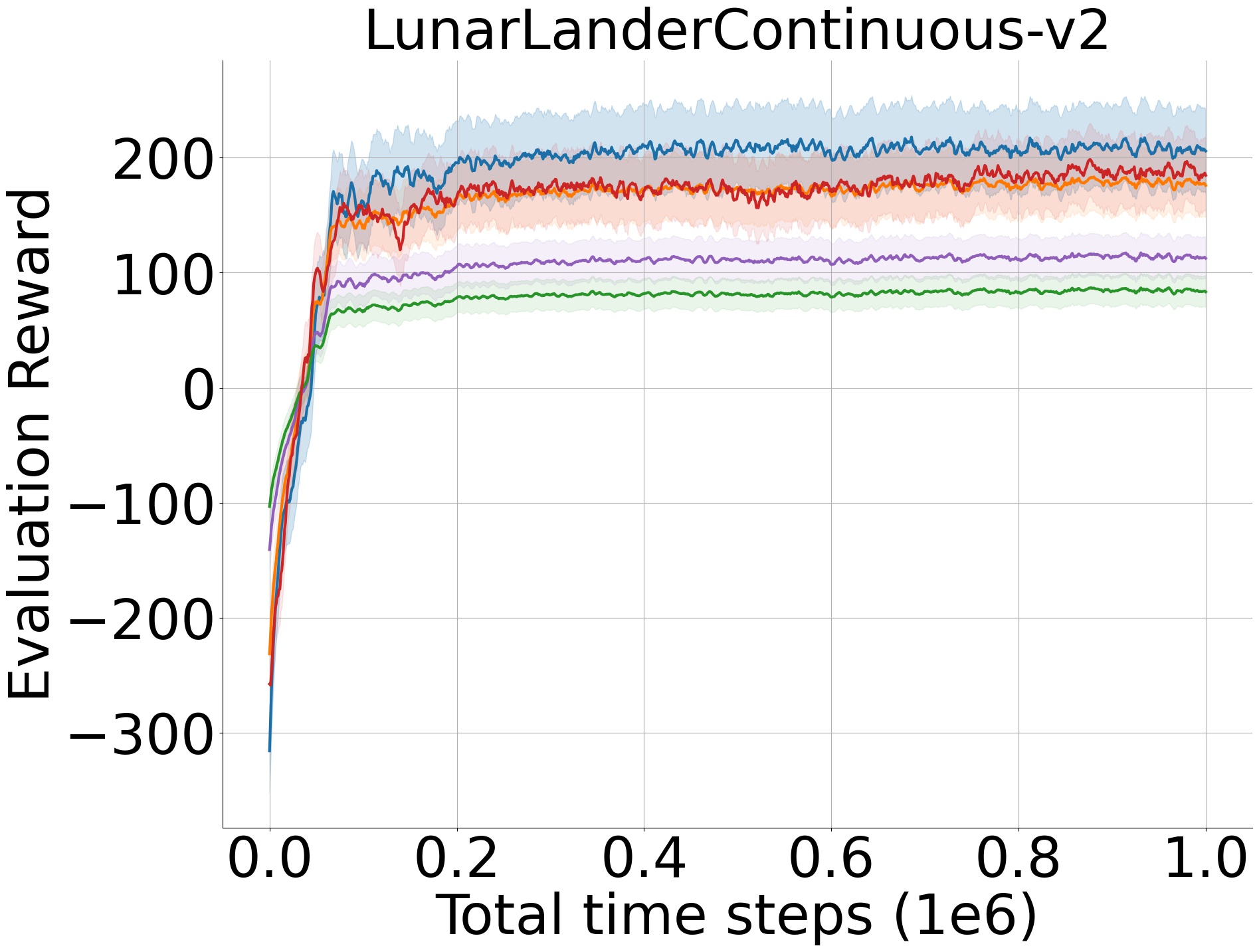}
		\includegraphics[width=2.0in, keepaspectratio]{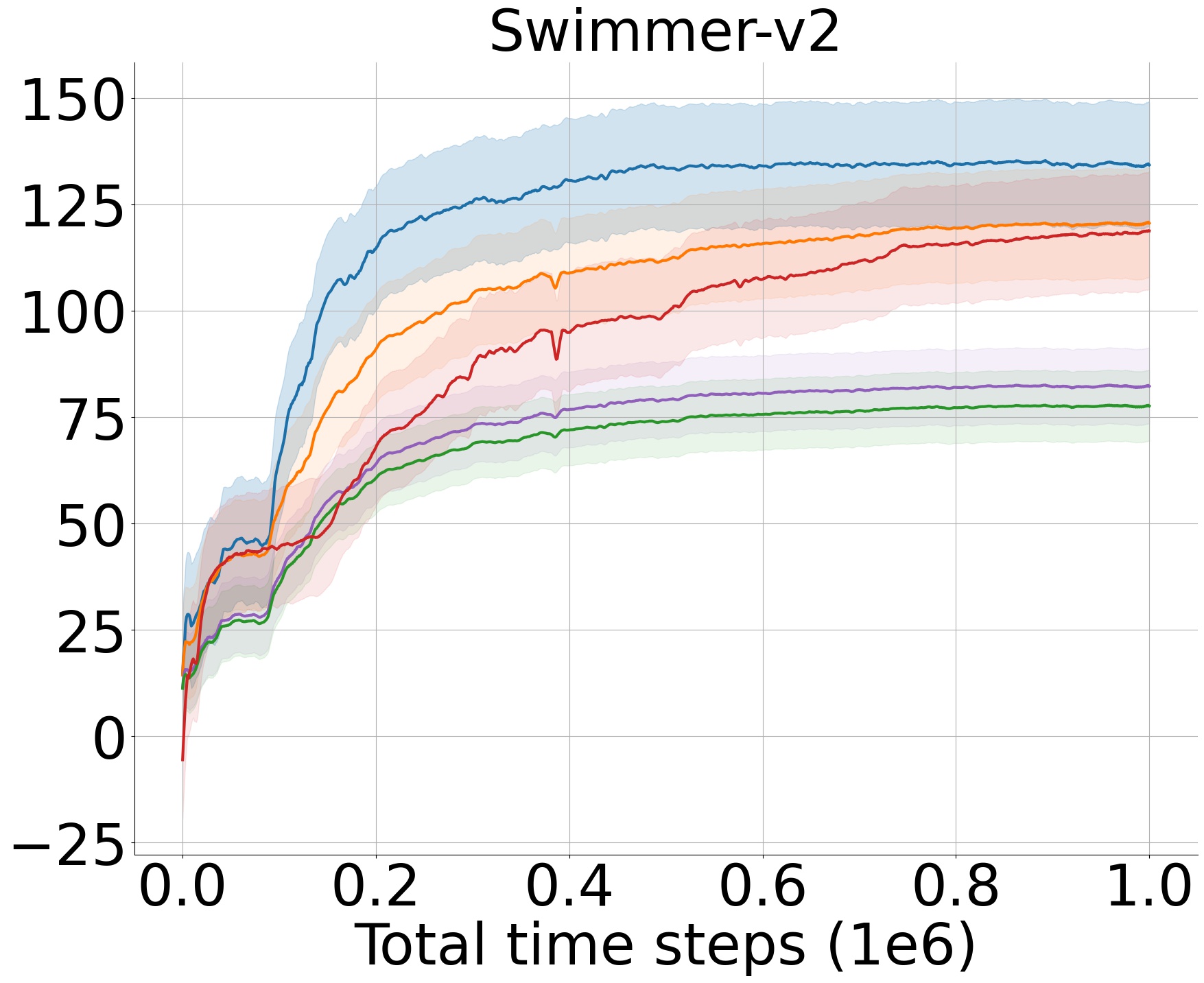}
		\includegraphics[width=2.0in, keepaspectratio]{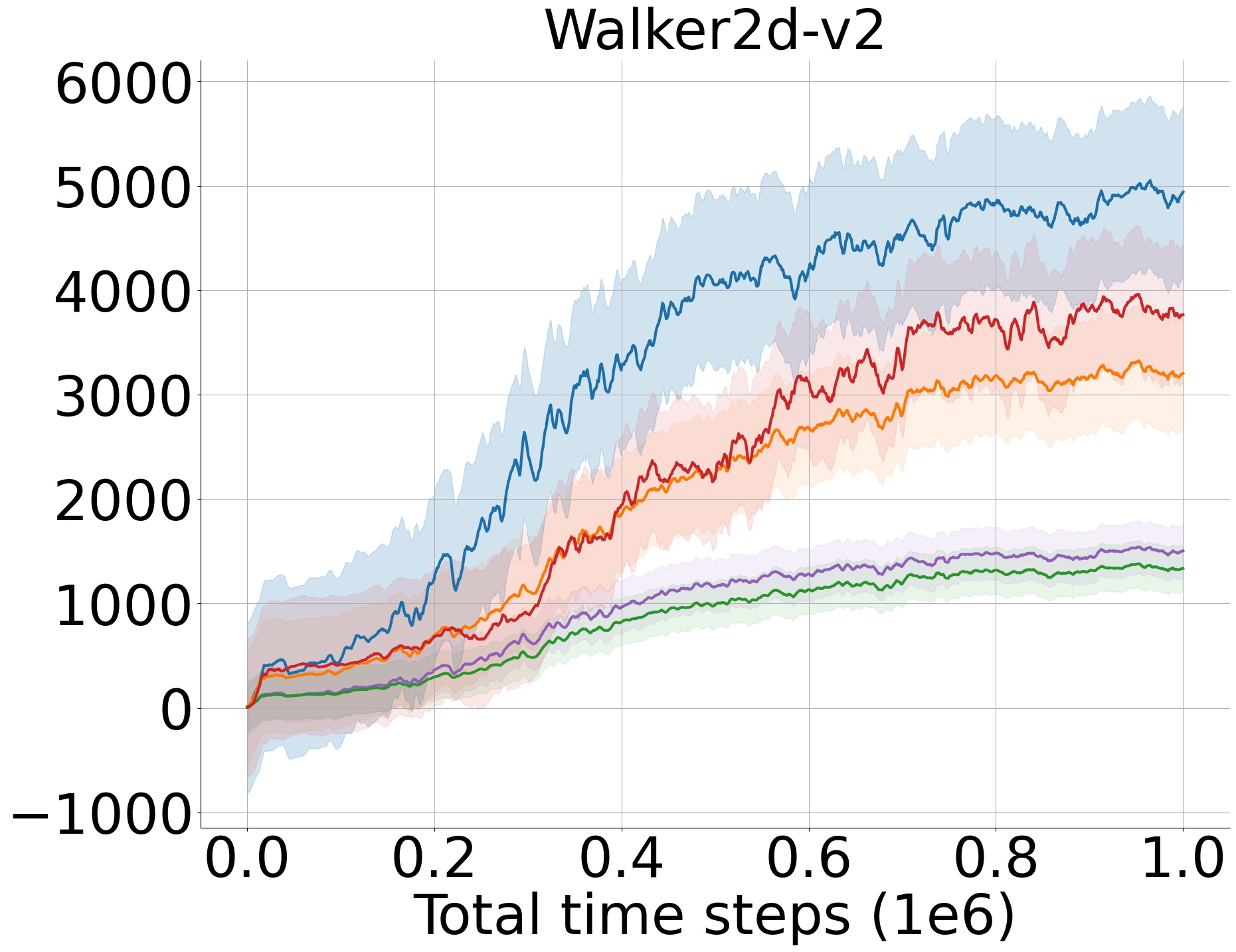}
    }
	\caption{Evaluation curves for the set of MuJoCo and Box2D continuous control tasks under the PPO algorithm. The shaded region represents half a standard deviation of the average evaluation return over 10 random seeds. A sliding window of size 5 smoothes curves for visual clarity.}
	\label{fig:eval_results_ppo}
\end{figure*}

\subsubsection{On-Policy Learning}
Learning curves under the on-policy algorithms A2C \citep{a2c} and PPO \citep{ppo} are reported in Figure \ref{fig:eval_results_a2c} and \ref{fig:eval_results_ppo}, respectively. Additional results for the InvertedDoublePendulum-v2, InvertedPendulum-v2, Reacher-v2 environments are found in Appendix \ref{app:comp_eval_results}. Note that the default exploration used in these algorithms is the inherent entropy maximization of the actions produced by the stochastic policy. As the actions are sampled from the Gaussian policy, a maximized entropy yields diverse actions and thus, more efficient exploration. From an overall perspective, we observe that On-Policy DISCOVER consistently improves the baseline algorithms and either matches or outperforms the competing approaches in all tasks regarding the convergence rate and highest evaluation returns. In some of the environments, it is demonstrated that the performance of our method matches the baseline. These environments, e.g., InvertedDoublePendulum, InvertedPendulum, and Reacher, are regarded as trivial in that they do not require a comprehensive exploration or complex learning approaches \citep{deep_rl_that_matters}. However, we observe a crucial performance improvement in more challenging tasks and high dimensional environments such as Ant, HalfCheetah, Hopper, Humanoid, Swimmer, and Walker2d. Therefore, we infer that when the underlying algorithm fails at learning, the effectiveness of the utilized exploration strategy becomes prominent. 

For the competing approaches, although our implementation of NoisyNet \citep{noisy_net} and PSNE \citep{psne} follow the authors' implementation and tuned hyper-parameters, we notice that they usually underperform the baseline, while Deep Coherent Exploration \citep{deep_coherent_exploration} exhibits a considerable performance. It was previously shown by \citet{deep_coherent_exploration} that NoisyNet \citep{noisy_net} and PSNE \citep{psne} fail in the continuous setting as they were initially proposed for the standard Atari games \citep{atari_ale} in which action spaces are discrete, as discussed in Section \ref{sec:related_work_exploration}. Furthermore, although the Coherent exploration \citep{deep_coherent_exploration} was shown to exhibit better performance in the on-policy setting due to the synchronized policy and exploration strategy updates, our approach outperforms it by a notable margin. Finally, On-Policy DISCOVER exhibits an improved but suboptimal behavior in environments when the baseline does not converge, such as A2C \citep{a2c} in Reacher, InvertedPendulum, InvertedDoublePendulumand, and PPO \citep{ppo} in HalfCheetah, Humanoid. Naturally, when the off-policy method cannot solve the environment sufficiently, the performance improvement offered by DISCOVER does not cover optimal policies.

\subsubsection{Off-Policy Learning}
\begin{figure*}[!hbt]
    \centering
    \begin{align*}
        &\text{{\blue} DISCOVER-DDPG ($\lambda$ = 0.3)}  &&\text{{\orange} Coherent-DDPG ($\beta$ = 0.01)} &&\text{{\purple} NoisyNet-DDPG} \\ &\text{{\green} PSNE-DDPG} &&\text{{\red} Gaussian-DDPG} &&\text{{\brown} Greedy-DDPG}
    \end{align*}
	\subfigure{
		\includegraphics[width=2.0in, keepaspectratio]{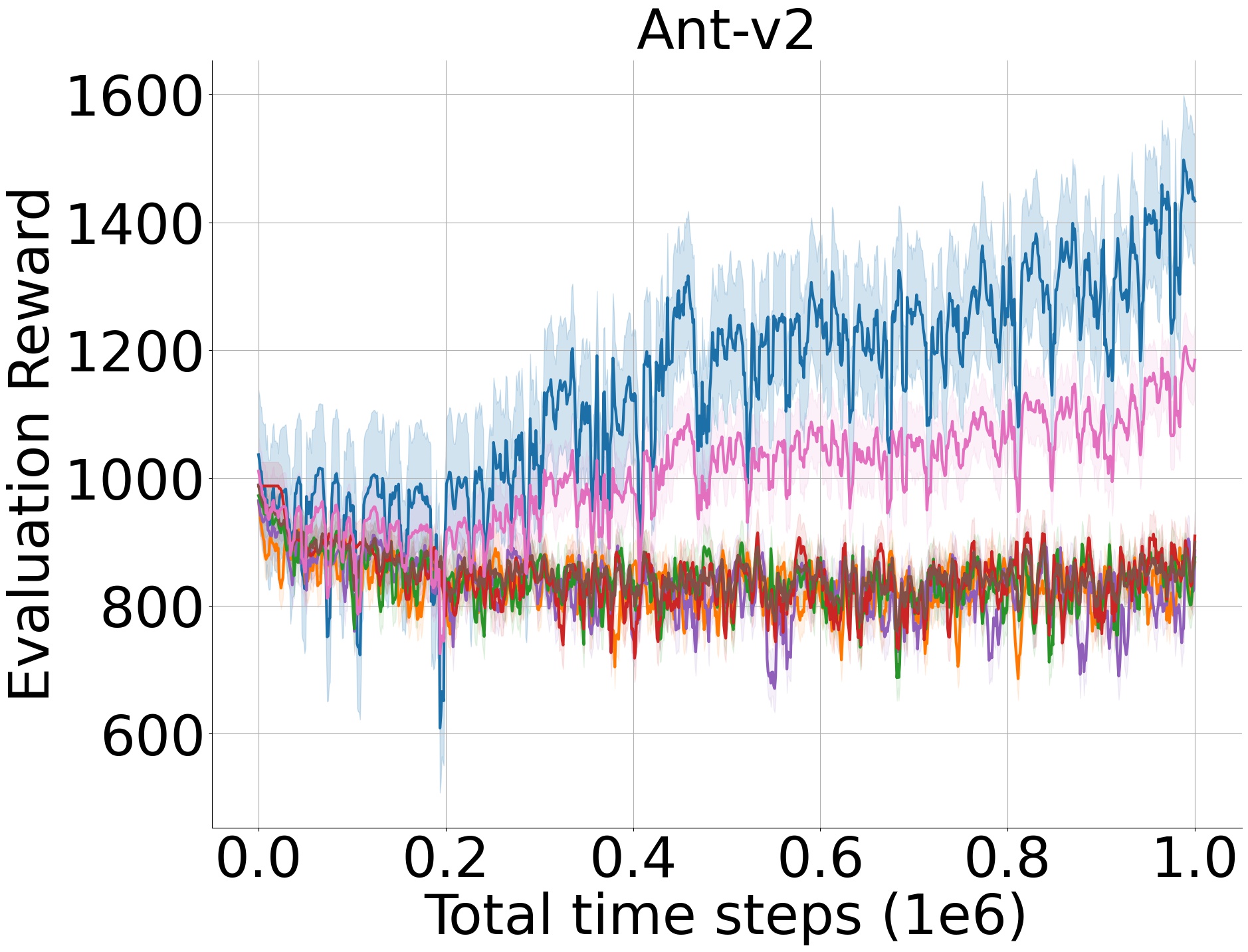}
		\includegraphics[width=2.0in, keepaspectratio]{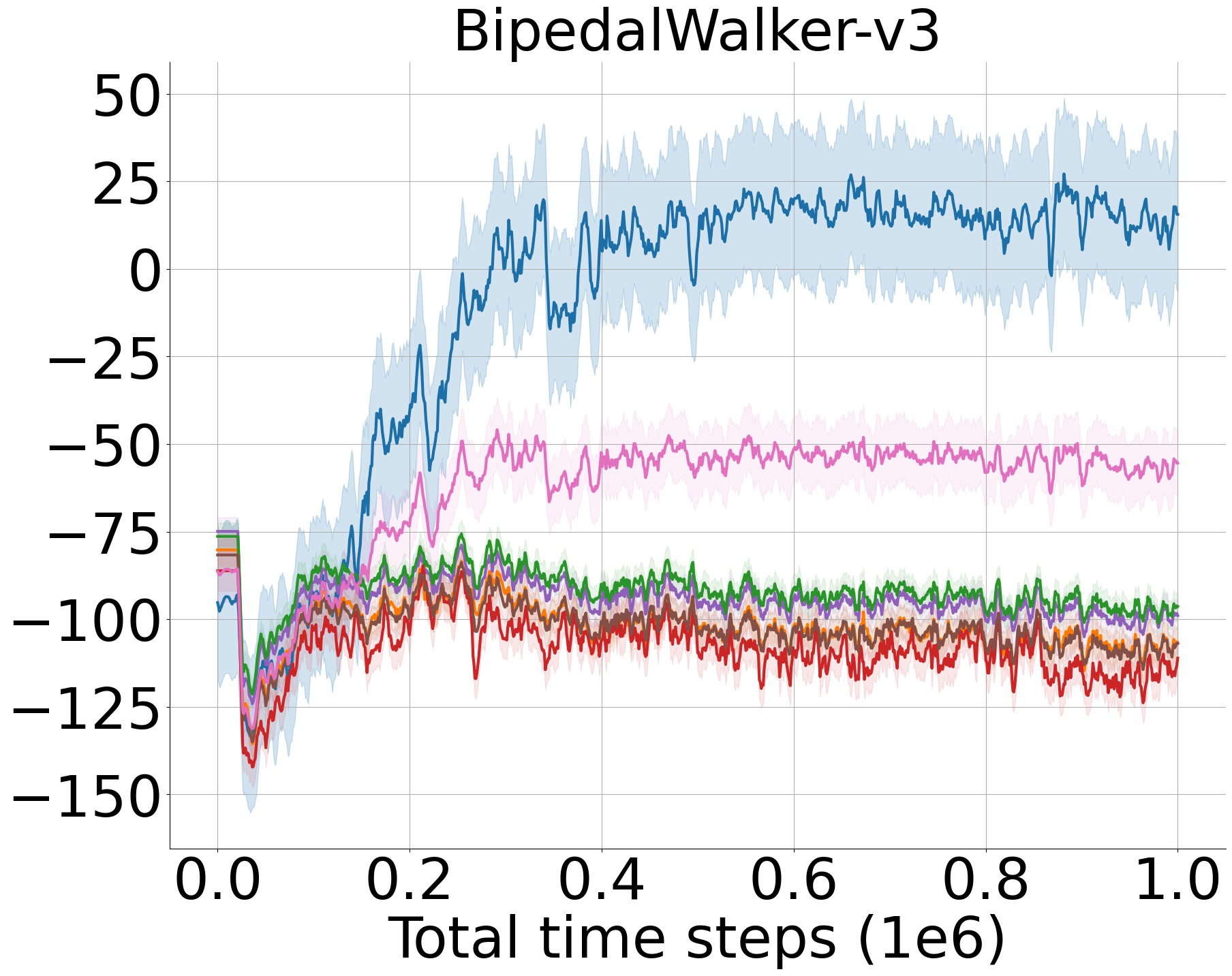}
		\includegraphics[width=2.0in, keepaspectratio]{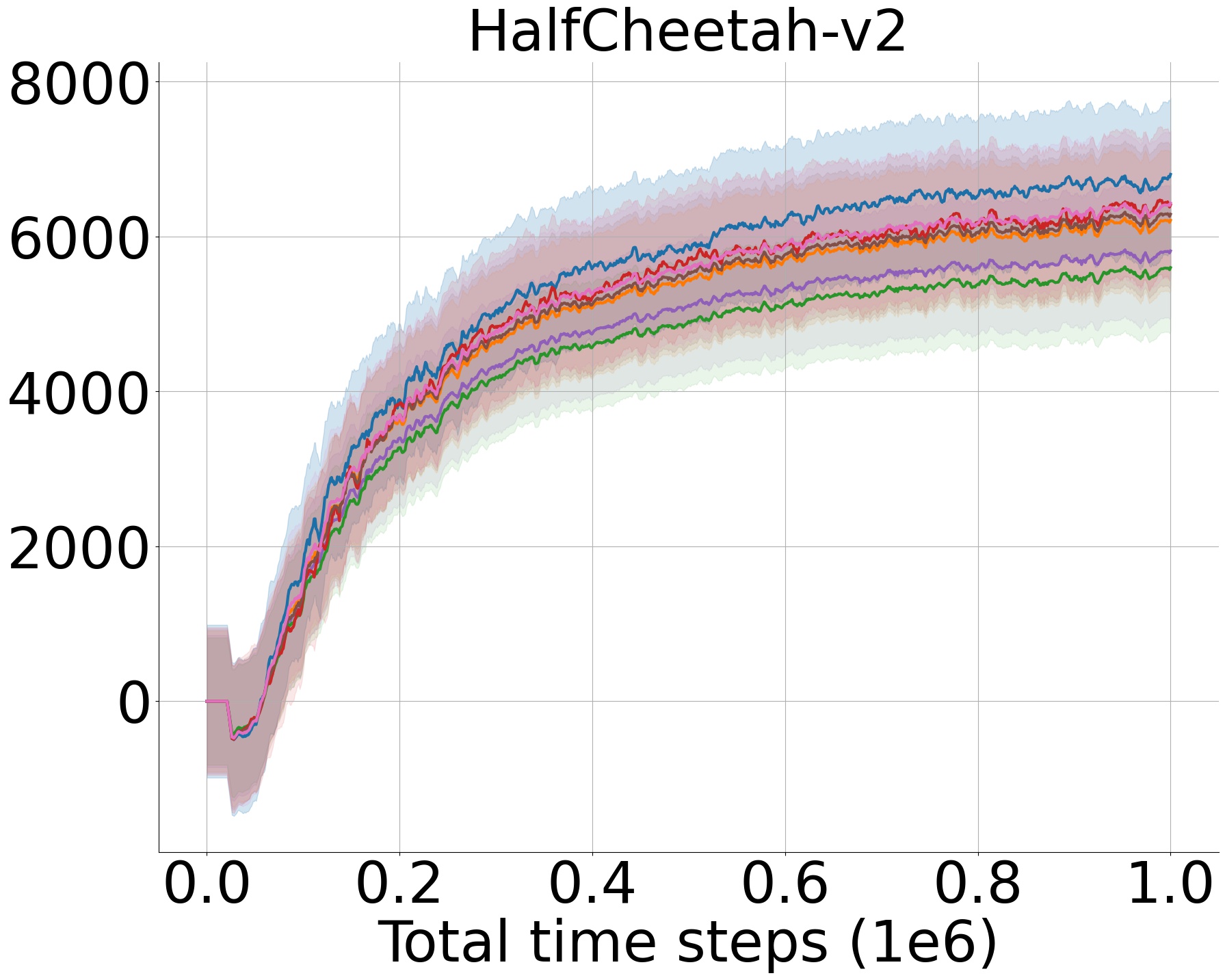}
	} \\
	\subfigure{
		\includegraphics[width=2.0in, keepaspectratio]{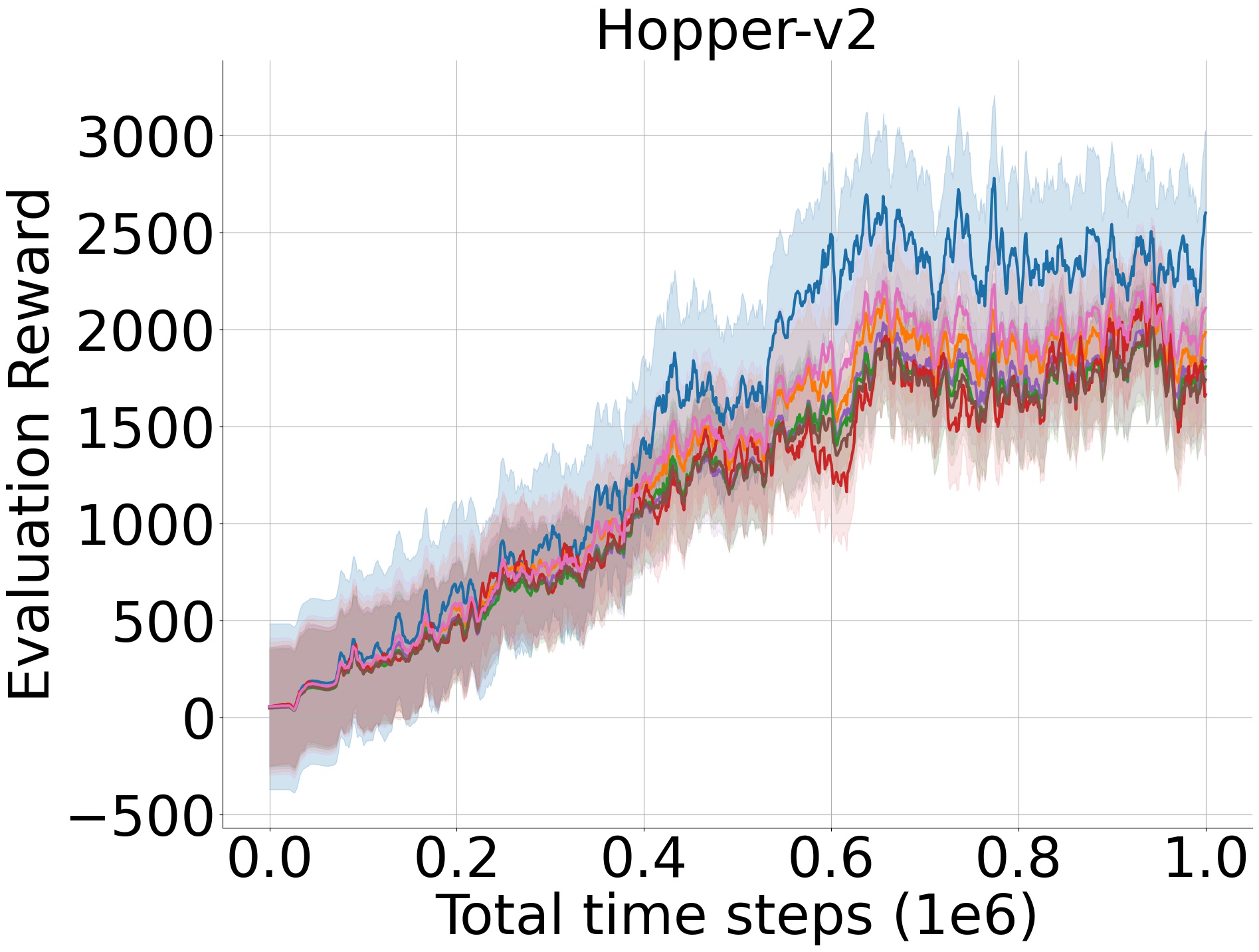}
		\includegraphics[width=2.0in, keepaspectratio]{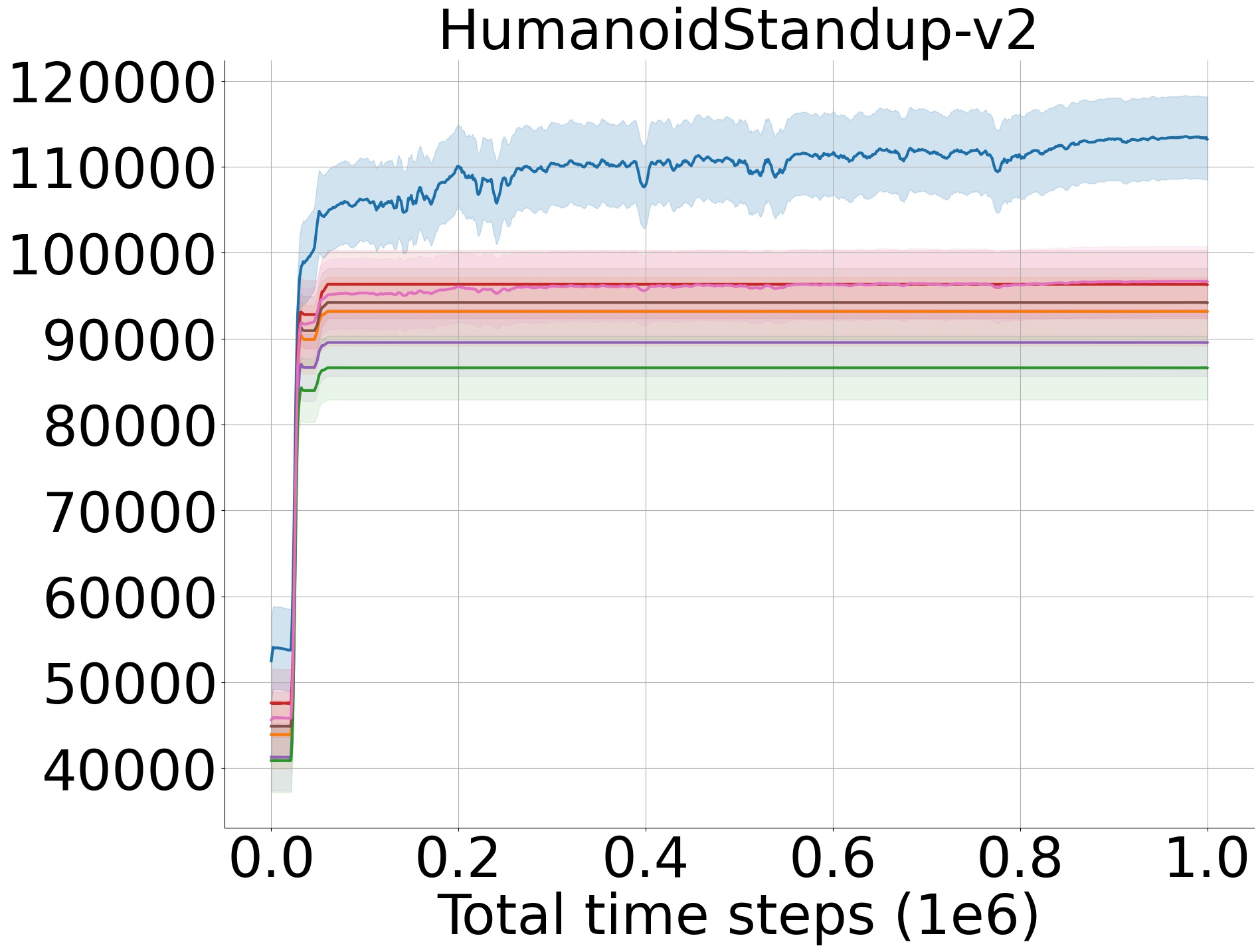}
		\includegraphics[width=2.0in, keepaspectratio]{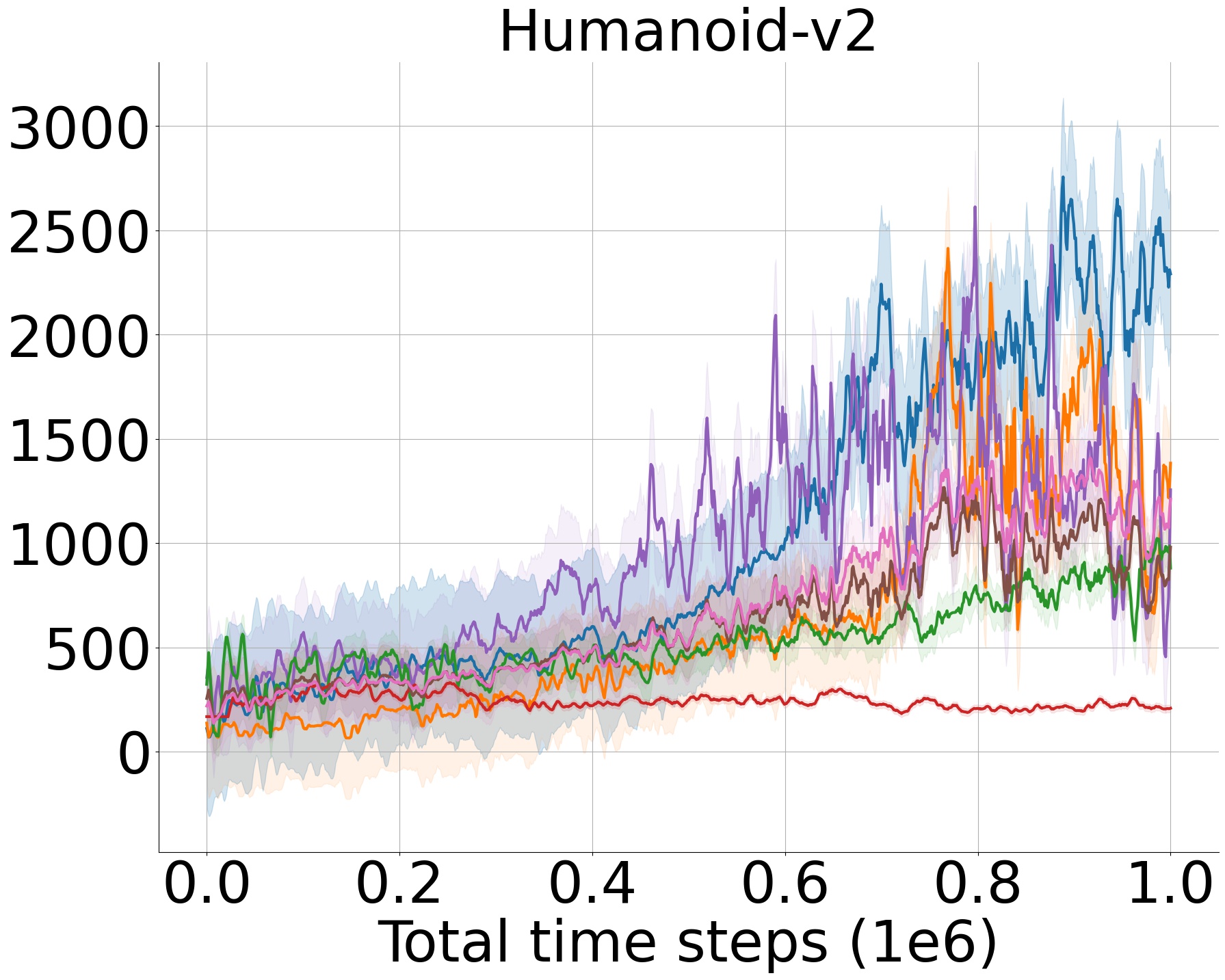}
    } \\
	\subfigure{
		\includegraphics[width=2.0in, keepaspectratio]{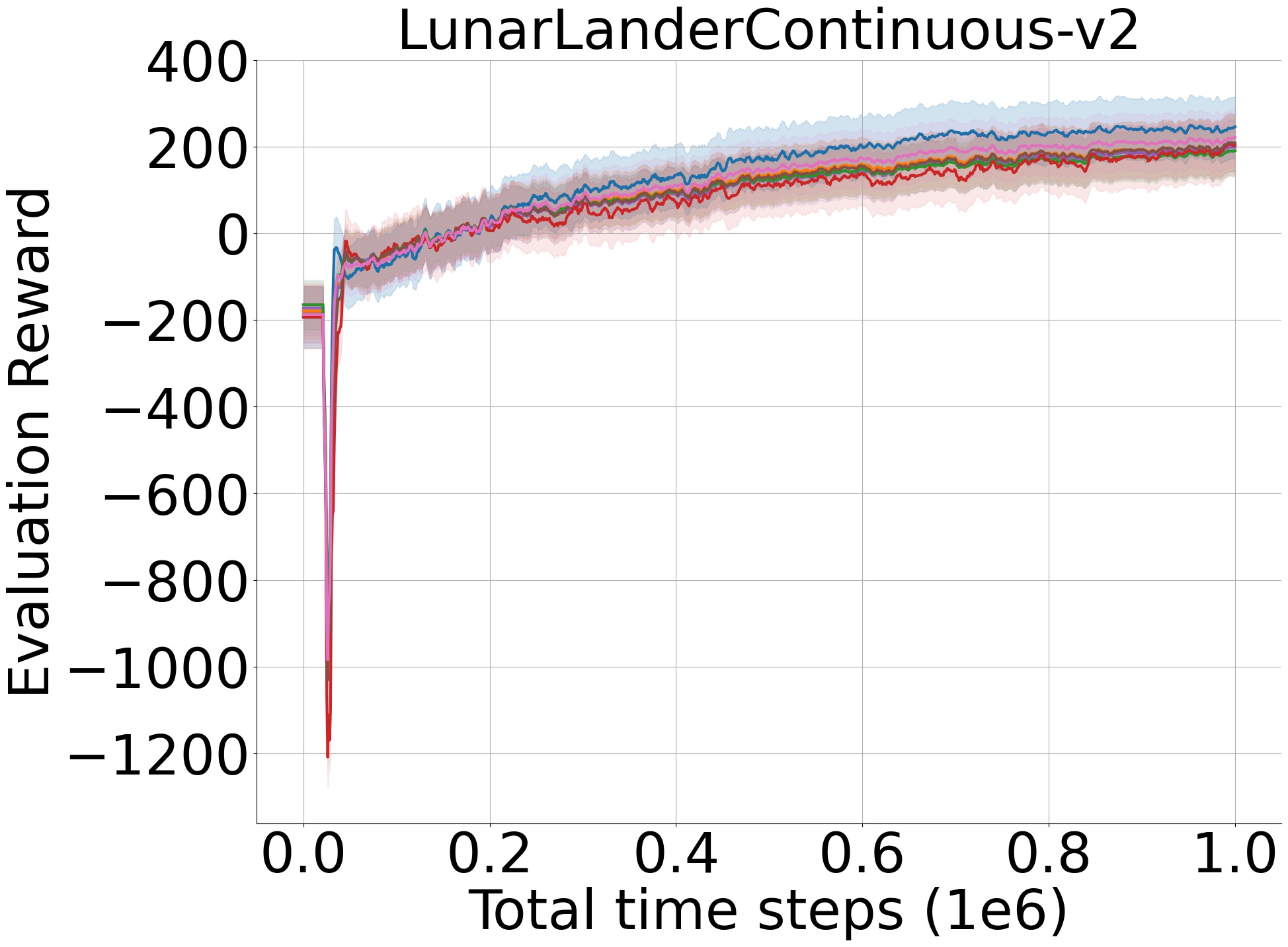}
		\includegraphics[width=2.0in, keepaspectratio]{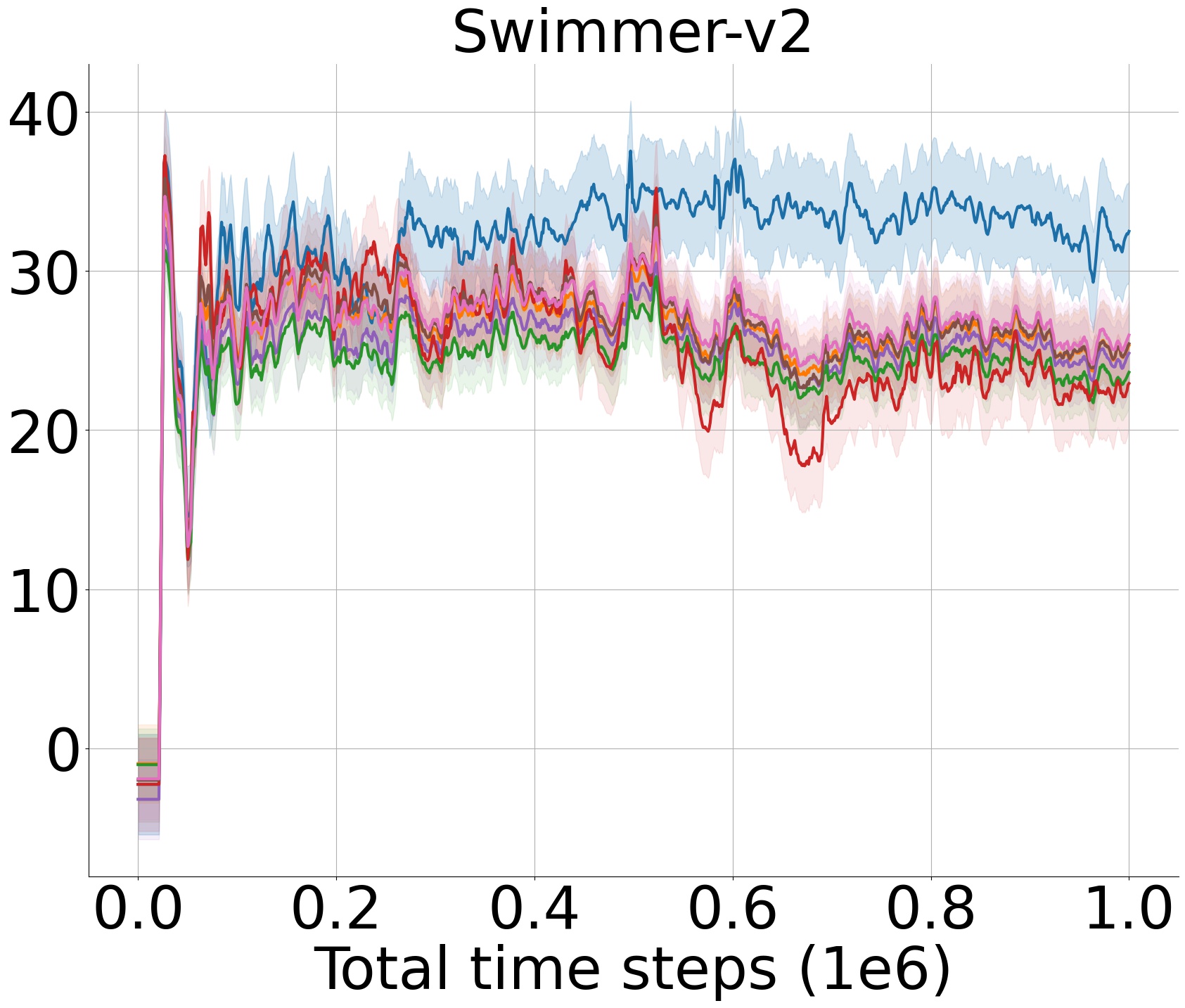}
		\includegraphics[width=2.0in, keepaspectratio]{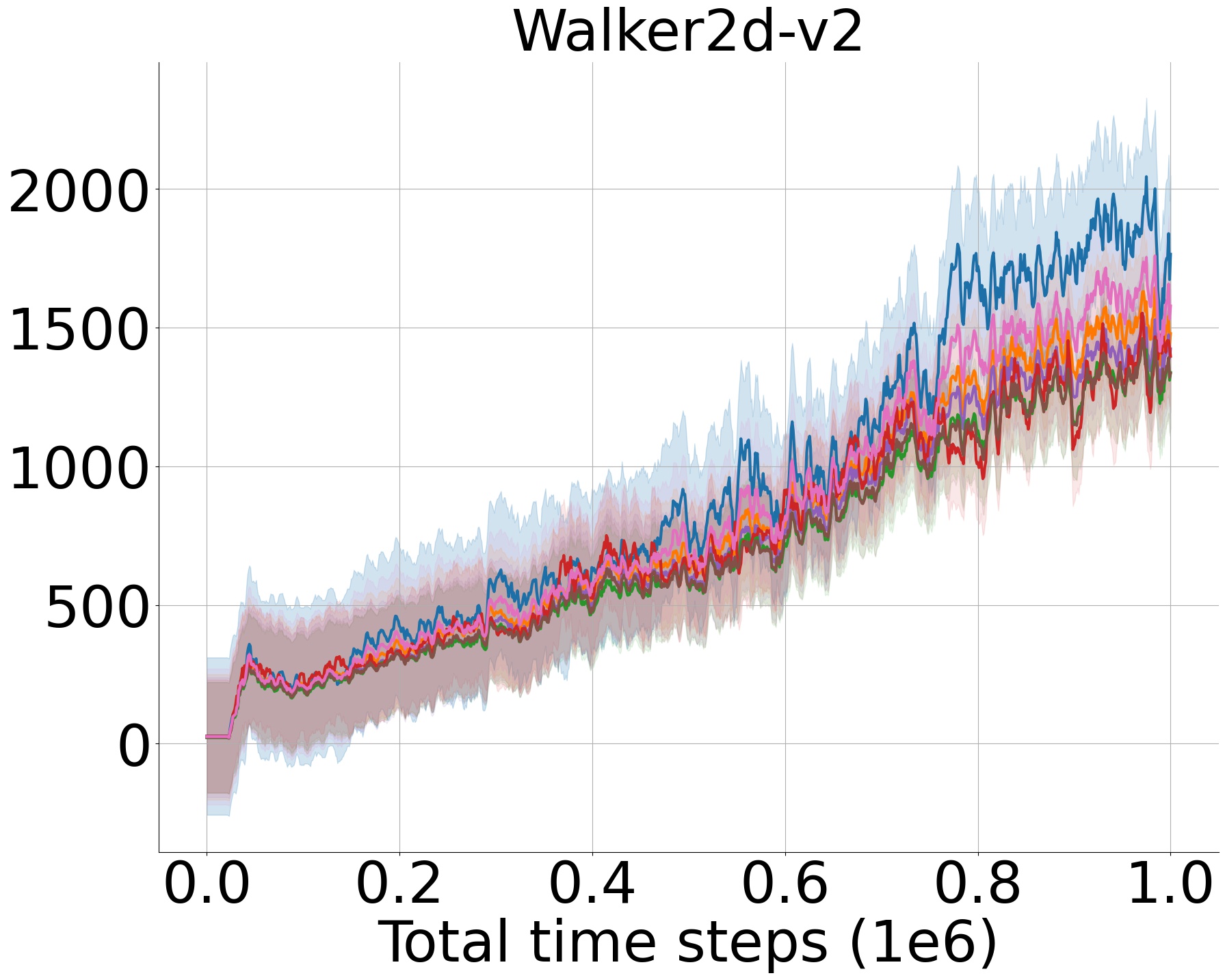}
    }
	\caption{Evaluation curves for the set of MuJoCo and Box2D continuous control tasks under the DDPG algorithm. The shaded region represents half a standard deviation of the average evaluation return over 10 random seeds. A sliding window of size 5 smoothes curves for visual clarity.}
	\label{fig:eval_results_ddpg}
\end{figure*}

\begin{figure*}[!hbt]
    \centering
    \begin{align*}
        &\text{{\blue} DISCOVER-SAC ($\lambda$ = 0.3)}  &&\text{{\orange} Coherent-SAC ($\beta$ = 0.01)} &&\text{{\purple} NoisyNet-SAC} \\ &\text{{\green} PSNE-SAC} &&\text{{\red} Gaussian-SAC} &&\text{{\brown} Greedy-SAC}
    \end{align*}
	\subfigure{
		\includegraphics[width=2.0in, keepaspectratio]{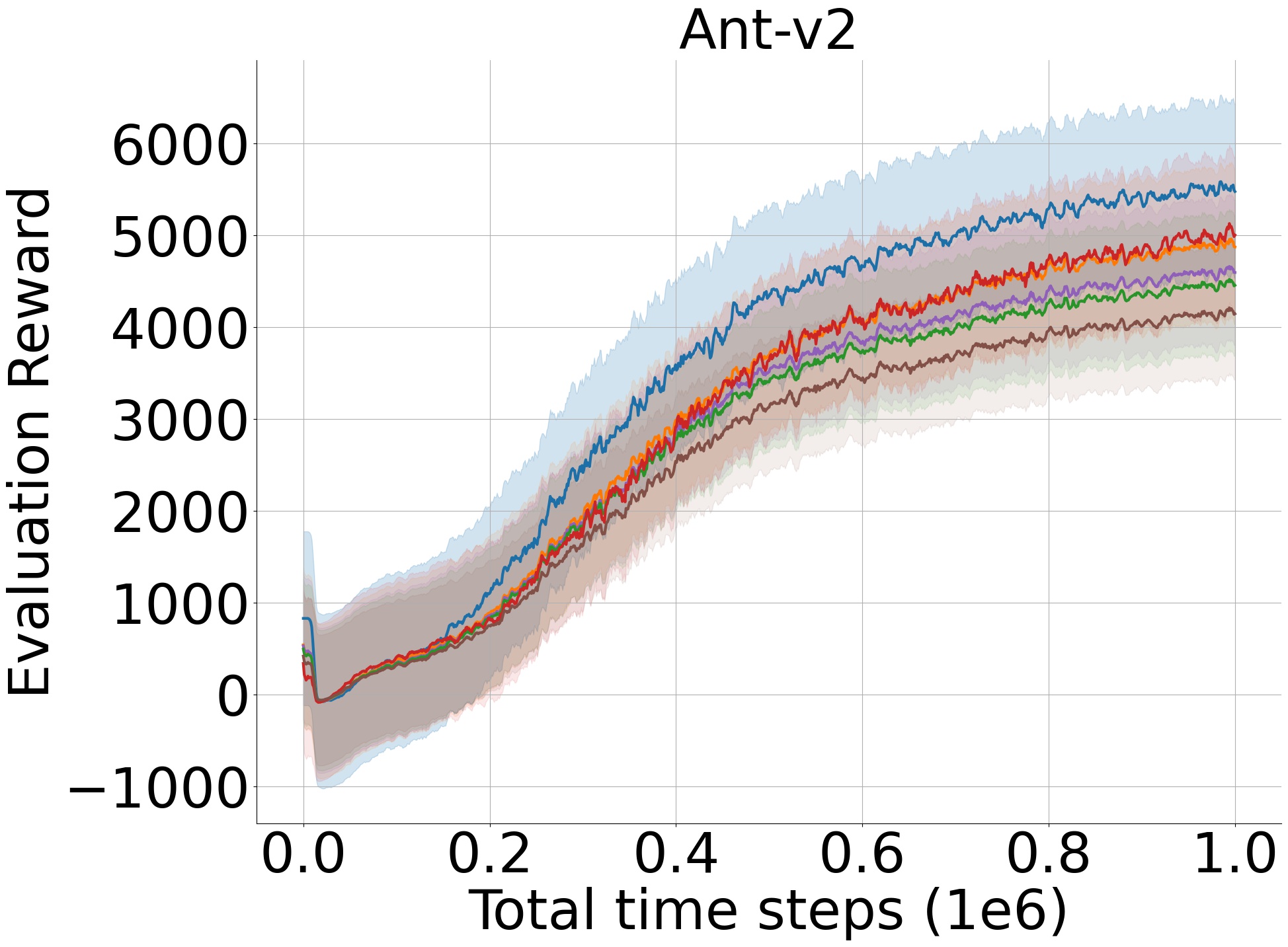}
		\includegraphics[width=2.0in, keepaspectratio]{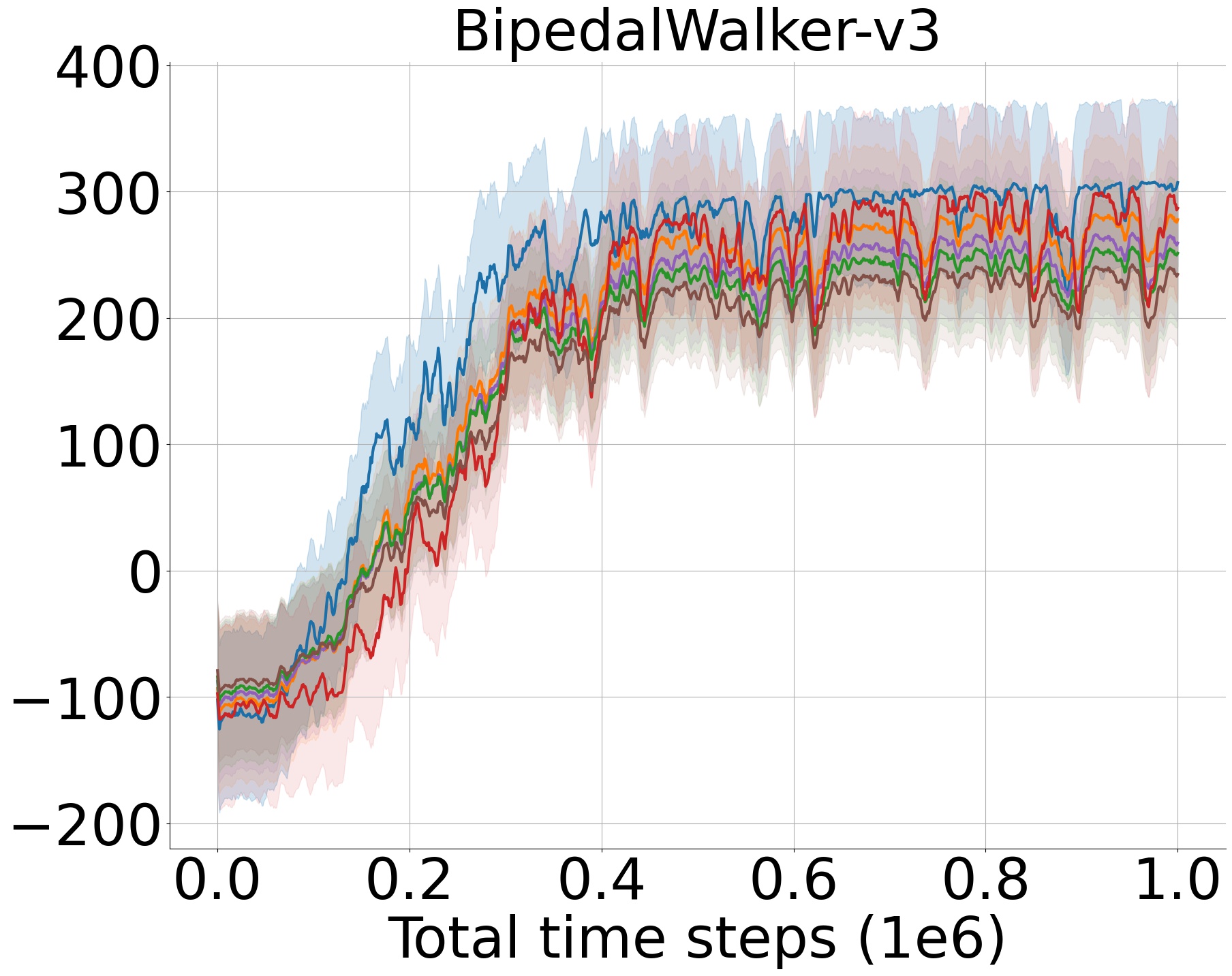}
		\includegraphics[width=2.0in, keepaspectratio]{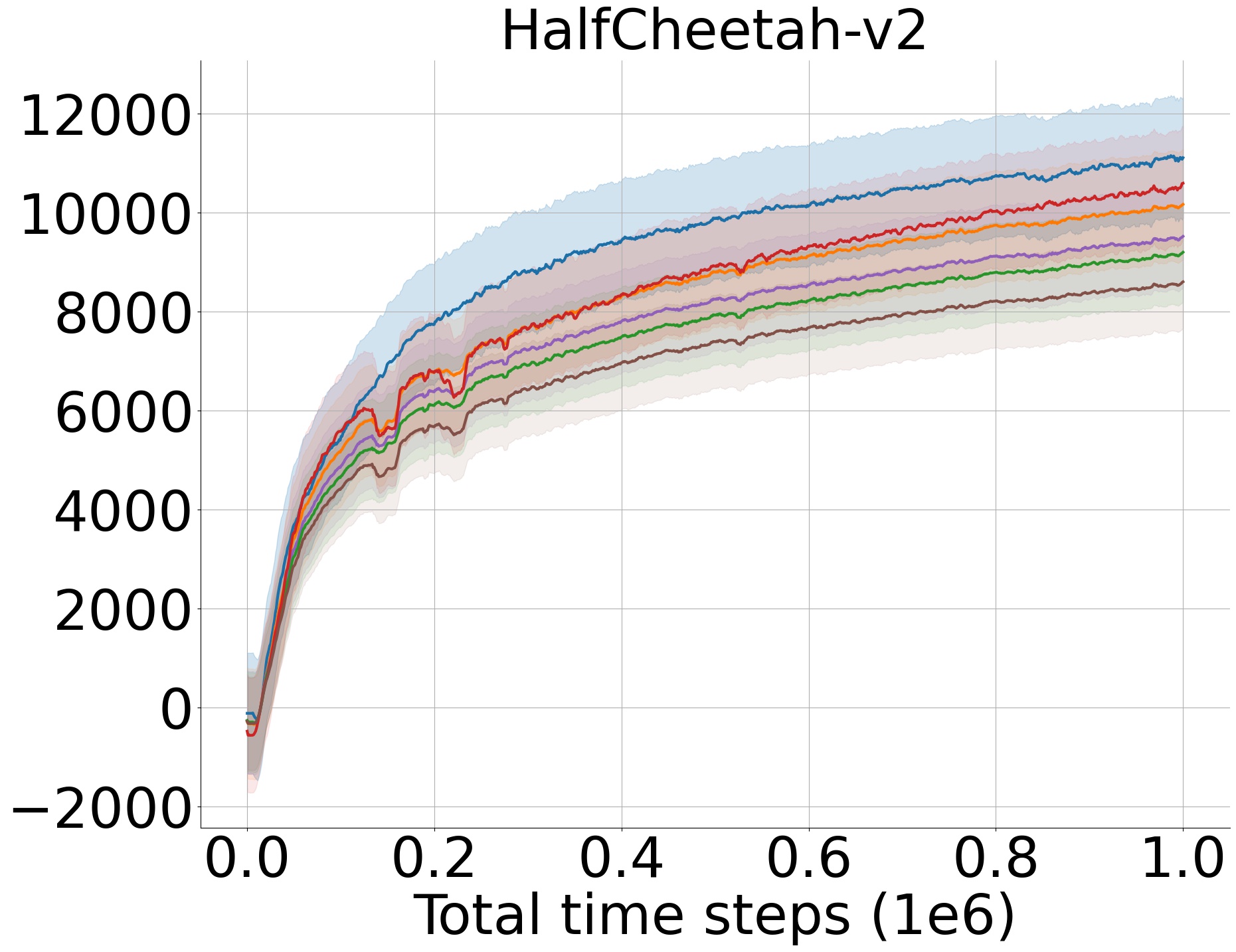}
	} \\
	\subfigure{
		\includegraphics[width=2.0in, keepaspectratio]{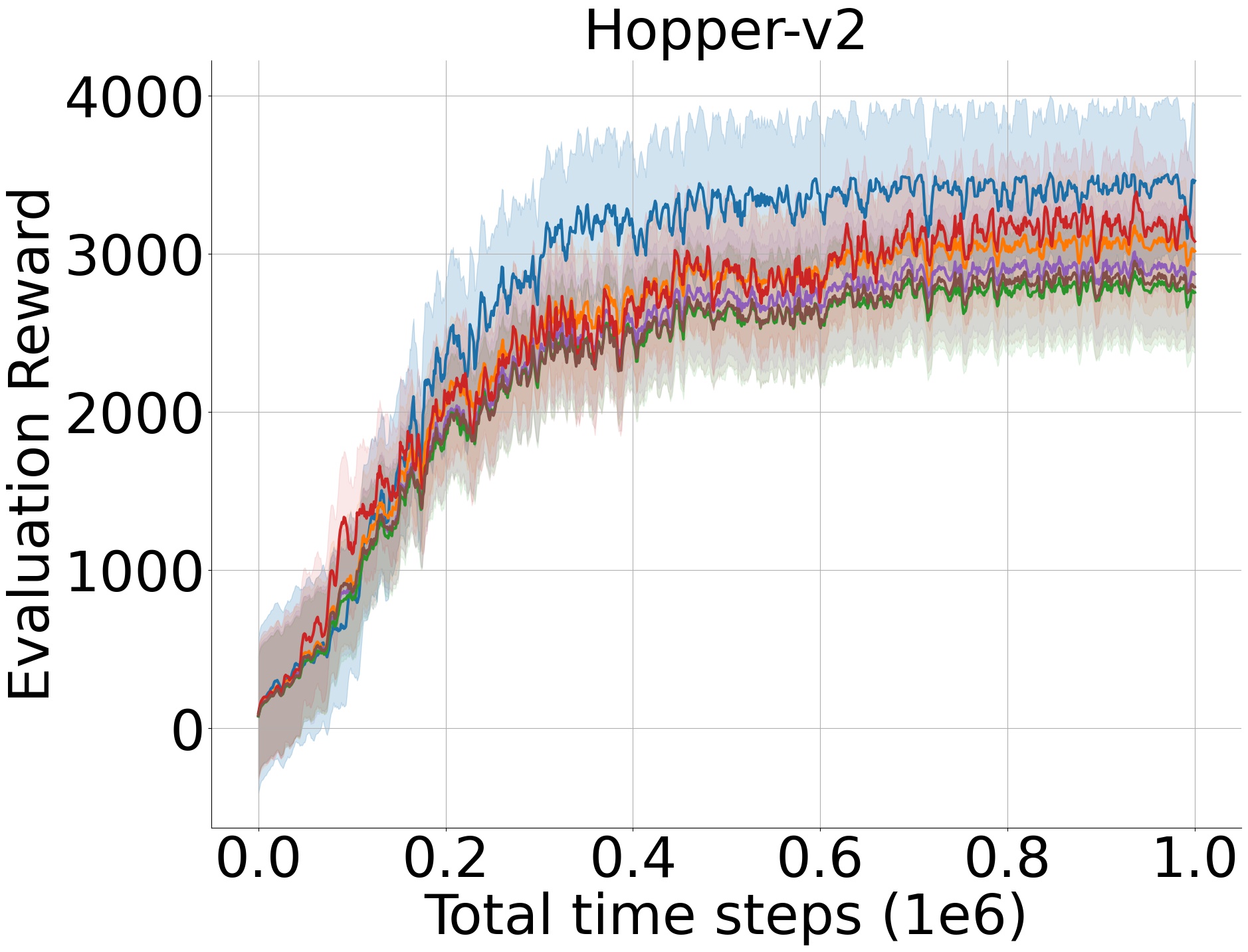}
		\includegraphics[width=2.0in, keepaspectratio]{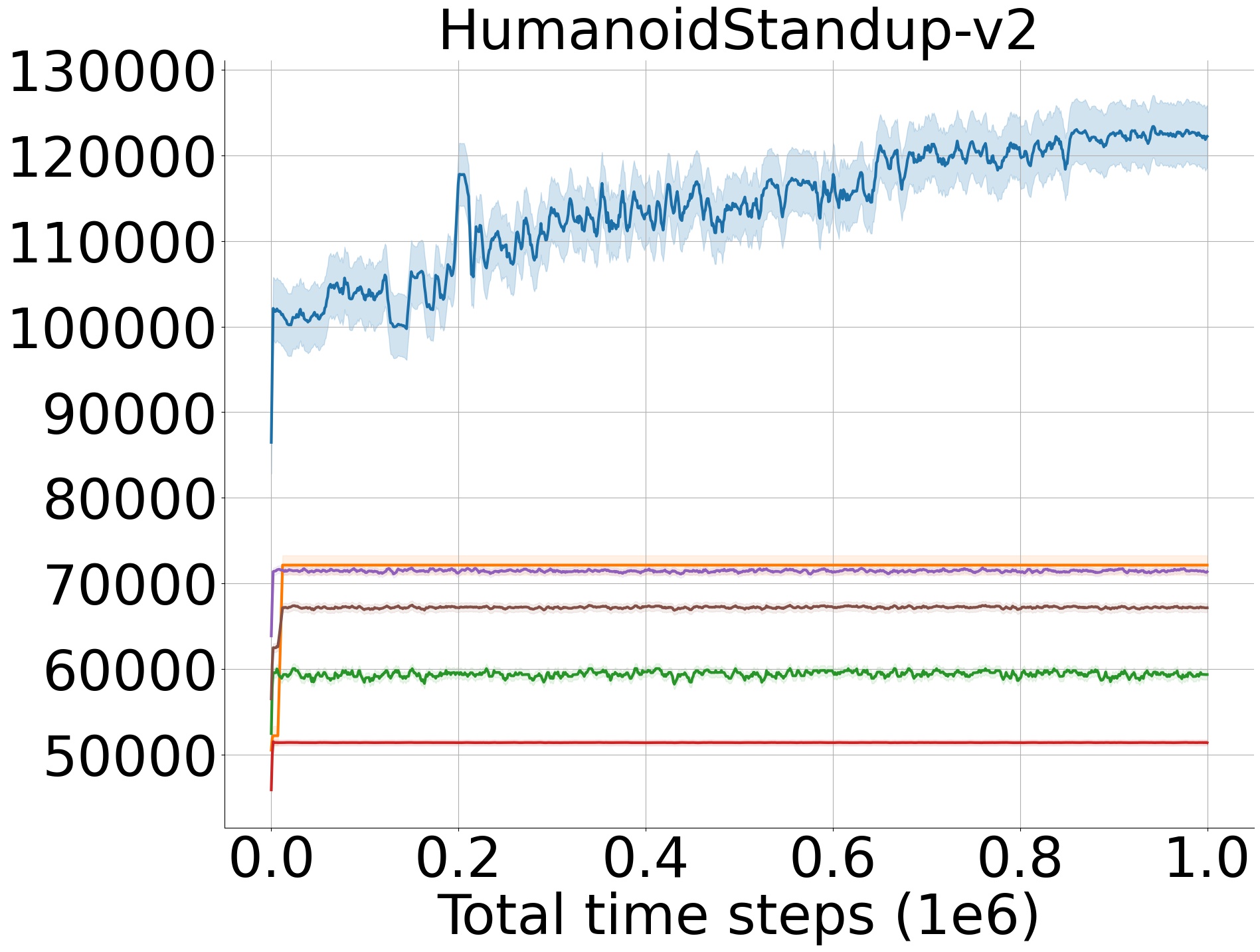}
		\includegraphics[width=2.0in, keepaspectratio]{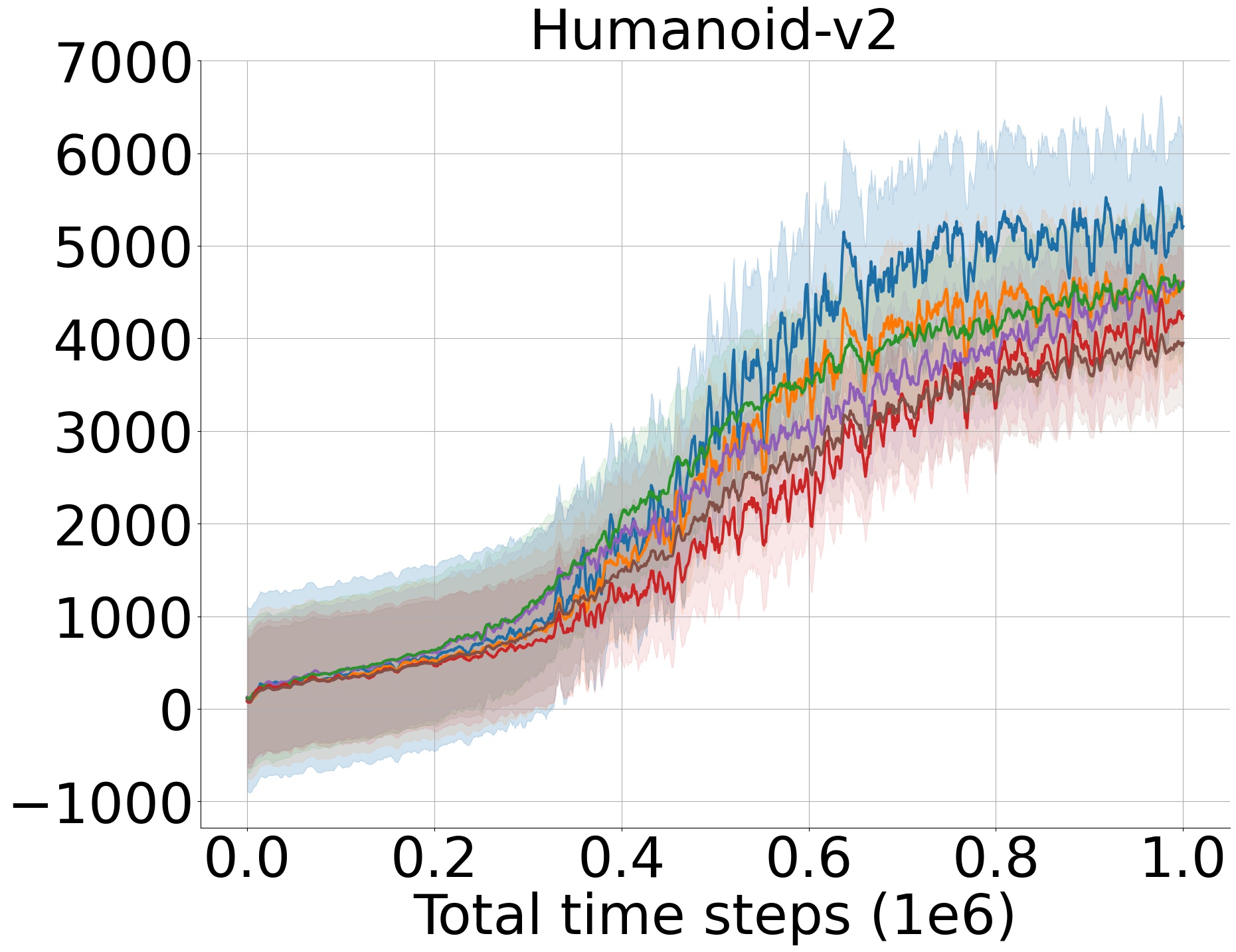}
    } \\
	\subfigure{
		\includegraphics[width=2.0in, keepaspectratio]{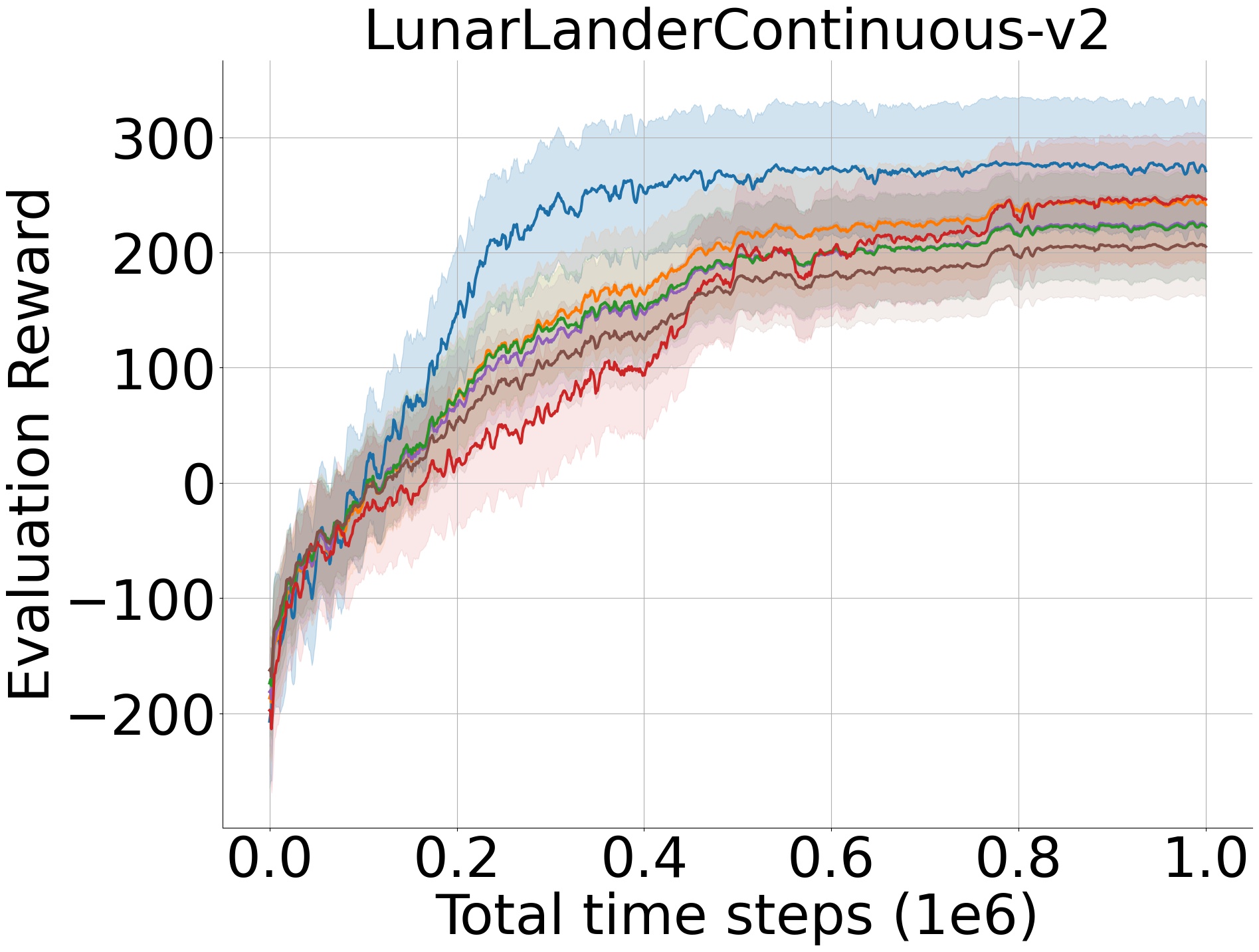}
		\includegraphics[width=2.0in, keepaspectratio]{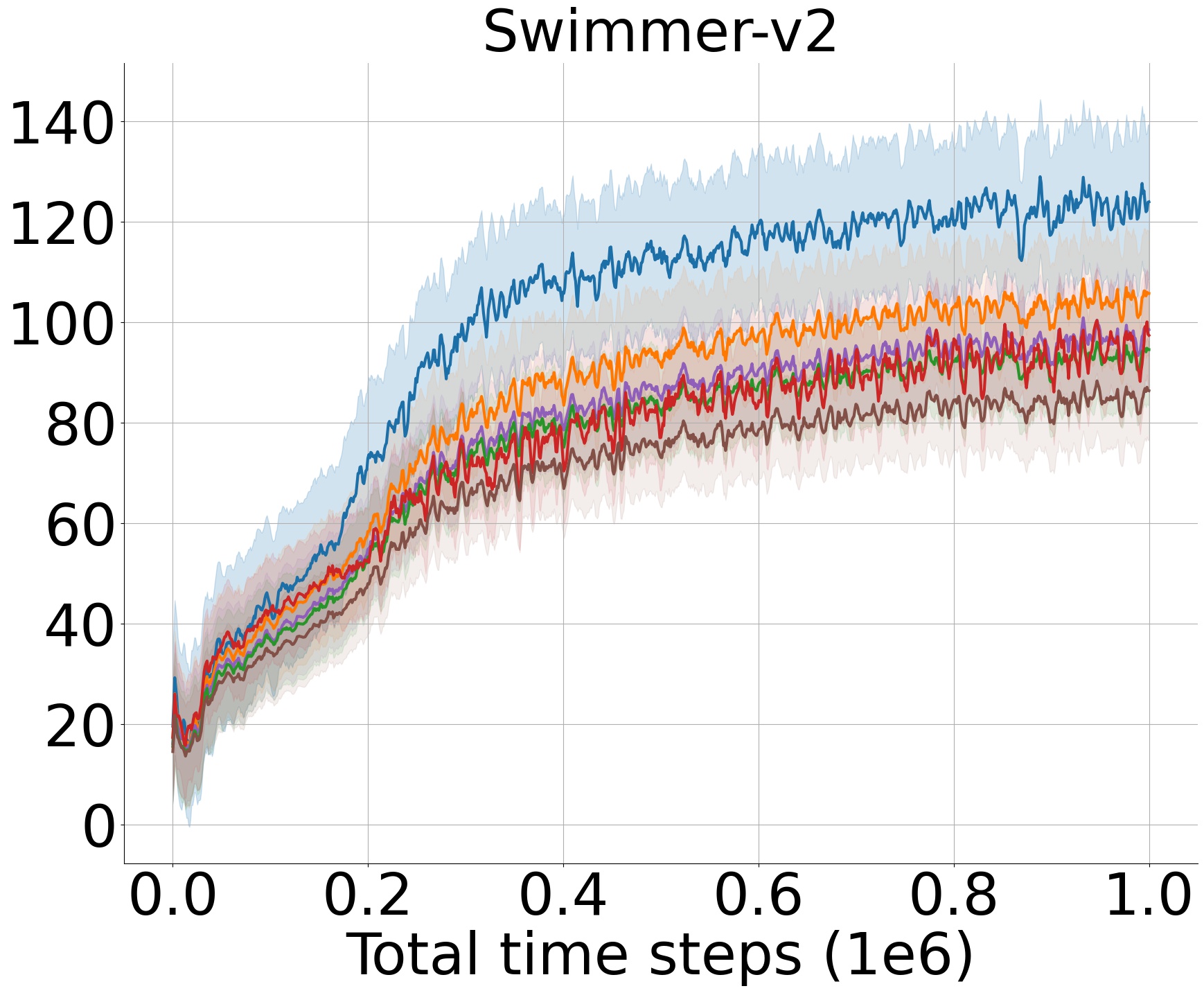}
		\includegraphics[width=2.0in, keepaspectratio]{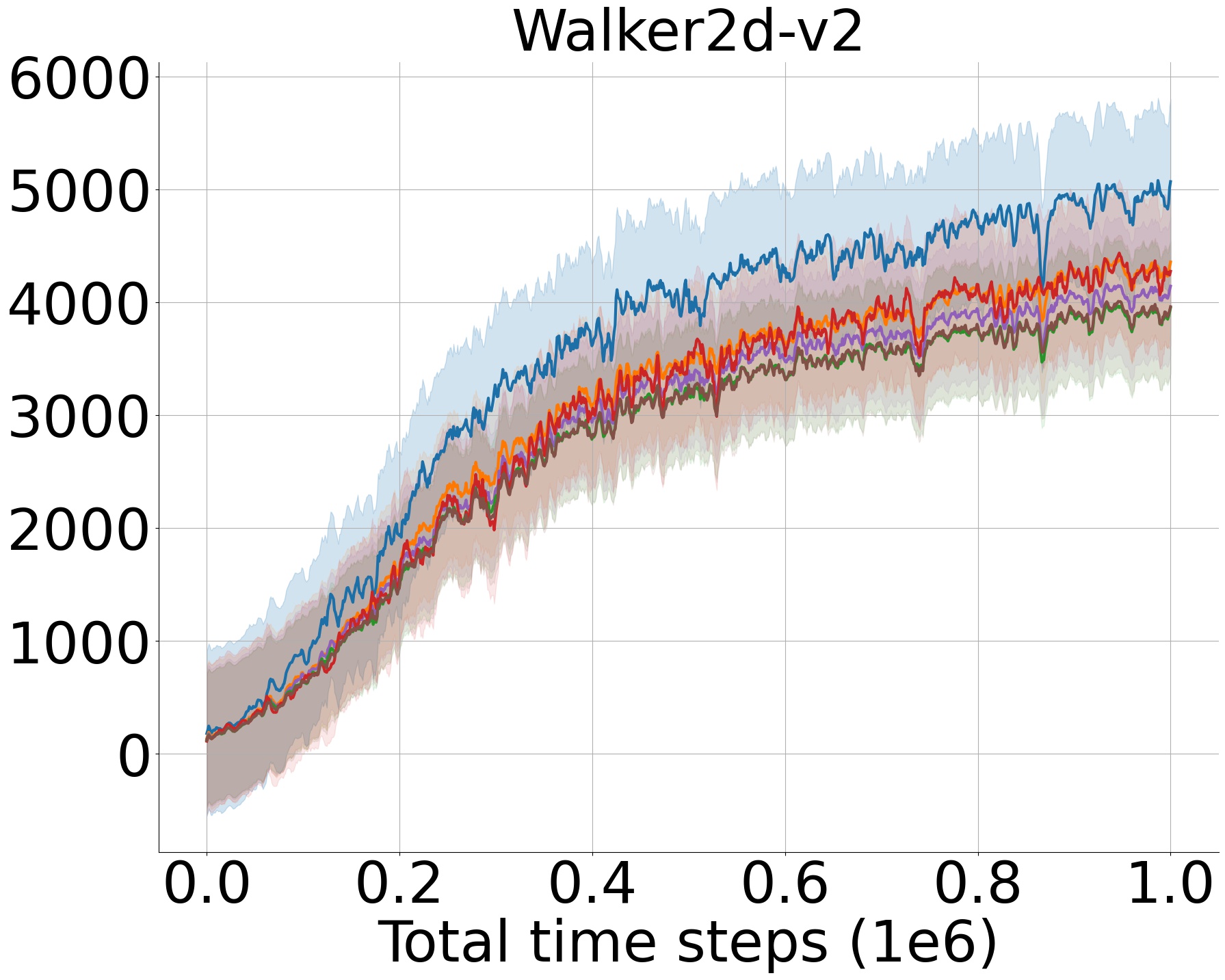}
    }
	\caption{Evaluation curves for the set of MuJoCo and Box2D continuous control tasks under the SAC algorithm. The shaded region represents half a standard deviation of the average evaluation return over 10 random seeds. A sliding window of size 5 smoothes curves for visual clarity.}
	\label{fig:eval_results_sac}
\end{figure*}

\begin{figure*}[!hbt]
    \centering
    \begin{align*}
        &\text{{\blue} DISCOVER-TD3 ($\lambda$ = 0.3)}  &&\text{{\orange} Coherent-TD3 ($\beta$ = 0.01)} &&\text{{\purple} NoisyNet-TD3} \\ &\text{{\green} PSNE-TD3} &&\text{{\red} Gaussian-TD3} &&\text{{\brown} Greedy-TD3}
    \end{align*}
	\subfigure{
		\includegraphics[width=2.0in, keepaspectratio]{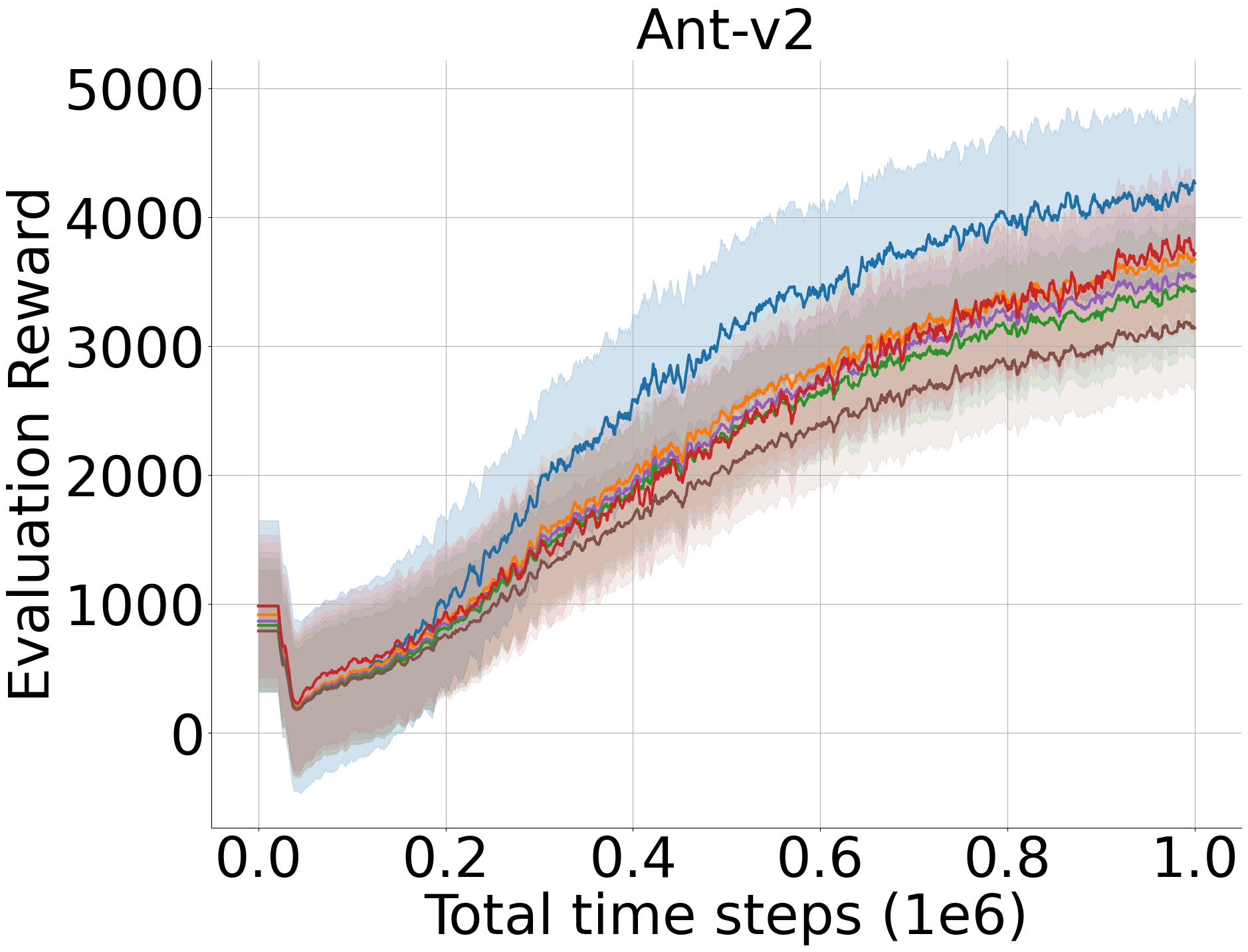}
		\includegraphics[width=2.0in, keepaspectratio]{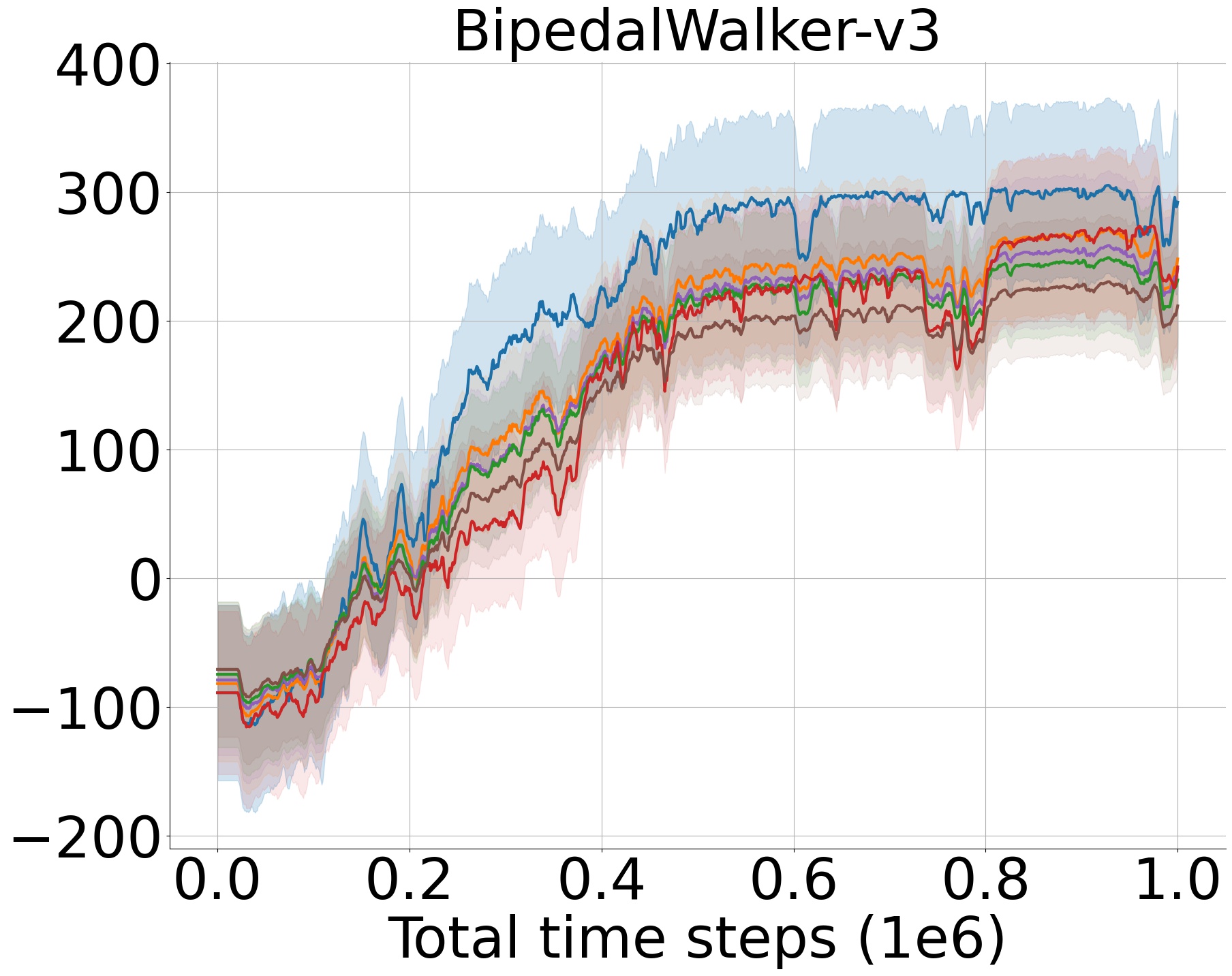}
		\includegraphics[width=2.0in, keepaspectratio]{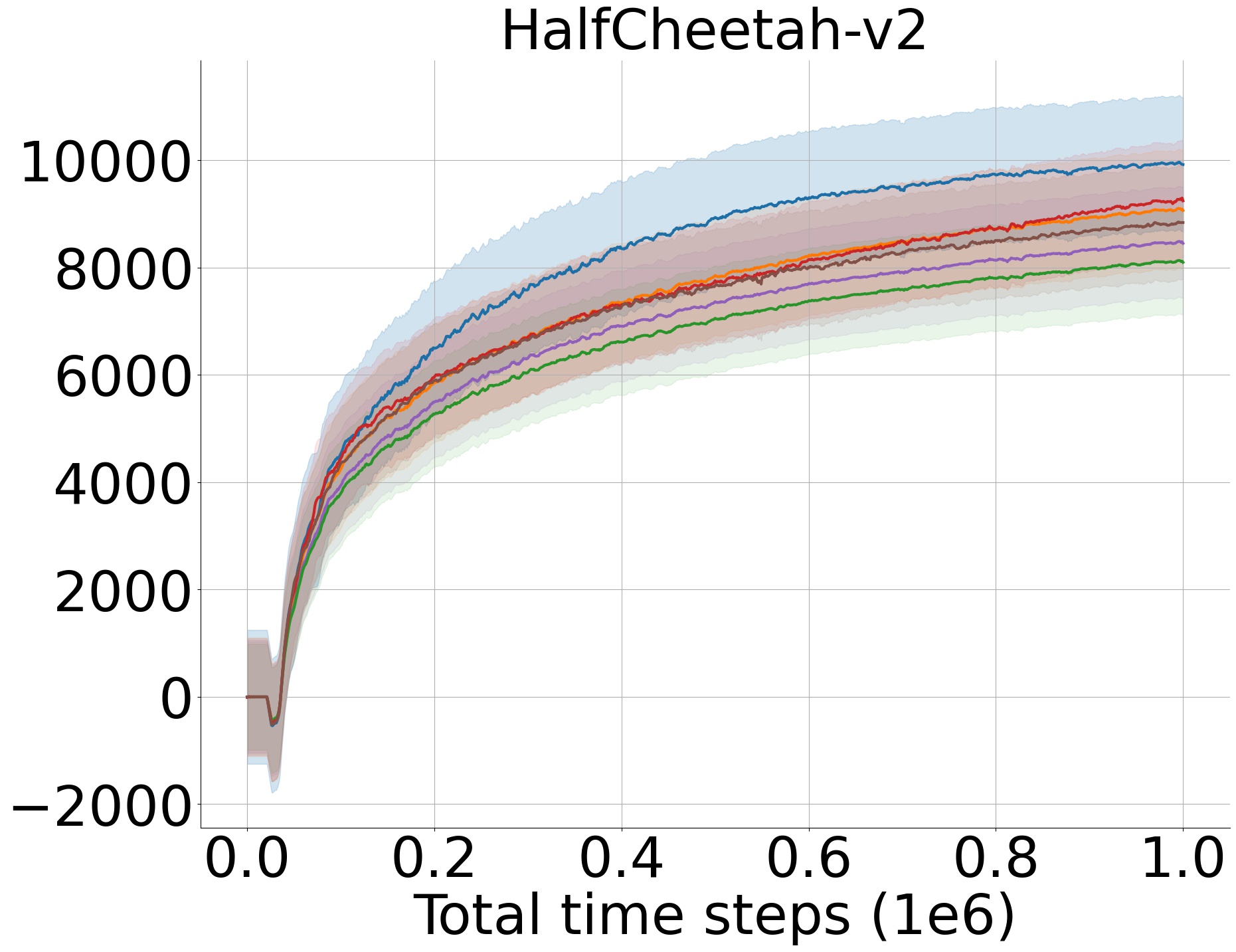}
	} \\
	\subfigure{
		\includegraphics[width=2.0in, keepaspectratio]{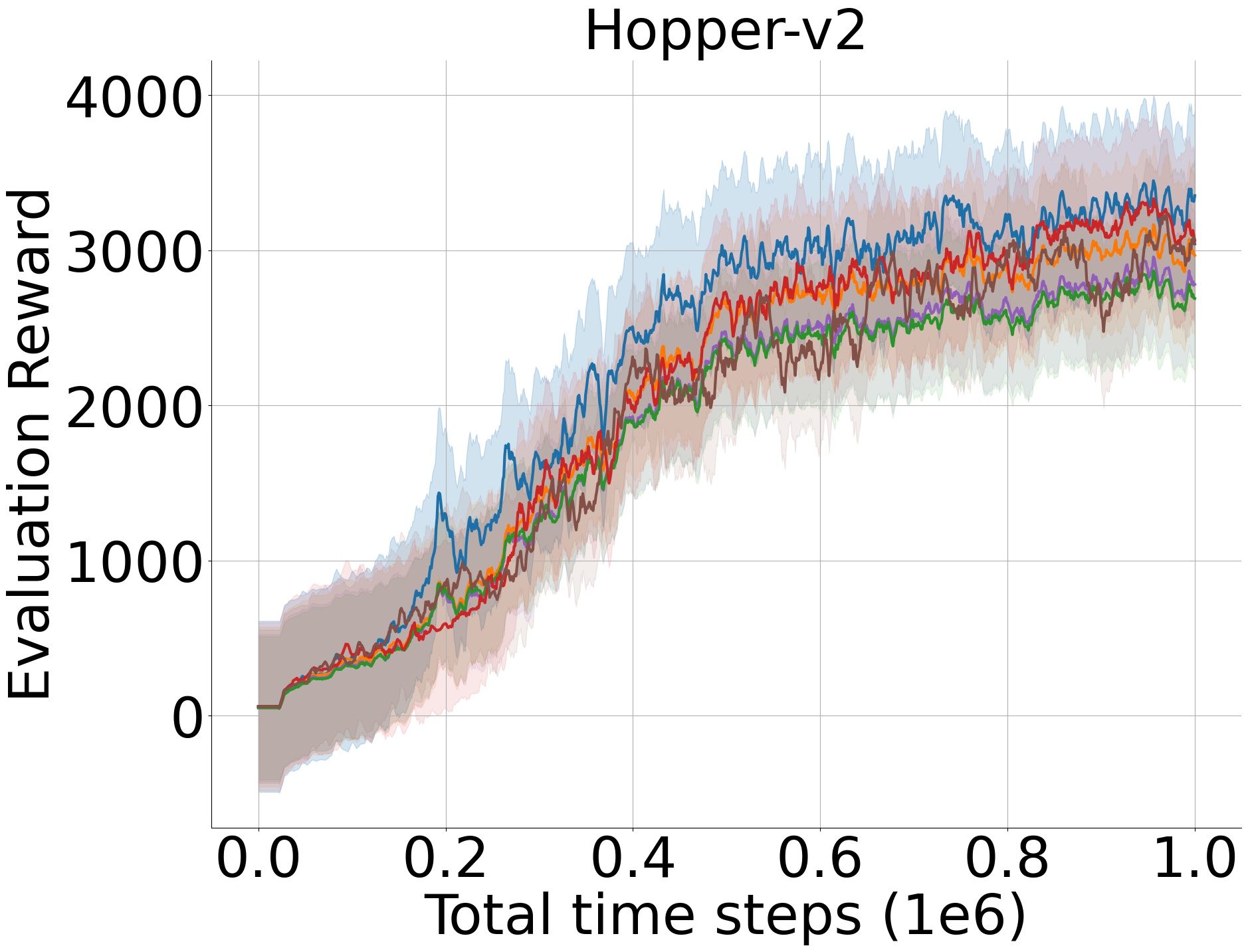}
		\includegraphics[width=2.0in, keepaspectratio]{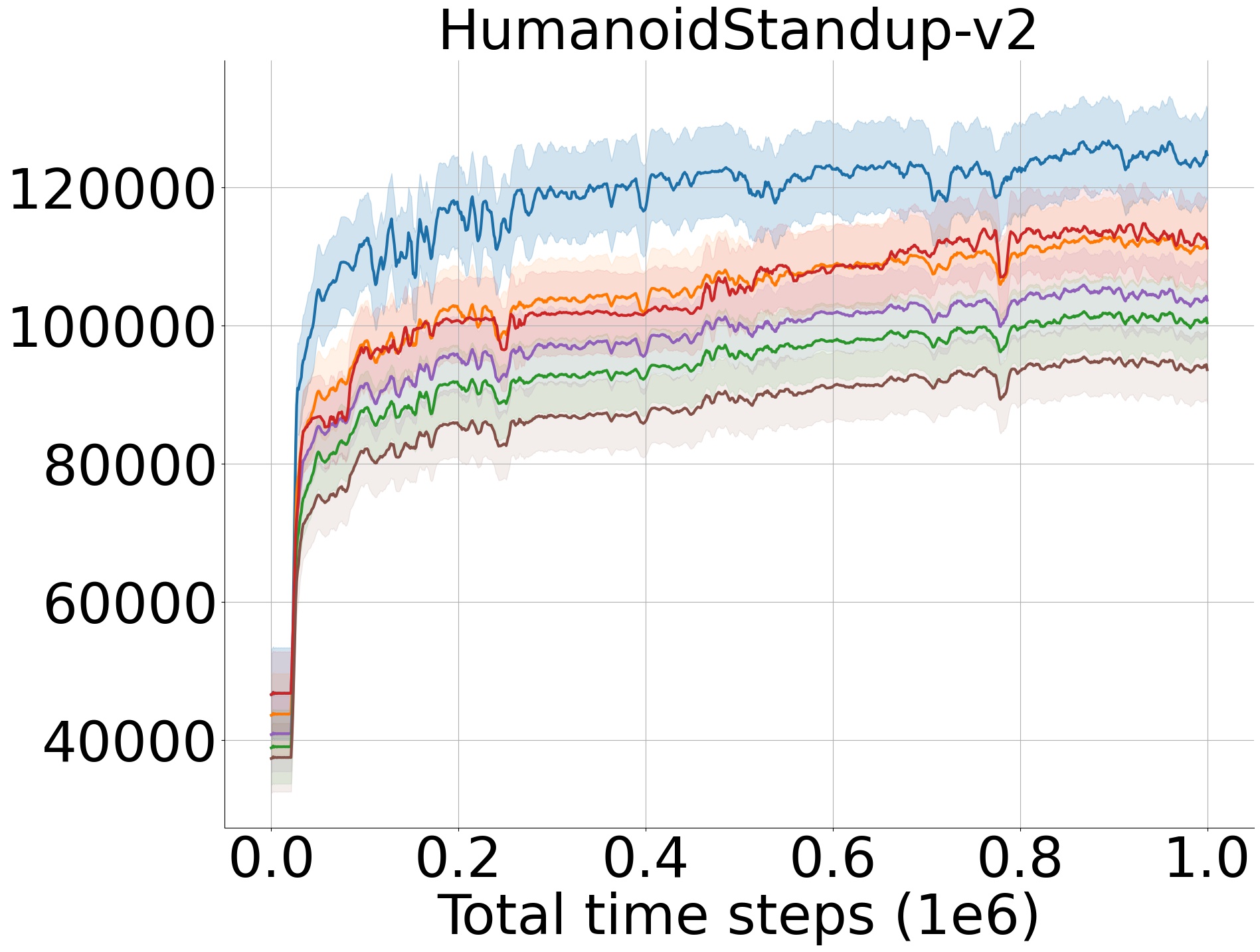}
		\includegraphics[width=2.0in, keepaspectratio]{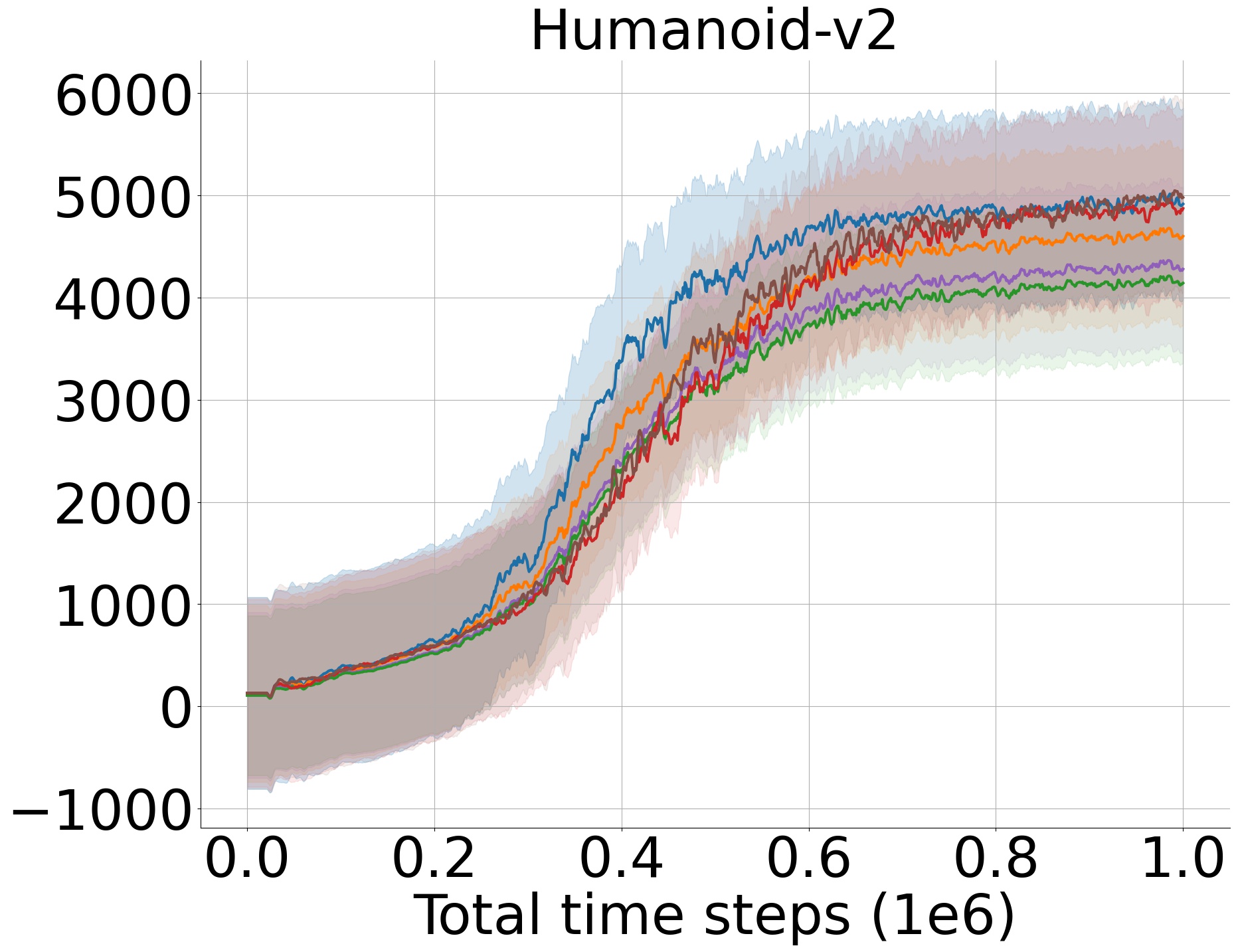}
    } \\
	\subfigure{
		\includegraphics[width=2.0in, keepaspectratio]{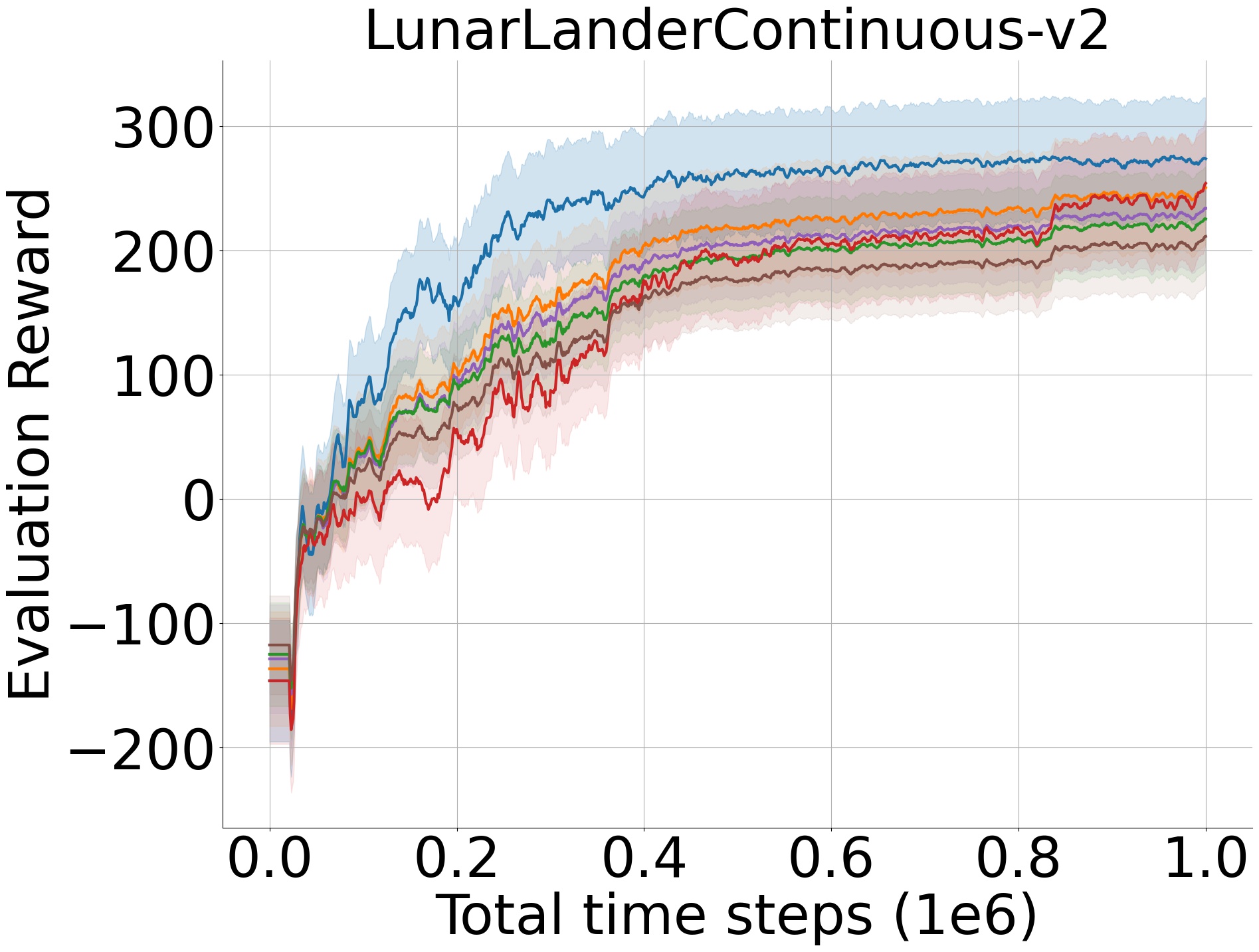}
		\includegraphics[width=2.0in, keepaspectratio]{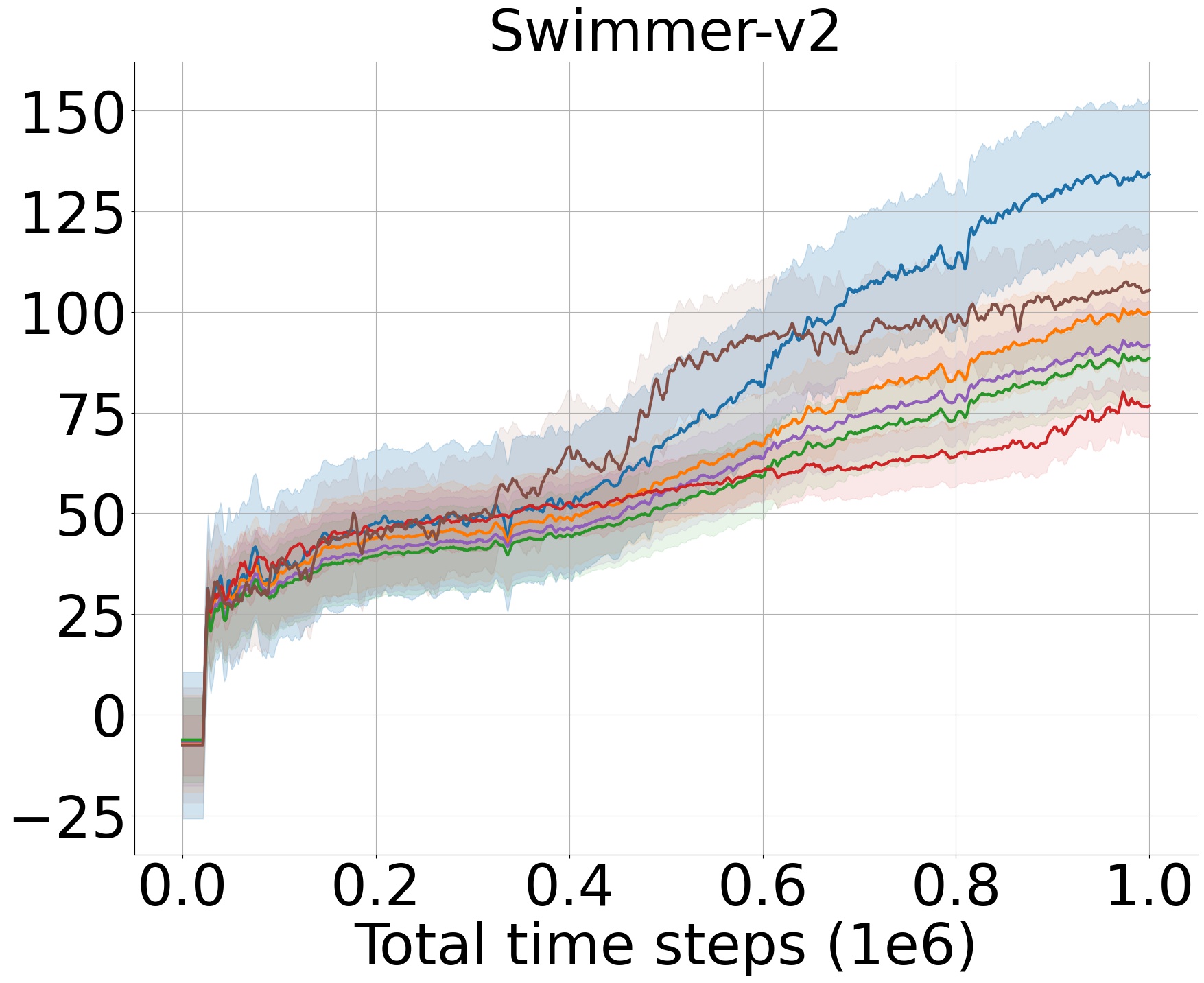}
		\includegraphics[width=2.0in, keepaspectratio]{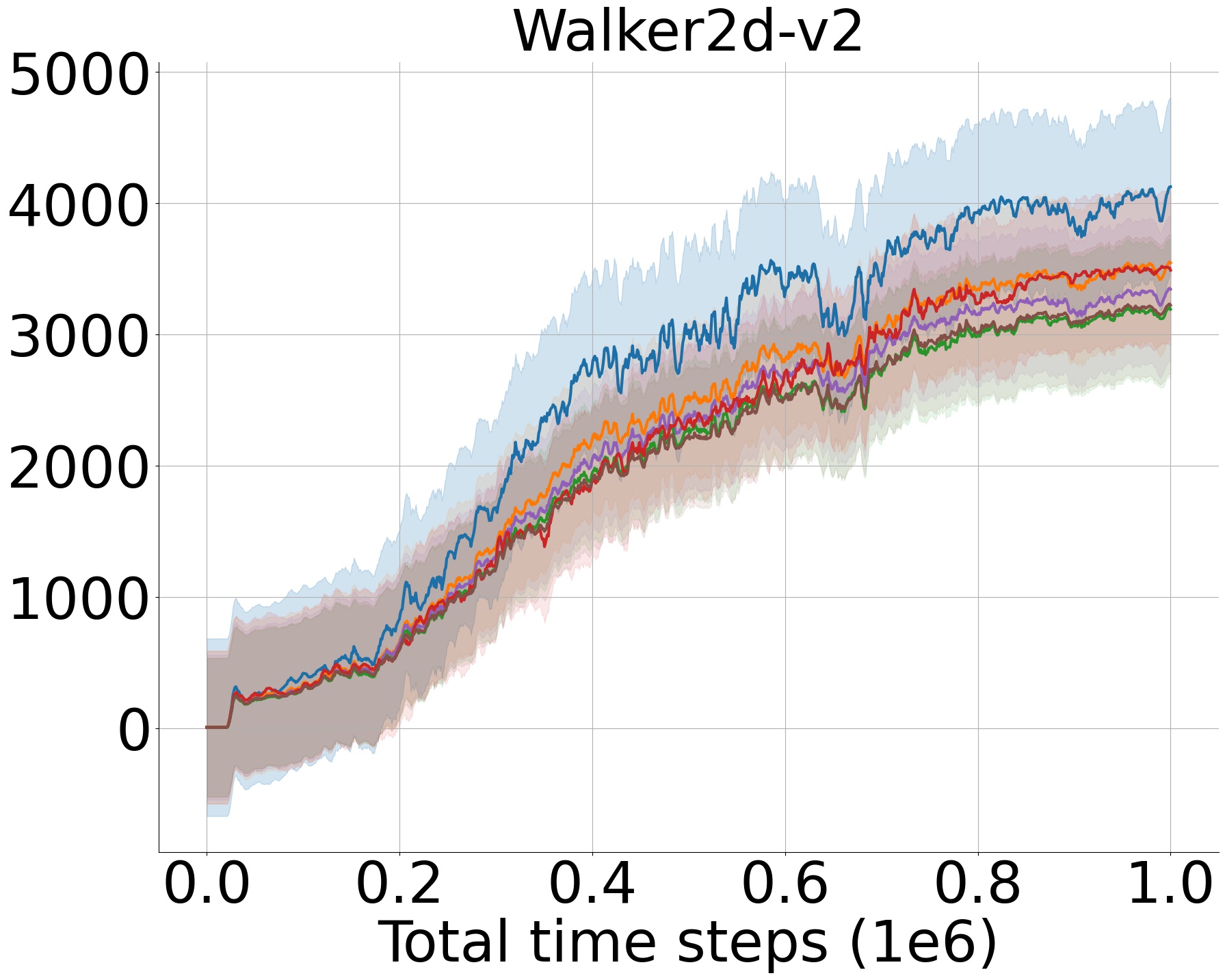}
    }
	\caption{Evaluation curves for the set of MuJoCo and Box2D continuous control tasks under the TD3 algorithm. The shaded region represents half a standard deviation of the average evaluation return over 10 random seeds. A sliding window of size 5 smoothes curves for visual clarity.}
	\label{fig:eval_results_td3}
\end{figure*}

Evaluation results for the off-policy setting are reported in Figure \ref{fig:eval_results_ddpg}, \ref{fig:eval_results_sac} and \ref{fig:eval_results_td3} for the DDPG \citep{ddpg}, SAC \citep{sac} and TD3 \citep{td3} algorithms, respectively. Complete evaluation results are provided in Appendix \ref{app:comp_eval_results}. The results demonstrate that Off-Policy DISCOVER substantially improves the baselines in the majority of the tasks. Specifically, in high-dimensional environments, Ant and Humanoid, which are considered challenging tasks \citep{deep_rl_that_matters}, DISCOVER significantly accelerates the learning while noise-based methods have a slight or no improvement. In addition, DDPG \citep{ddpg} is known exhibit a random behavior in high-dimensional environments \citep{deep_rl_that_matters} and Swimmer. Similarly, TD3 \citep{td3} and SAC \citep{sac} cannot surpass the local optima in Swimmer. We observe in these cases that the competing algorithms have a high dependence on the underlying algorithm, i.e., performance improvement is considerable only when the baseline performs well. Our method, in contrast, can break this correlation to an extent and converge to an improved policy.   

% Evaluation Results - DDPG

% Evaluation Results - SAC

% Evaluation Results - TD3

We observe that DISCOVER has a faster convergence to higher evaluation rewards for the rest of the environments and off-policy algorithms. In comparison, the undirected methods either barely improve the baseline or closely follows the no exploration setting in terms of the learning speed and attained evaluation returns. Although the undirected algorithms are previously shown to have a considerable impact on the on-policy algorithms such as A2C \citep{a2c} and PPO \citep{ppo}, they exhibit a poor improvement in the off-policy setting. In fact, for some of the environments, they underperform the Gaussian action noise \citep{williams_1992}. This poor performance is due to the asynchronous exploration and policy updates in the off-policy learning \citep{sml_11}. The undirected methods only evaluate and update their deep exploration strategy when the exploration is performed. The decoupled updates of the agent's policy and exploration strategy cannot lead agents to useful state spaces. Conversely, our method updates its exploration policy on par with the agent's policy update in an off-policy fashion, which results in an apparent and significant improvement in the off-policy setting, verifying Remark \ref{rem:imitation}. 

Moreover, Meta-Policy Gradient \citep{meta_exploration} notably improves the DDPG algorithm \citep{ddpg} compared to the action and parameter-space noise techniques. This shows the effectiveness of the directed exploration over the undirected methods in off-policy learning. Nevertheless, the highest returns are attained by DISCOVER in our experiments with DDPG \citep{ddpg}. This is due to the additional time duration spent on the teacher network's data collection in the Meta-Policy Gradient algorithm \citep{meta_exploration}. In comparison, Figure \ref{fig:eval_results_ddpg} shows that our method exhibits a more robust and better performance as it simultaneously learns to explore with the agent's policy.

Although DISCOVER is shown to attain higher returns in all tasks, for stable environments where the number of time steps is fixed, such as HalfCheetah, our method shows a slight advantage over the competing methods. This is due to the smaller exploration degree requirement in the stable environments as stated by \citet{deep_rl_that_matters}. In contrast, the performance improvement in unstable environments is more considerable as undirected exploration strategies can cause sudden failures in off-policy learning \citep{deep_rl_that_matters}. Moreover, we still observe an improved but suboptimal behavior when the underlying off-policy algorithm does not converge to the optimal policy, mainly under the DDPG algorithm \citep{ddpg}. As discussed, it is natural for an exploration strategy not to cover optimal policies when the baseline is incapable of sufficiently solving the environment. Nonetheless, our empirical studies reflect the psychological approach to the exploration-exploitation dilemma, that is, informative and directed guidance through the gradient of the evaluation function prevents the boredom of the agents to encourage the exploration, which validates Remark \ref{rem:in_mot_on} and \ref{rem:in_mot_off}. Thus, we conclude that the introduced directed exploration scheme can overcome the limitations induced by the random walk behavior in the undirected approaches. 

\subsection{Ablation Studies}
We perform ablation studies to analyze the effects of the components: exploration direction regularization $\lambda$ and delayed exploration policy updates, target explorer network, and target smoothing regularization in the off-policy TD-learning \citep{td3}. As delayed policy updates, target policy networks, and target smoothing regularization are employed in the TD3 algorithm \citep{td3}, we conduct our ablations on TD3 \citep{td3} for a comprehensive study. Therefore, the resulting performance of Off-Policy DISCOVER is analyzed when removing or altering each of these components. Moreover, we provide the results for On-Policy DISCOVER under the PPO algorithm \citep{ppo} for the considered $\lambda$ values. The effect of the rest of the DISCOVER components is not available in the on-policy setting since the policy updates are not delayed, and the target network is not utilized in on-policy algorithms.

\begin{table}[!hbt]
\begin{center}
    \caption{Average return over the last 10 evaluations over 10 trials
    of 1 million time steps, comparing ablation over DISCOVER under $\lambda = \{0.0, 0.1, 0.3, 0.6, 0.9, 1.0\}$, Off-Policy DISCOVER without delayed explorer policy updates (DPU), target network (TN), and target smoothing regularization (TSR). Bold values represent the maximum for each environment under on- or off-policy setting. The PPO and TD3 algorithms are used as baselines for On-Policy and Off-Policy DISCOVER, respectively.}
    \label{table:ablation}
    \begin{tabular}{@{} cl*{4}c @{}}
        \toprule
        & \textbf{Setting} & \textbf{HalfCheetah-v2} & \textbf{Humanoid-v2} & \textbf{Hopper-v2} & \textbf{Swimmer-v2} \\
        \midrule
        & $\lambda = 0.0$             & 3179.35 & 700.42 & 3126.34 & 127.85 \\
        & $\lambda = 0.1$             & \textbf{5224.88} & \textbf{1010.48} & \textbf{3436.32} & \textbf{134.25} \\
        & $\lambda = 0.3$             & 2500.32 & 634.86 & 1890.07 & 122.57 \\
        & $\lambda = 0.6$             & 2621.43 & 626.17 & 2140.21 & 90.26 \\
        & $\lambda = 0.9$             & 1822.65 & 528.45 & 1607.60 & 56.78 \\
        \rot{\rlap{~\textbf{On-Policy}}}
        & $\lambda = 1.0$             & 1933.92 & 555.24 & 1581.94 & 74.23 \\ 
        \midrule
        & $\lambda = 0.0$             & 8853.83 & 3041.59 & 4993.90 & 105.19 \\
        & $\lambda = 0.1$             & 8762.55 & 2704.40 & \textbf{5080.31} & 107.76 \\
        & $\lambda = 0.3$             & \textbf{9939.39} & \textbf{3350.27} & 4941.76 & \textbf{133.94} \\
        & $\lambda = 0.6$             & 7955.40 & 2651.23 & 4719.90 & 88.81 \\
        & $\lambda = 0.9$             & 4594.68 & 795.43 & 393.91 & 43.13 \\
        & $\lambda = 1.0$             & 2732.76 & 253.12 & 203.45 & 25.92 \\ \\
        
        \rot{\rlap{~\textbf{Off-Policy}}}
        
        & w/o DPU                     & 8900.64 & 2794.50 & 4924.07 & 105.55 \\
        & w/o TN                      & 8756.69 & 3248.61 & 4463.95 & 115.48 \\
        & w/o TSR                     & 8795.56 & 2953.44 & 4857.31 & 113.02 \\
        \bottomrule
    \end{tabular}
\end{center}
\end{table}

As the significance of each component may vary from task to task, we consider four environments with different characteristics in our ablation studies. We choose the low-dimensional environment Hopper, high-dimensional environment Humanoid, and environments that vastly require on- and off-policy samples to be solved, Swimmer and HalfCheetah, respectively \citep{deep_rl_that_matters}. Ablation results are reported in Table \ref{table:ablation}. Unless otherwise stated, $\lambda = 0.3$ is used for Off-Policy DISCOVER. Learning curves for the ablation studies are given in Appendix \ref{app:ab_study_learning_curves}.

Since larger values of $\lambda$ increase the exploration, $\lambda = 1$ indicates a full exploration while $\lambda = 0$ represents the greedy action selection. From Table \ref{table:ablation}, we observe that after around $\lambda = 0.3$, the performance of Off-Policy DISCOVER dramatically decreases due to the highly perturbed action selections, which prevents agents from learning the environment. The exploration becomes insufficient for smaller $\lambda$ and yields suboptimal policies. This is similar for the on-policy setting for which after around $\lambda = 0.1$, the underlying distribution of the observed state is highly perturbed, which degrades the learning efficiency. In addition, $\lambda = 0.0$ results in greedy action selection under which the agent is stuck at a suboptimal policy.

We additionally compare the effectiveness of the target explorer network, perturbing the next action in the TD-learning \citep{sutton_1988}, i.e., target smoothing regularization \citep{td3}, and delayed exploration policy updates \citep{td3} in the off-policy setting. We perform the target policy regularization \citep{td3} with the behavioral explorer network when the target network is removed from the complete algorithm. Therefore, we observe that the resulting performance is similar to the no target policy smoothing \citep{td3} case. Although these minor modifications have a slight advantage, we still observe a further improvement in Swimmer. Our ablation studies infer that designing our exploration framework to mimic the policy framework of the baseline can achieve maximum performance due to the coupled explorer and policy networks and verifies Remark \ref{rem:imitation}. Hence, when applied to an actor-critic algorithm, DISCOVER should follow the exact setting of the policy framework. This results in a single hyper-parameter optimized, the exploration regularization term $\lambda$.

\subsection{Visualization of the State Visitations}
To study the behavior of our exploration policy, we investigate the collected set of transitions throughout the training of Off-Policy DISCOVER under the TD3 algorithm \citep{td3}. We focus on off-policy learning for this study since it allows a step-wise progressive learning rather than the on-policy algorithms in which a single behavioral policy with frozen parameters collects the experiences for a long rollout duration. Hence, the transitions stored in the experience replay buffer \citep{experience_replay} can provide an interpretable visualization on the state space visitations, as shown in \citep{meta_exploration}. 

\begin{figure*}[!hbt]
    \centering
    \begin{align*}
        &\text{\textcolor{custom_blue}{$\blacksquare$} No Exploration}  &&\text{\textcolor{custom_orange}{$\blacksquare$} DISCOVER} \\
        &\text{\textcolor{custom_green}{$\blacksquare$} No Exploration (TD-error)} 
        &&\text{\textcolor{custom_purple}{$\blacksquare$} DISCOVER (TD-error)}
    \end{align*}
	\subfigure[Early]{
		\includegraphics[width=1.5in, keepaspectratio]{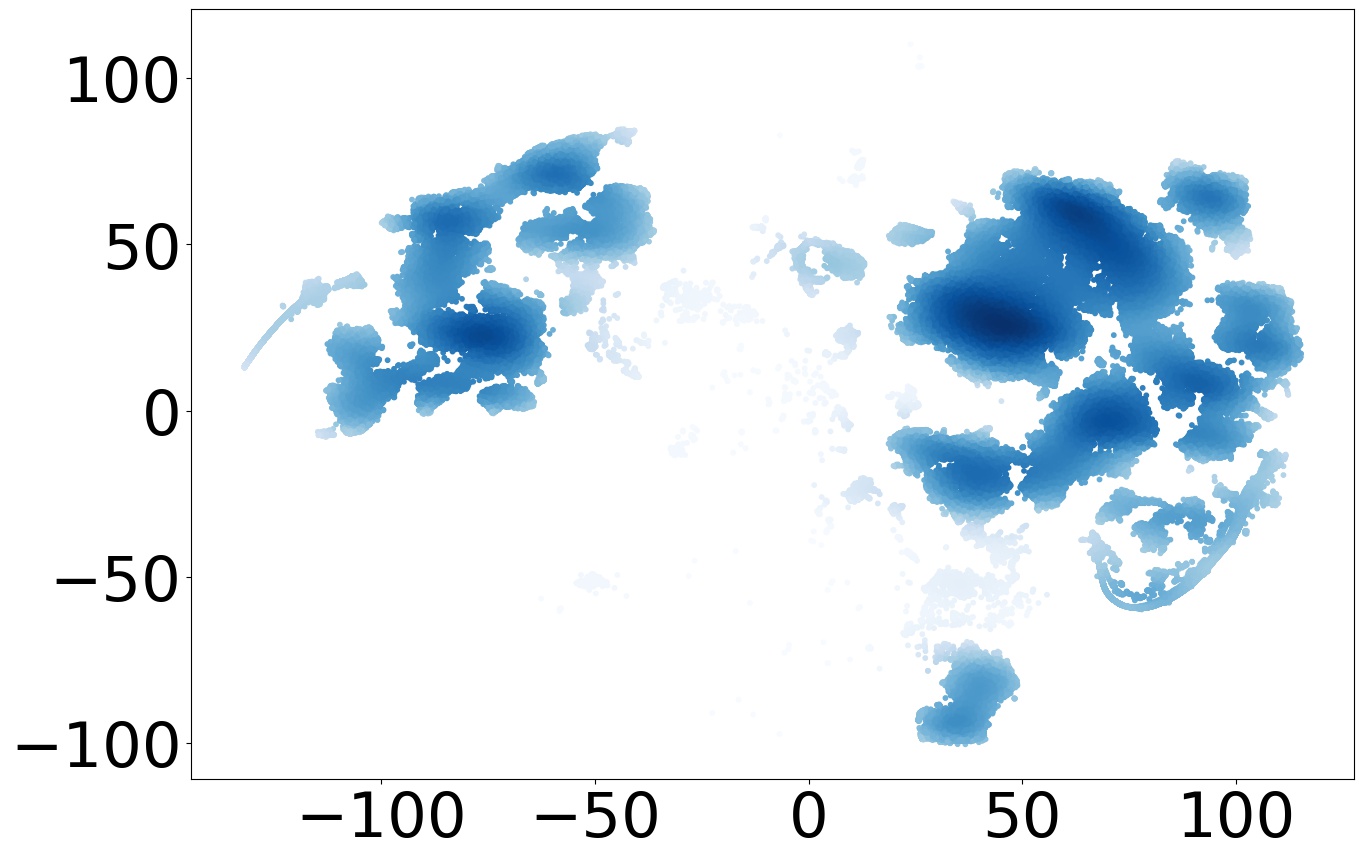}
		\includegraphics[width=1.5in, keepaspectratio]{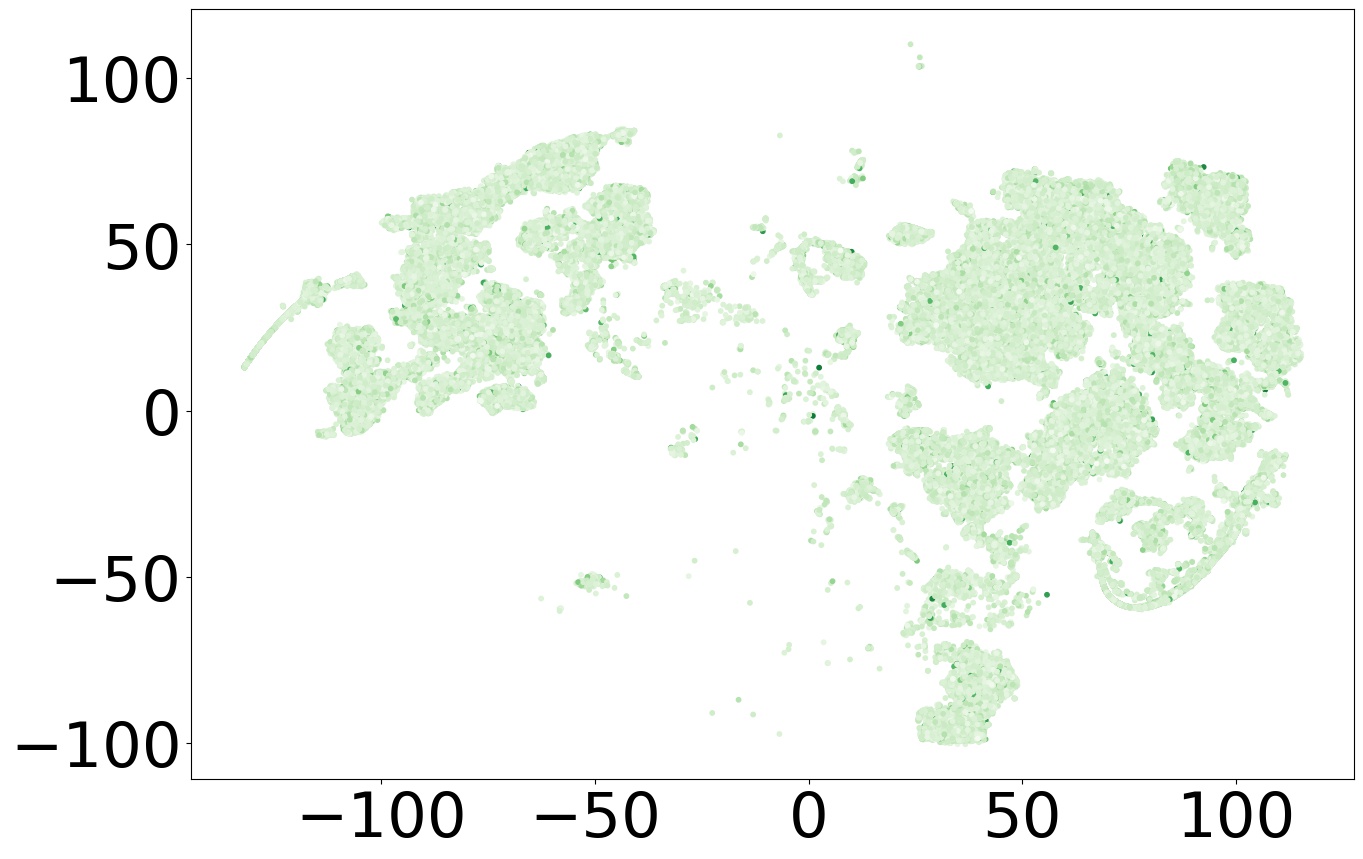}
		\includegraphics[width=1.5in, keepaspectratio]{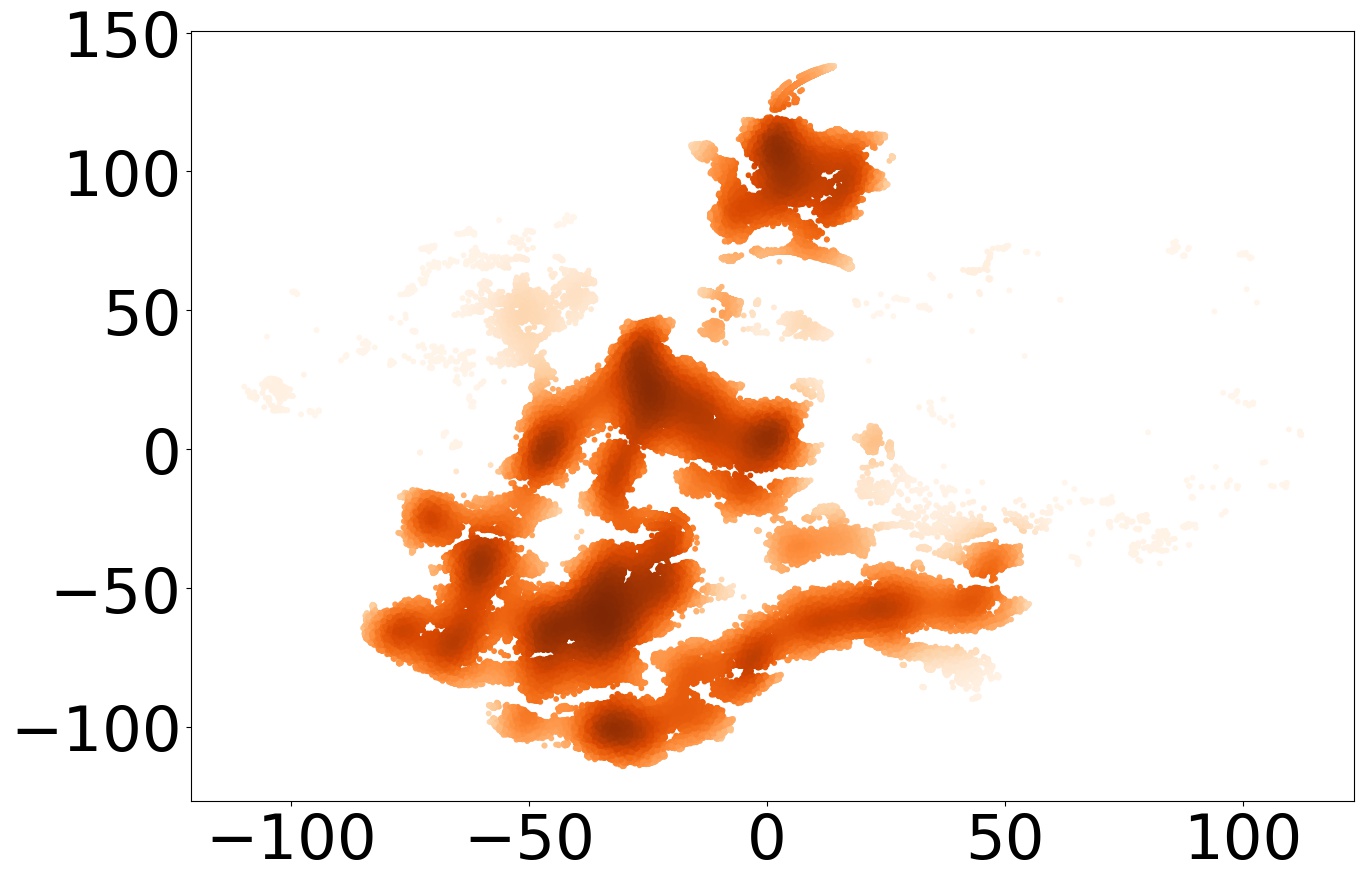}
		\includegraphics[width=1.5in, keepaspectratio]{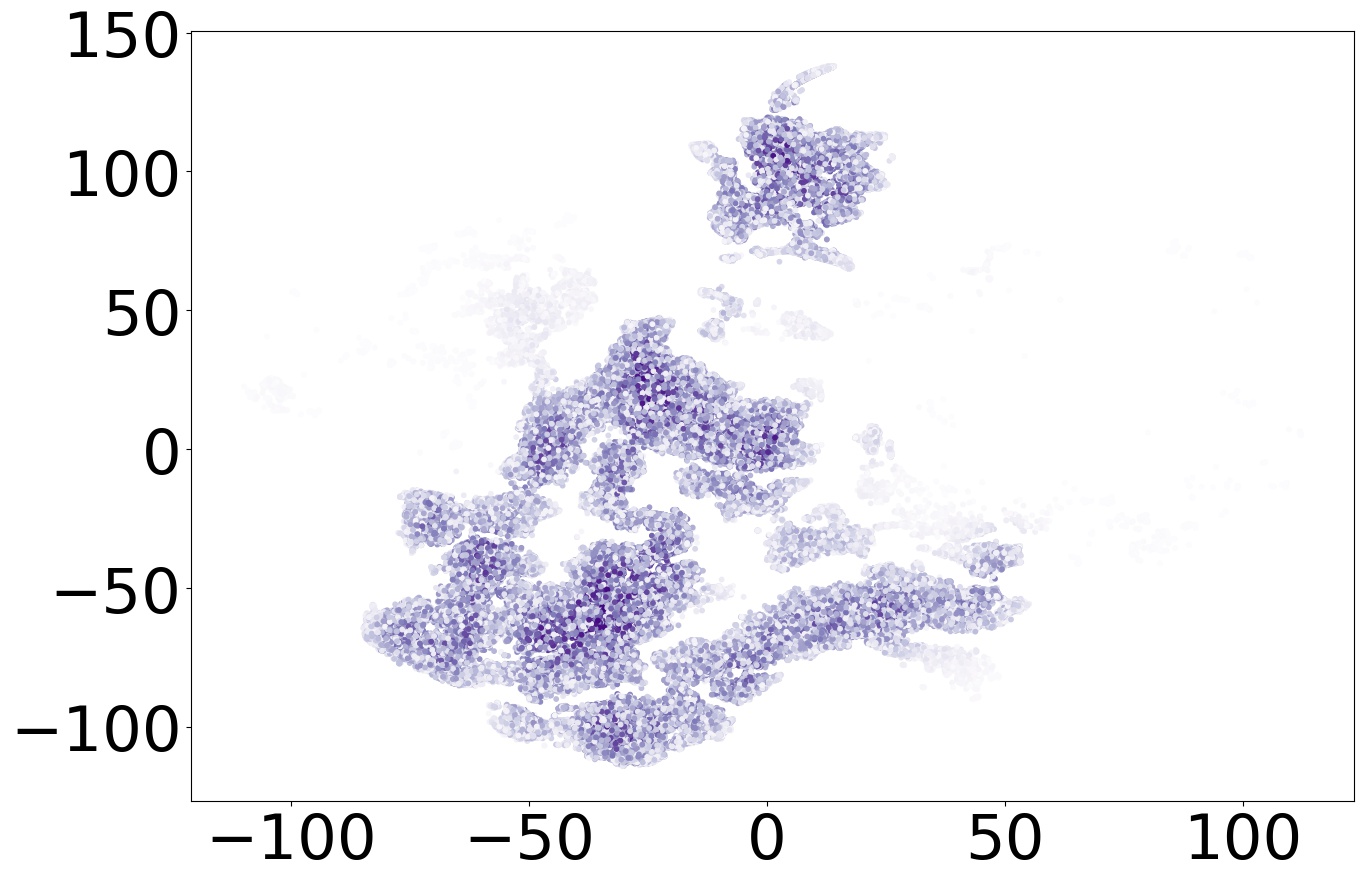}
	} \\
	\subfigure[Intermediate]{
		\includegraphics[width=1.5in, keepaspectratio]{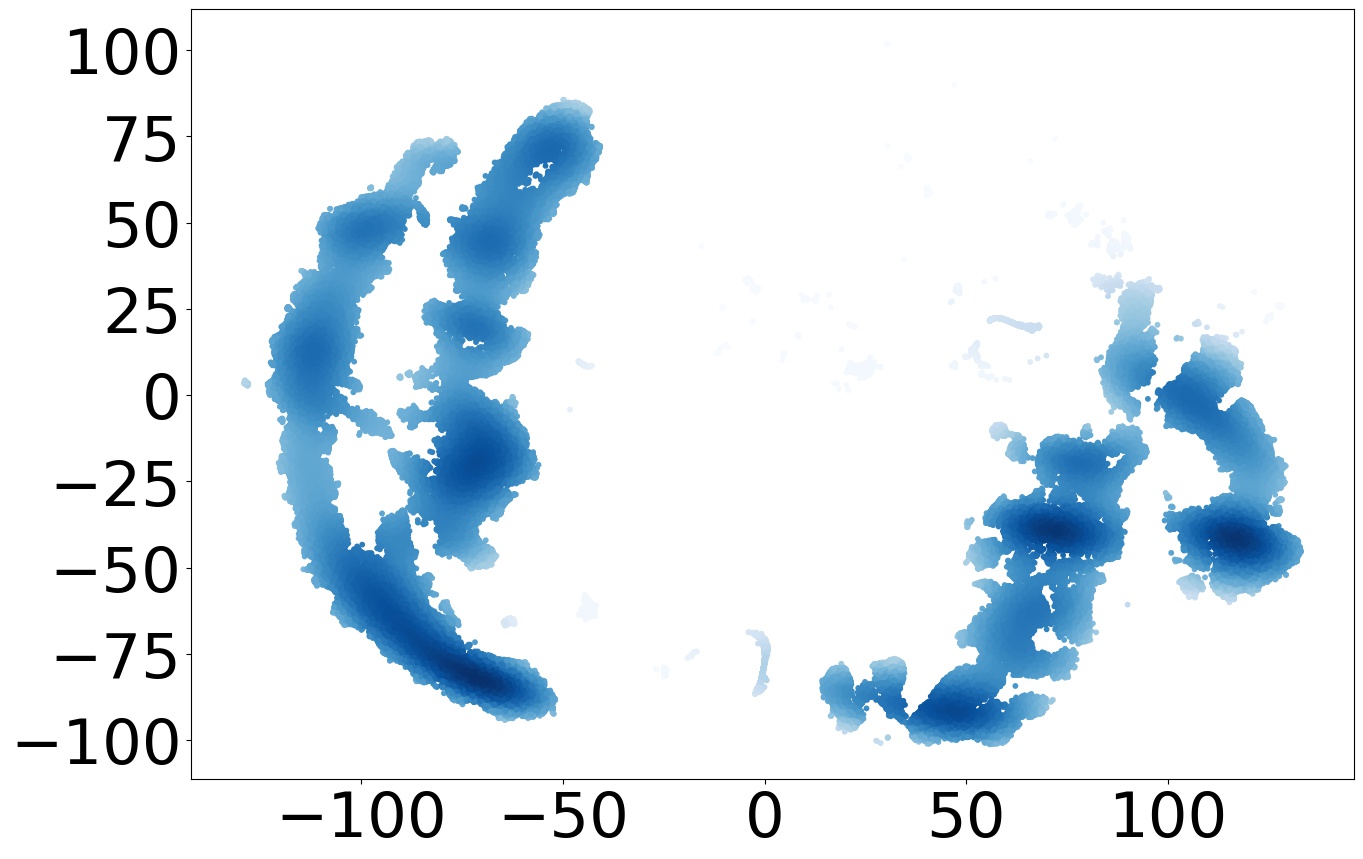}
		\includegraphics[width=1.5in, keepaspectratio]{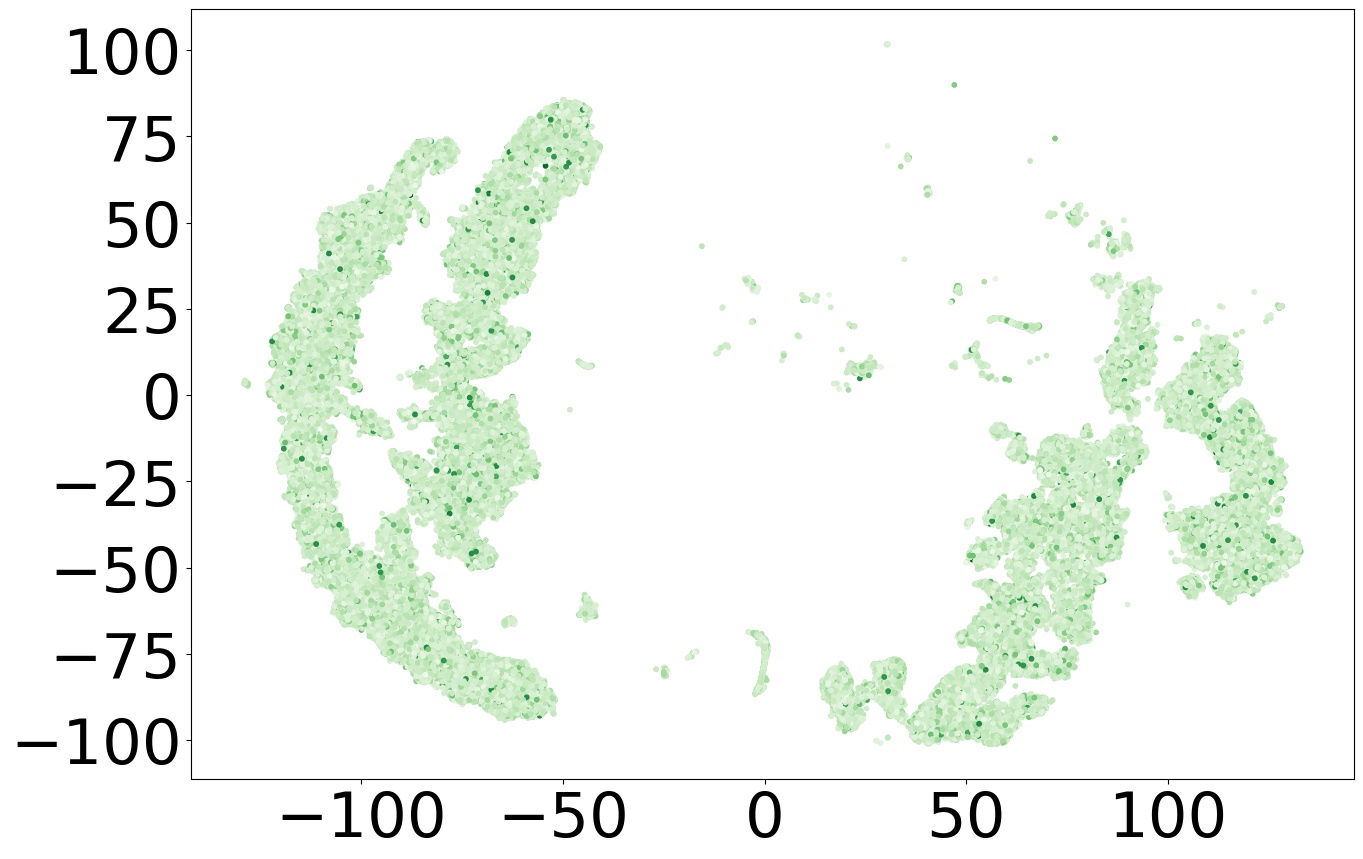}
		\includegraphics[width=1.5in, keepaspectratio]{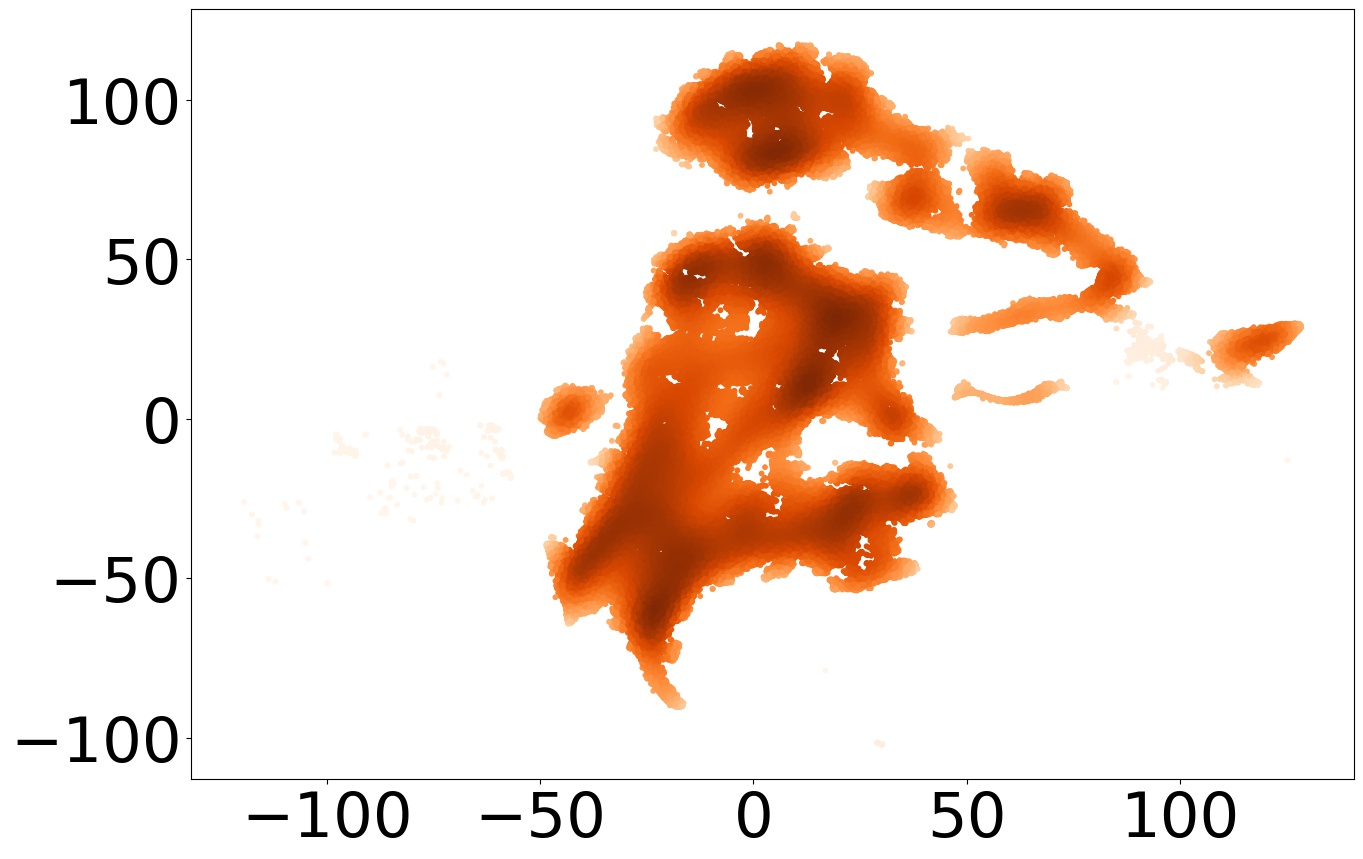}
		\includegraphics[width=1.5in, keepaspectratio]{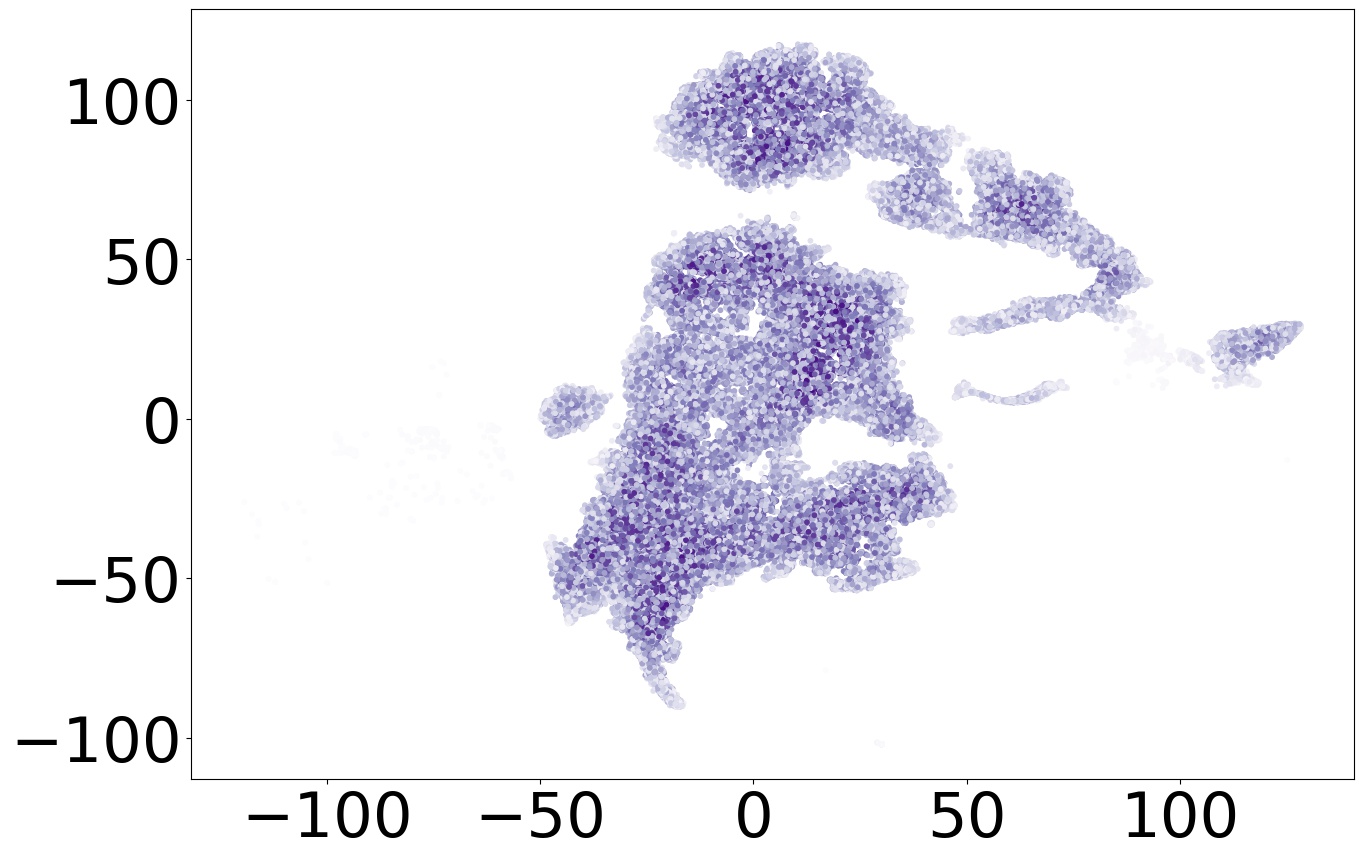}
	} \\
	\subfigure[Late]{
		\includegraphics[width=1.5in, keepaspectratio]{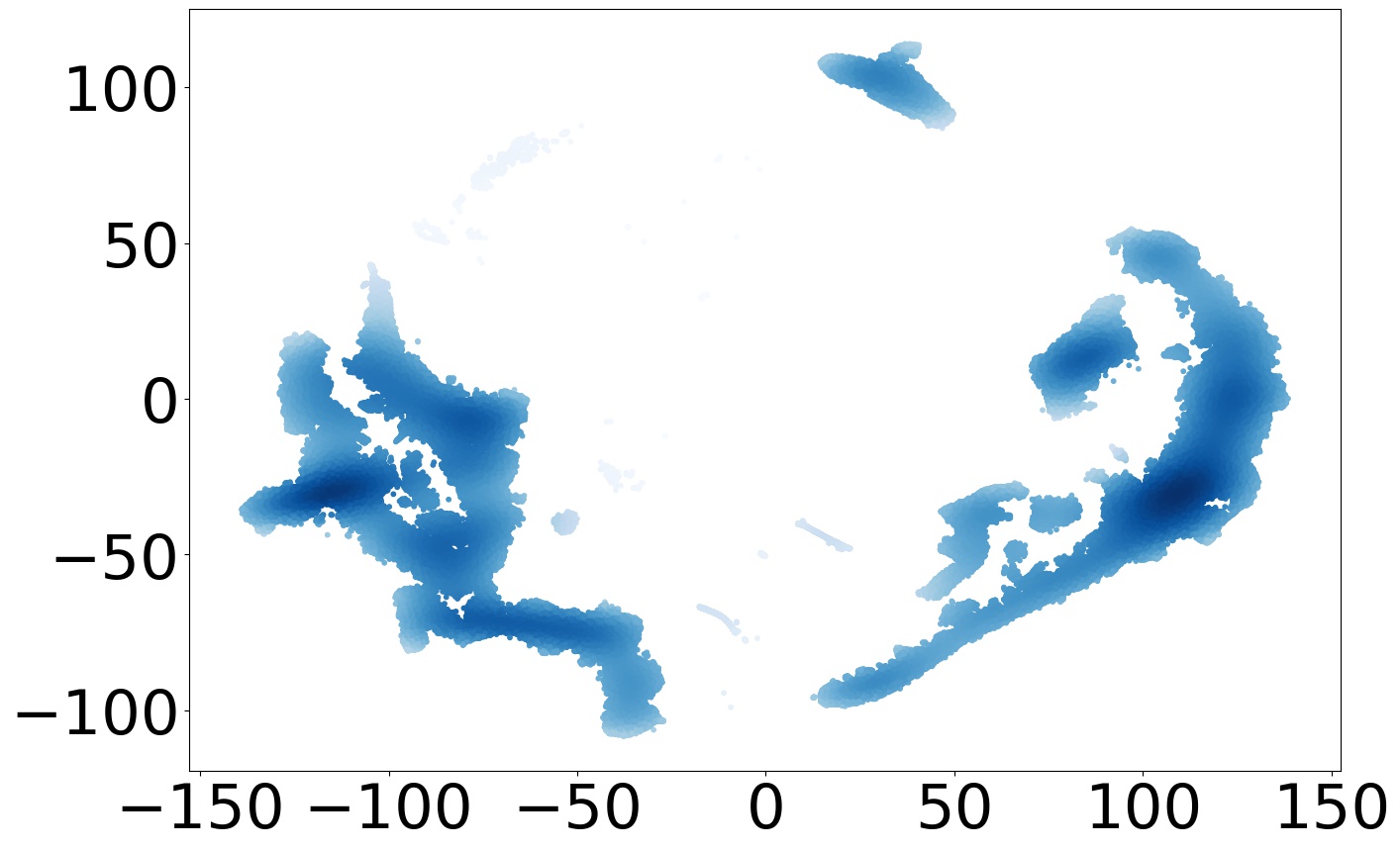}
		\includegraphics[width=1.5in, keepaspectratio]{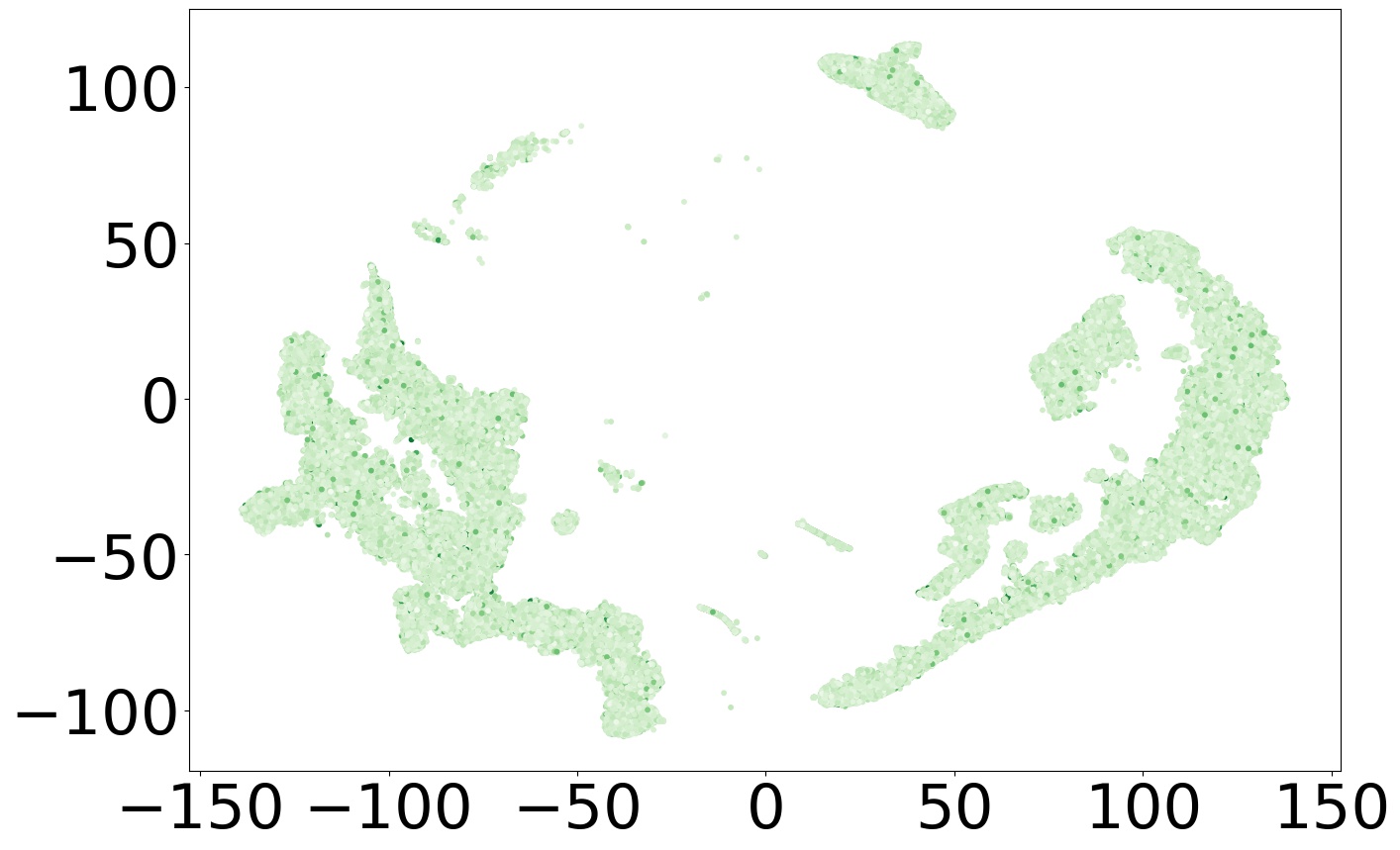}
		\includegraphics[width=1.5in, keepaspectratio]{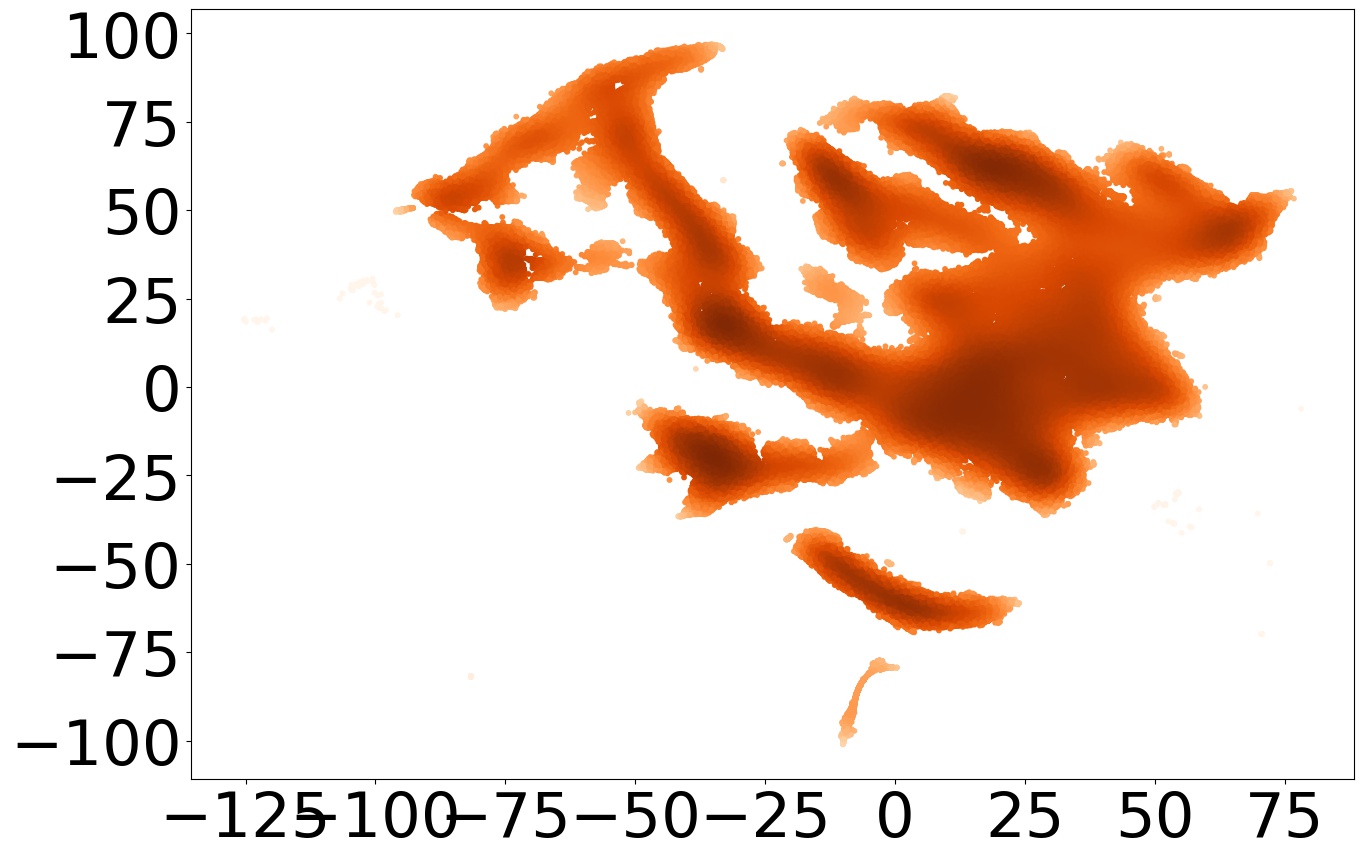}
		\includegraphics[width=1.5in, keepaspectratio]{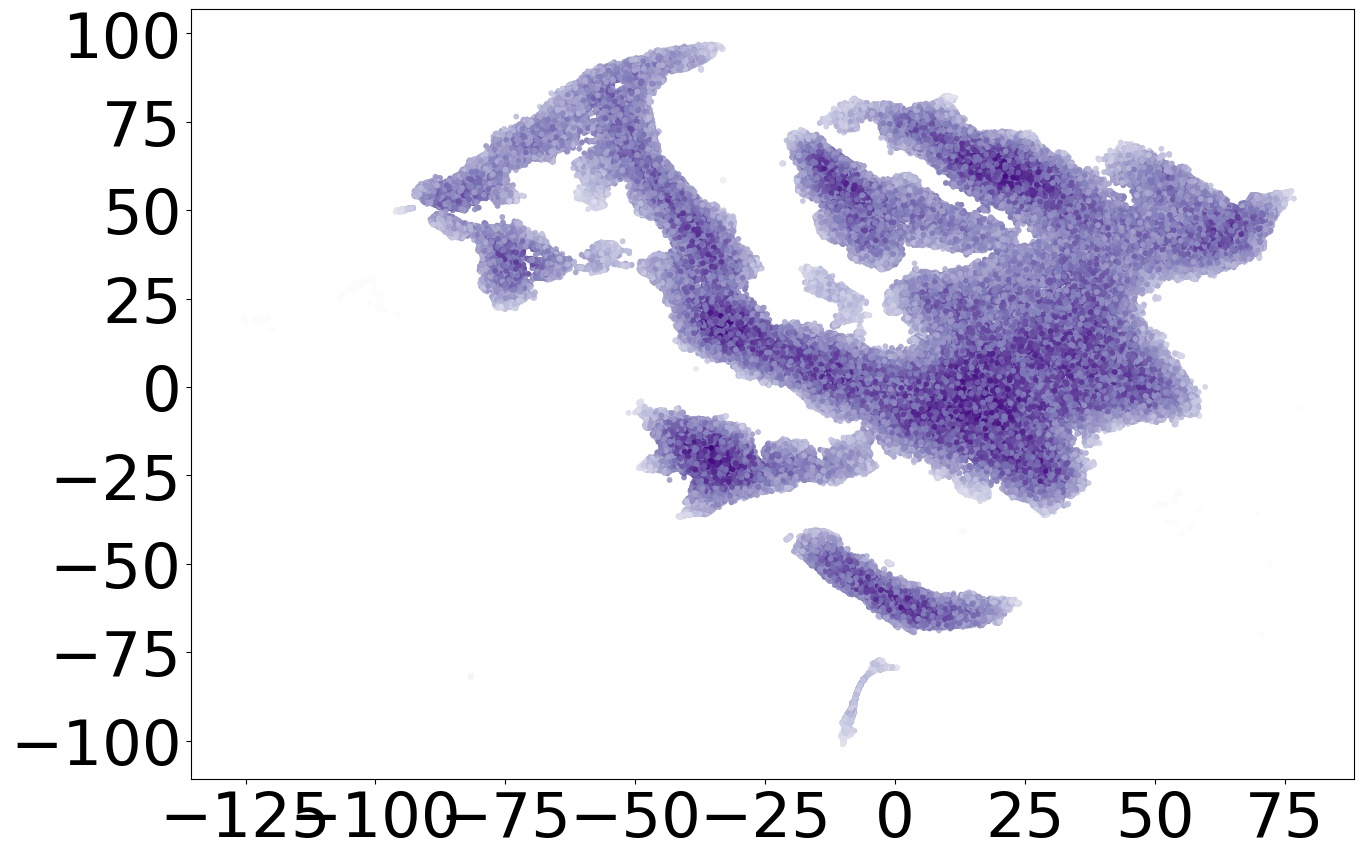}
	}
	\caption{State visitation density contours of the TD3 algorithm under greedy action selection and DISCOVER. Large temporal difference errors are represented by darker regions in the TD-error contours.}
	\label{fig:tsne_results}
\end{figure*}

We run the TD3 algorithm \citep{td3} without exploration and with Off-Policy DISCOVER over 1 million time steps in the Swimmer environment. We use a replay buffer \citep{experience_replay} of size 1 million transitions where random sampling is performed. Therefore, each visited state is accessible at the end of the training under both exploration settings. We plot the state visitation probabilities in Figure \ref{fig:tsne_results}. We employ Kernel Density Estimation based on the samples in 2D embedding space to learn the state visitation probabilities. The 2D embeddings are performed through the joint projection of the data sets onto 2D embedding space by t-SNE \citep{tsne}. We choose the Swimmer environment as its state space dimension is relatively small compared to the other environments, preventing possible artifacts in the low-dimensional visualization. Moreover, in Section \ref{sec:comp_eval}, we observe that actor-critic methods usually perform poorly in the Swimmer environment, which can be considered challenging \citep{deep_rl_that_matters}. To further reduce the possibility of projection error, we first perform PCA to reduce the state dimension from 8 to 4 prior to t-SNE \citep{tsne}. The PCA operation results in a proportion of variance explained value of 0.984 and 0.966 for no exploration and Off-Policy DISCOVER data sets, respectively.

% State Visitations Visualization

As shown in Figure \ref{fig:tsne_results}, we have three groups of visualization for each data set. Moreover, to give insights on the TD-error maximization in the exploration, we colorize the states with the TD-errors respective to their transitions. Thus, darker regions correspond to large absolute TD-errors. We group the state space visualizations into the states visited in the earlier, intermediate, and later stages of the training. In the earlier stages of the data collection, the greedy action selection spans the two sides of the state space. The agent nearly does not see the middle space throughout the training, and visited states settle to a smaller region on both sides of the state space. Furthermore, the TD-errors of the corresponding states are practically the same. This is an outcome of the greedy action selection in that the agent usually chooses the same actions and observes the same states, TD-error of which are constantly reduced by Q-learning \citep{watkins_1989}. 

However, our approach diversifies the state visitation distributions ranging from different modes of regions, compared to the greedy action selection. From the earlier to the later stages of data collection, the exploration policy learns how to direct the agent to state-action pairs with high TD-error, which we observe from the increasing similarity of the state visitation probabilities and TD-error heat map. Hence, this comparison indicates that our exploration policy is progressively learning a stationary policy that maximizes the TD-error and provides a robust exploration strategy to improve the underlying continuous control method.  

Our empirical studies also demonstrate an interesting insight. In all stages of the training, DISCOVER consistently visits almost complementary distinct states in the distribution space compared to the greedy action selection. We deduce that our exploration policy also learns how the agent greedily selects the actions and systematically directs it to explore diverse state regions based on this selection. This enables a consistent and continuous improvement in the agent's performance and designates our global and directed exploration strategy, which is notably different from noise-based exploration, which remains local. Finally, we obtain a considerable performance improvement by our exploration framework in this comparison, where the maximum return obtained by greedy and DISCOVER agents are 121.56 and 172.35, respectively. 

\section{Conclusion}
This paper introduces a general and scalable deep directed exploration framework that adapts the psychological animal behavior to continuous control deep reinforcement learning. Motivated by the gradients of the value functions, our method directs agents to less known state spaces with high temporal difference error. Our work shows that the introduced deep deterministic exploration policy can extract useful information from the collected experiences for exploration and combine the intrinsic motivations in the current literature into a single exploration objective. Furthermore, by theoretical justifications, our directed exploration framework is guaranteed to have a faster convergence than undirected strategies in any ergodic Markov Decision Process.

When combined with actor-critic algorithms, we empirically show that our strategy leads the agents to diverse state regions that are orthogonal to the greedy actions and outperforms the competing exploration methods by a significant margin in most of the tested tasks. Moreover, an extensive set of ablation studies demonstrate that when our exploration framework adopts the same update rule, hyper-parameter setting, and structure of the baselines' policy, it achieves maximum performance. Hence, our effective exploration strategy comes with the cost of training a single neural network and optimizing a single hyper-parameter which adjusts the exploration degree. Finally, the introduced exploration framework is readily applicable to various on- and off-policy actor-critic algorithms through a generic approach, and code to implement DISCOVER is publicly available at our repository\footref{our_repo} to ensure reproducibility. 

\bibliographystyle{unsrtnat}
\bibliography{references} 

\clearpage

\begin{appendices}

\section{Detailed Experimental Setup}
\label{app:imp_details}

\subsection{Software and Environment}

All networks are trained with PyTorch (version 1.8.1) \citep{pytorch}, using default values for all unmentioned hyper-parameters. Performances of all methods are evaluated in MuJoCo (mujoco-py version 1.50) \citep{mujoco}, and Box2D (version 2.3.10) \citep{box2d} physics engines interfaced by OpenAI Gym (version 0.17.3) \citep{gym}, using v3 environment for BipedalWalker and v2 for rest of the environments. The environment dynamics, state and action spaces, and reward functions are not pre-processed and modified for easy reproducibility and fair evaluation procedure with the baseline and competing algorithms. Each environment episode runs for a maximum of 1000 steps until a terminal condition is encountered. The multi-dimensional action space for all environments is within the range (-1, 1) except for Humanoid, which uses the range of (-0.4, 0.4).

\subsection{Evaluation}

All experiments are run for 1 million time steps with evaluations every 1000 time steps, where an evaluation of an agent records the average reward over 10 episodes without exploration noise and network updates. We utilize a new environment with a fixed seed (the training seed + a constant) for each evaluation to decrease the variation caused by different seeds. Therefore, each evaluation uses the same set of initial start states. 

We report the average evaluation return of 10 random seeds for each environment, including the initialization of behavioral policies, simulators, network parameters, and dependencies. Unless stated otherwise, each agent is trained by one training iteration after each time step. Agents are trained by batches of transitions uniformly sampled from the experience replay \citep{experience_replay}. Learning curves are used to show performance, and they are given as an average of 10 trials with a shaded zone added to reflect a half standard deviation across the trials. The curves are smoothed uniformly over a sliding window of 5 evaluations for visual clarity.

\subsection{Visualization of the State Visitations}

We visualize the states within the collected transitions for greedy action selection and DISCOVER under the TD3 algorithm \citep{td3} while learning in the Swimmer environment over 1 million time steps. The results are reported over a single seed. We consider the last 975000 transitions in the replay buffer \citep{experience_replay} as the first 25000 is sampled from the environment's action space. We first separately project the datasets onto a 4D state space through PCA to reduce the visual artifacts. 

Later, we jointly embed the resulting datasets onto the 2D state space through the t-SNE \citep{tsne} implementation of the openTSNE library \citep{open_tsne} \footnote{\url{https://opentsne.readthedocs.io/en/latest/}}. We use a perplexity value of 1396 and euclidean metric in measuring the distances between each state. The t-SNE \citep{tsne} is run over 1500 iterations. Default values in openTSNE \citep{tsne} is used for all unmentioned parameters. We split the datasets into three portions and separately visualize them, where each portion contains 325000 samples. The PCA operation yields a proportion of variance explained by the value of 0.984 and 0.966 for the greedy and DISCOVER datasets. 

\subsection{Implementation}

Our implementation of A2C \citep{a2c} and PPO \citep{ppo} is based on the code from the well-known repository\footref{ppo_repo}, following the tuned hyper-parameters for the considered continuous control tasks. For the implementation of TD3 \citep{td3}, we use the author's GitHub repository\footref{td3_repo} for the fine-tuned version of the algorithm and DDPG \citep{ddpg} implementation. For SAC \citep{sac}, we follow structure outlined in the original paper. 

% DISCOVER adaptations follow the pseudocodes provided in Algorithm \ref{alg:discover_ddpg}, \ref{alg:discover_sac} and \ref{alg:discover_td3}.

We implement NoisyNet \citep{noisy_net} by adapting the code from authors' GitHub repository\footref{noisy_net_repo} to the baseline actor-critic algorithms. Authors' OpenAI Baselines implementation\footref{baselines_repo} \citep{openai_baselines} is used to implement PSNE. Similar to SAC \citep{sac}, we refer to the original papers in implementing Deep Coherent Exploration \citep{deep_coherent_exploration} and Meta-DDPG \citep{meta_exploration} as the authors did not provide a valid code repository. 

\subsection{Architecture and Hyper-Parameter Setting}
The on-policy methods, A2C \citep{a2c} and PPO \citep{ppo} follow the tuned hyper-parameters for the MuJoCo \citep{mujoco} and Box2D \citep{box2d} tasks provided by the repository\footref{ppo_repo}. Our implementation of the off-policy actor-critic algorithms, DDPG \citep{ddpg}, SAC \citep{sac} and TD3 \citep{td3}, closely follows the set of hyper-parameters given in the respective papers. For DDPG \citep{ddpg} and SAC \citep{sac}, we use the fine-tuned environment-specific hyper-parameters provided by the OpenAI Baselines3 Zoo\footref{zoo_repo} \citep{openai_zoo}. TD3 \citep{td3} uses the fine-tuned parameters provided in the author's GitHub repository\footref{td3_repo}. Shared, and environment and algorithm specific hyper-parameters for the off-policy methods are given in Table \ref{table:shared_parameters_off}, \ref{table:algo_specific_parameters_off} and \ref{table:sac_spec_parameters}. Additionally Table \ref{table:shared_parameters_on} and \ref{table:algo_specific_parameters_on} reports the shared and algorithm specific tuned hyper-parameters for the on-policy baselines, respectively. Note that entropy coefficient used for A2C \citep{a2c} and PPO \citep{ppo} is 0, meaning that there is no entropy maximization and hence, inherent exploration. Nevertheless, the tuned 0 value corresponds to greedy action selection, which is found to perform best.

For the parameter-space noise algorithms, we initialize the parameter noise at 0.017 and 0.034, which gives the best results in practice for on- and off-policy algorithms, respectively, as suggested in \citep{deep_coherent_exploration}. Furthermore, we use $\beta = 0.01$ for all environments in the Deep Coherent Exploration algorithm \citep{deep_coherent_exploration}, and set the mean-squared error threshold in PSNE \citep{psne} to 0.1. 

The exploration framework of DISCOVER strictly follows the actor framework in the corresponding baseline algorithms. This includes the depth and size of the networks, learning rate, optimizer, nonlinearity, target, and behavioral policy update frequency, target network learning rate, and the number of gradient steps in the updates. Moreover, we still use the exploration policy in the exploration time steps at the beginning of each training. 

\begin{table}[!hbt]
\begin{center}
\caption{Shared hyper-parameters of the baseline off-policy actor-critic algorithms.}
\label{table:shared_parameters_off}
    \begin{tabular}{@{} lc @{}}
    \toprule
    \textbf{Hyper-parameter} & \textbf{Value} \\
    \midrule
        Actor regularization & None \\
        Optimizer & Adam \citep{adam} \\
        Nonlinearity & ReLU \\
        Gradient clipping & False \\
        Number of hidden layers (all networks) & 2 \\
    \bottomrule
    \end{tabular}
\end{center}
\end{table}

\begin{table}[!hbt]
\begin{center}
\caption{Algorithm specific hyper-parameters used for the implementation of the baseline on-policy actor-critic algorithms.}
\label{table:algo_specific_parameters_on}
\begin{threeparttable}
    \begin{tabular}{@{} lccc @{}}
        \toprule
        \textbf{Hyper-parameter} & \textbf{A2C} & \textbf{PPO} \\ 
        \midrule
        Learning rate\tnote{a} & 0.0013 & 0.0003 \\
        Regularization\tnote{a} & None & None \\ 
        Explorer learning rate & 0.0013 & 0.0003 \\
        Optimizer & RMSProp & Adam \\
        Number of rollout steps (horizon) & 32 & 2048 \\
        Number of mini-batches\tnote{b} & 1 & 32 \\
        GAE\tnote{c} & False & True \\
        GAE $\lambda$ & None & 0.95 \\
        Entropy coefficient & 0.0 & 0.0 \\
        Linear decay\tnote{d} & False & True \\ 
        \bottomrule
    \end{tabular}
    \begin{tablenotes}    
        \item[a]{Used for both value and actor networks}
        \item[b]{The mini-batch size can be computed by:\\\texttt{\# of rollout steps / \# of mini-batches}}
        \item[c]{Usage of General Advantage Estimation \citep{gae}}
        \item[d]{Usage of linear decay on the learning rate}
    \end{tablenotes}
\end{threeparttable}
\end{center}
\end{table}

\begin{table}[!hbt]
\begin{center}
\caption{SAC specific hyper-parameters.}
\label{table:sac_spec_parameters}
\begin{tabular}{@{} lcc @{}}
    \toprule
    \textbf{Hyper-parameter} & \textbf{Value} \\ 
    \midrule
        Entropy target ($\tau$) & \texttt{-action dimensions} \\
        Temperature parameter ($\alpha$) & See\footref{zoo_repo} \\
        Reward scale & See \citep{sac} and See\footref{zoo_repo} \\
        Standard deviation clipping & (-20, 2) \\
        $\epsilon$ & $10^{-6}$ \\
    \bottomrule
    \end{tabular}
\end{center}
\end{table}

\begin{table}[!hbt]
\begin{center}
\caption{Algorithm specific hyper-parameters used for the implementation of the baseline off-policy actor-critic algorithms.}
\label{table:algo_specific_parameters_off}
    \begin{tabular}{@{} lcccc @{}}
        \toprule
        \textbf{Hyper-parameter} & \textbf{DDPG} & \textbf{SAC} & \textbf{TD3} \\ 
        \midrule
        Critic learning rate & $10^{-3}$ & $3 \times 10^{-4}$ & $3 \times 10^{-4}$ \\
        Critic regularization & $10^{-2} \times ||\theta||^{2}$ & None & None \\
        Actor learning rate & $10^{-4}$ & $3 \times 10^{-4}$ & $3 \times 10^{-4}$ \\
        Actor regularization & None & None & None \\
        Explorer learning rate & $10^{-4}$ & $3 \times 10^{-4}$ & $3 \times 10^{-4}$ \\
        Policy type & Deterministic & Gaussian & Deterministic \\
        Discount factor ($\gamma$) & 0.99  & See\footref{zoo_repo} & 0.99 \\
        Target update rate ($\tau$) & $10^{-3}$ & See\footref{zoo_repo} & $5 \times 10^{-3}$ \\
        Batch size & 64 & 256 & 256 \\
        Critic update period & 1 & See\footref{zoo_repo} & 1 \\
        Actor update period & 1 & See\footref{zoo_repo} & 2 \\
        Explorer update period & 1 & See\footref{zoo_repo} & 2 \\
        Critic updates per period  & 1 & See\footref{zoo_repo} & 1 \\
        Actor updates per period & 1 & See\footref{zoo_repo} & 1 \\
        Explorer updates per period & 1 & See\footref{zoo_repo} & 1 \\
        Normalized observations & True & False & False \\
        Gaussian exploration & $\mathcal{N}(0, 0.1)$ & $\mathcal{N}(0, 0.1)$ & $\mathcal{N}(0, 0.1)$ \\
        Exploration time steps & See\footref{zoo_repo} & See\footref{zoo_repo} & 25000 \\
        $1^{\mathrm{st}}$ layer hidden units & 400 & 256 & 256 \\
        $2^{\mathrm{nd}}$ layer hidden units & 300 & 256 & 256 \\
        \bottomrule
    \end{tabular}
\end{center}
\end{table}

\begin{table}[!hbt]
\begin{center}
\caption{Shared hyper-parameters of the baseline on-policy actor-critic algorithms.}
\label{table:shared_parameters_on}
    \begin{tabular}{@{} lc @{}}
    \toprule
    \textbf{Hyper-parameter} & \textbf{Value} \\
    \midrule
        Actor regularization & None \\
        Policy type & Diagonal Gaussian \\
        Update period & After each horizon \\ 
        Updates per period & Number of mini-batch sizes \\
        Explorer update period & After each horizon \\  
        Explorer updates per period & Number of mini-batch sizes \\
        Normalized observations & True \\
        Optimizer $\epsilon$ & 0.00001 \\
        Optimizer $\alpha$ & 0.99 \\
        Nonlinearity & Tanh \\
        Discount factor ($\gamma$) & 0.99 \\
        Value loss coefficient & 0.5 \\
        Maximum gradient norm & 0.5 \\ 
        Number of hidden layers (all networks) & 2 \\
        $1^{\mathrm{st}}$ layer hidden units & 64 \\
        $2^{\mathrm{nd}}$ layer hidden units & 64 \\
    \bottomrule
    \end{tabular}
\end{center}
\end{table}

\clearpage

\section{Complete Experimental Results}
\subsection{Additional Evaluation Results}
\label{app:comp_eval_results}

% Additional Evaluation Results - A2C
\begin{figure*}[!hbt]
    \centering
    \begin{align*}
        &\text{{\blue} DISCOVER-A2C ($\lambda$ = 0.1)}  &&\text{{\orange} Coherent-A2C ($\beta$ = 0.01)} &&\text{{\purple} NoisyNet-A2C} \\ &\text{{\green} PSNE-A2C} &&\text{{\red} A2C}
    \end{align*}
	\subfigure{
		\includegraphics[width=2.0in, keepaspectratio]{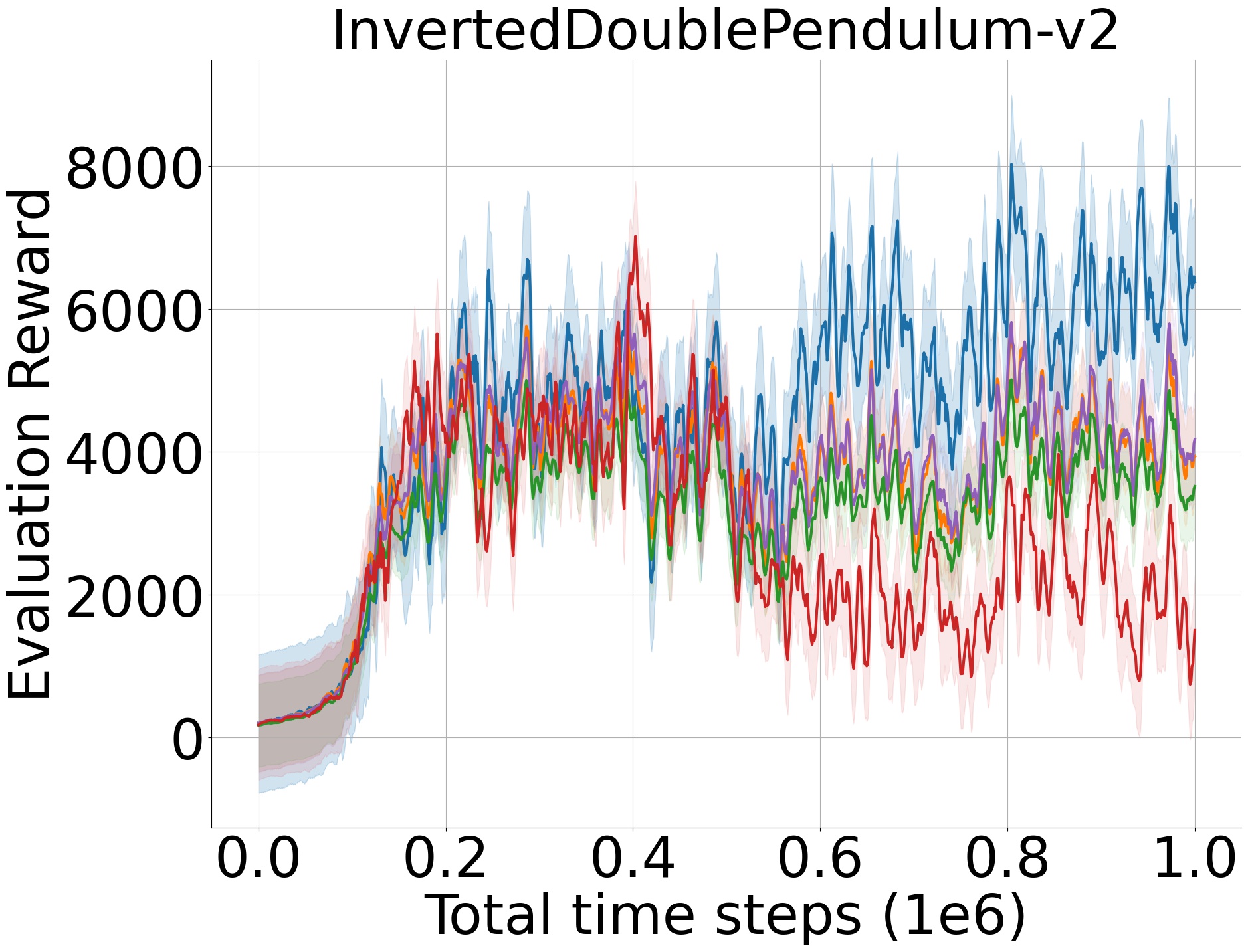}
		\includegraphics[width=2.0in, keepaspectratio]{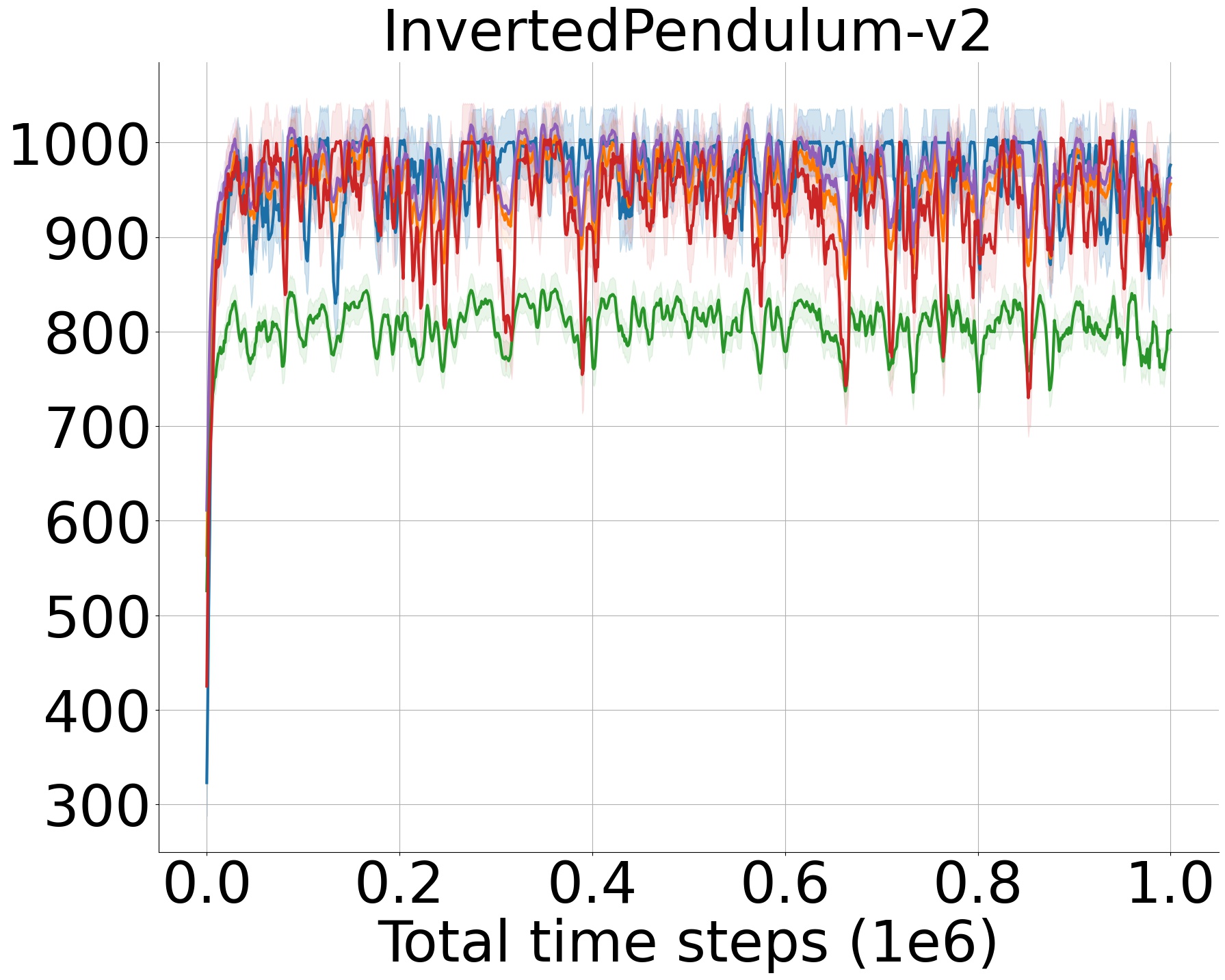}
		\includegraphics[width=2.0in, keepaspectratio]{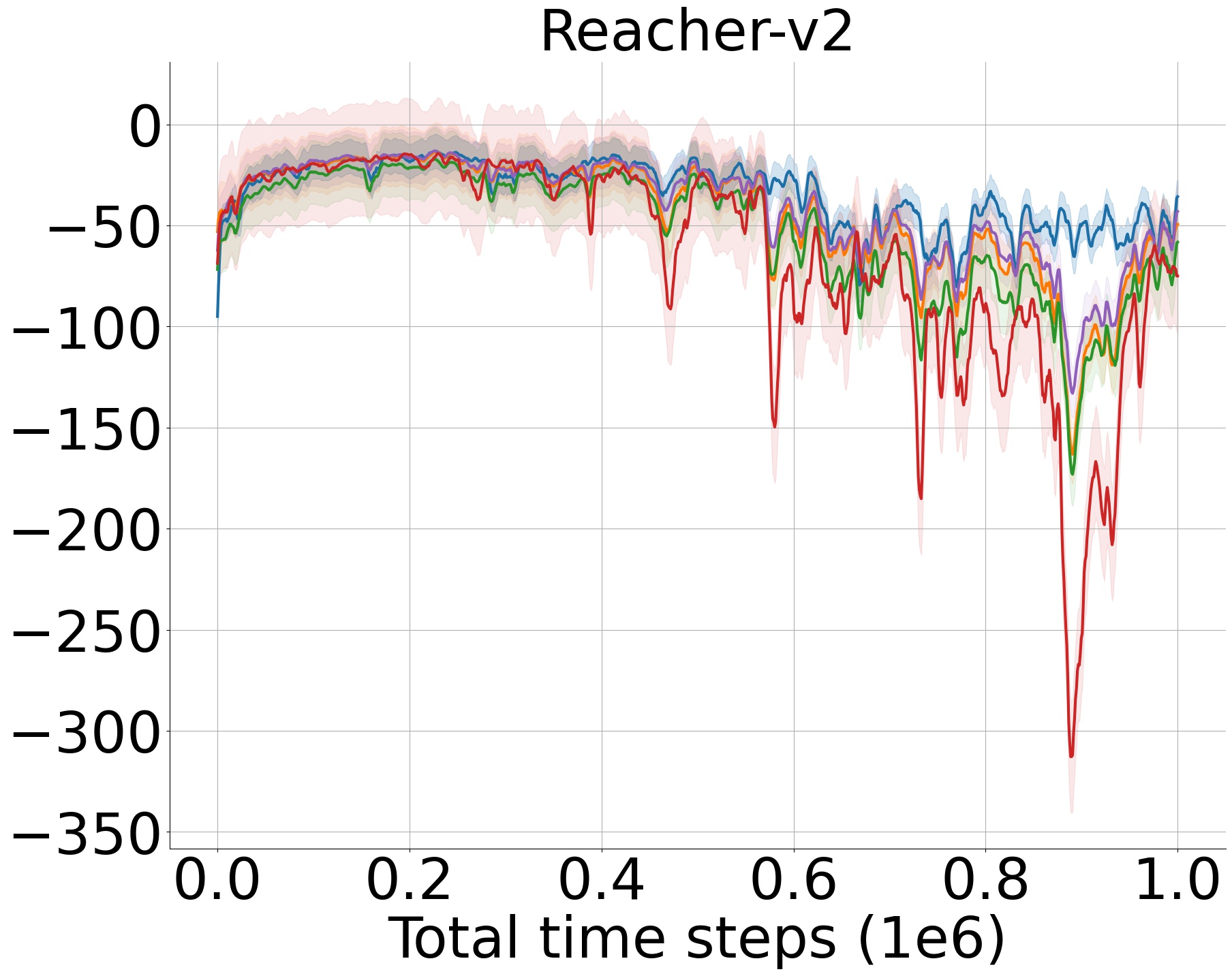}
	}
	\caption{Additional evaluation curves for the set of MuJoCo continuous control tasks under the A2C algorithm. The shaded region represents half a standard deviation of the average evaluation return over 10 random seeds. A sliding window of size 5 smoothes curves for visual clarity.}
\end{figure*}

% Additional Evaluation Results - PPO
\begin{figure*}[!hbt]
    \centering
    \begin{align*}
        &\text{{\blue} DISCOVER-PPO ($\lambda$ = 0.1)}  &&\text{{\orange} Coherent-PPO ($\beta$ = 0.01)} &&\text{{\purple} NoisyNet-PPO} \\ &\text{{\green} PSNE-PPO} &&\text{{\red} PPO}
    \end{align*}
	\subfigure{
		\includegraphics[width=2.0in, keepaspectratio]{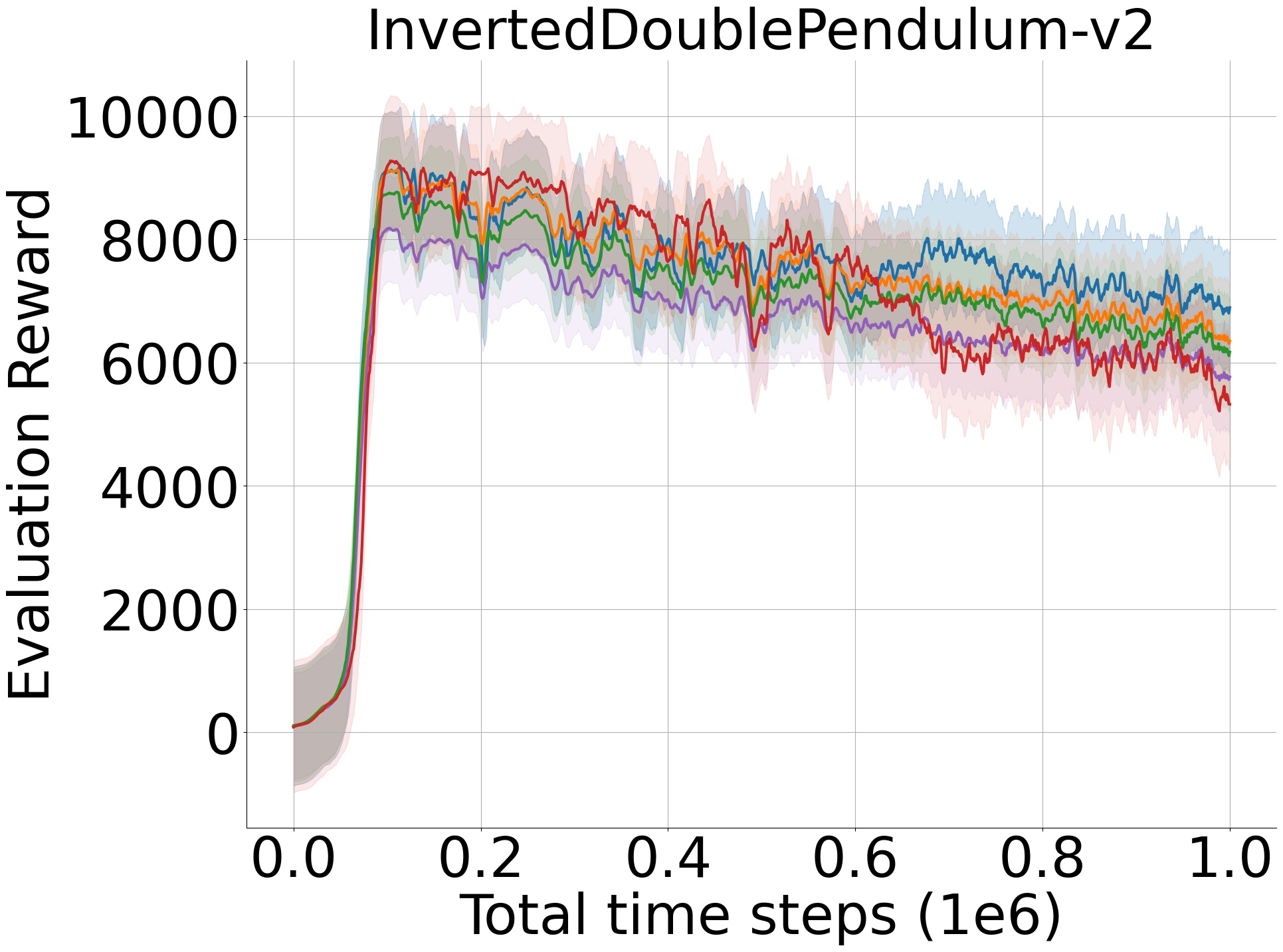}
		\includegraphics[width=2.0in, keepaspectratio]{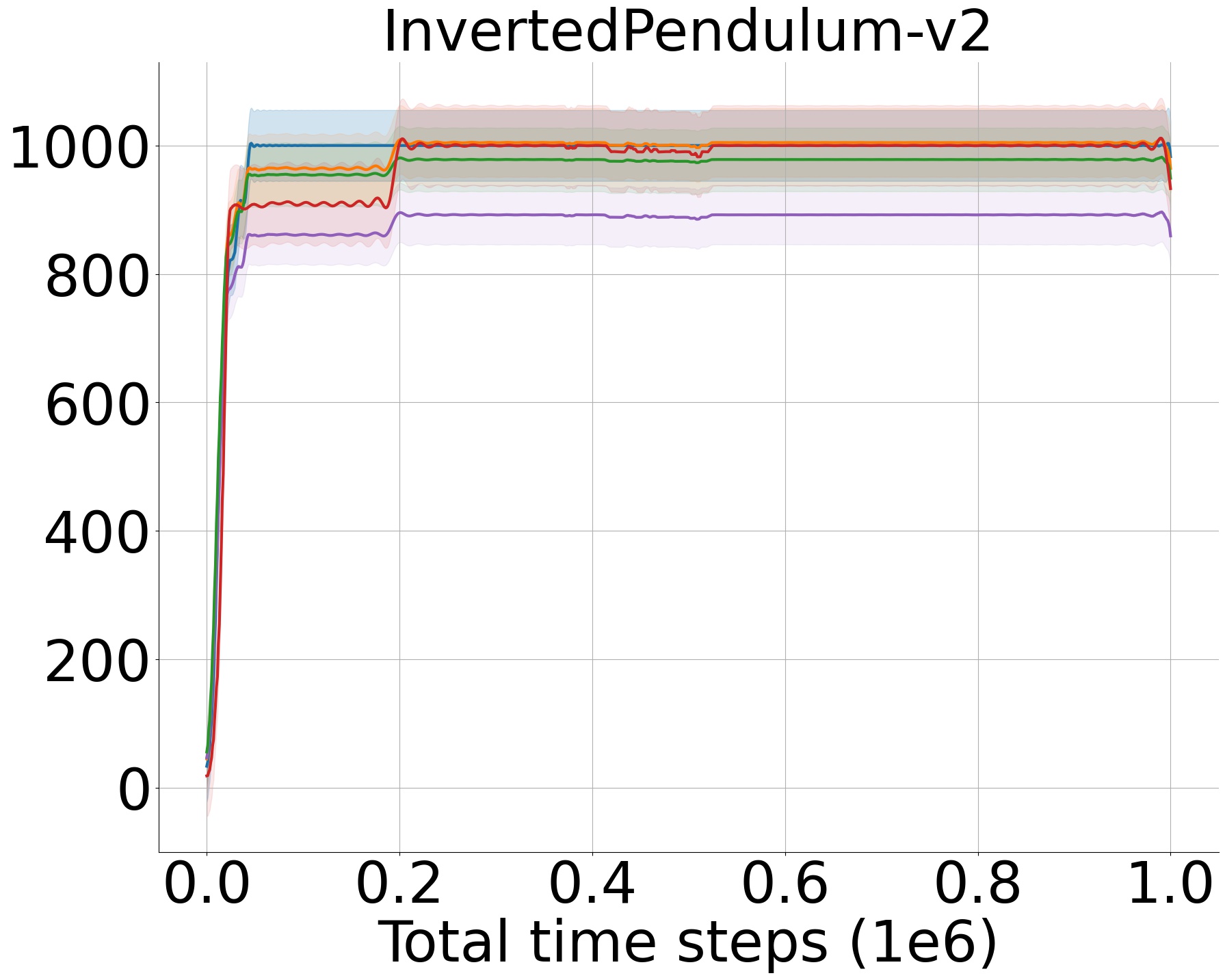}
		\includegraphics[width=2.0in, keepaspectratio]{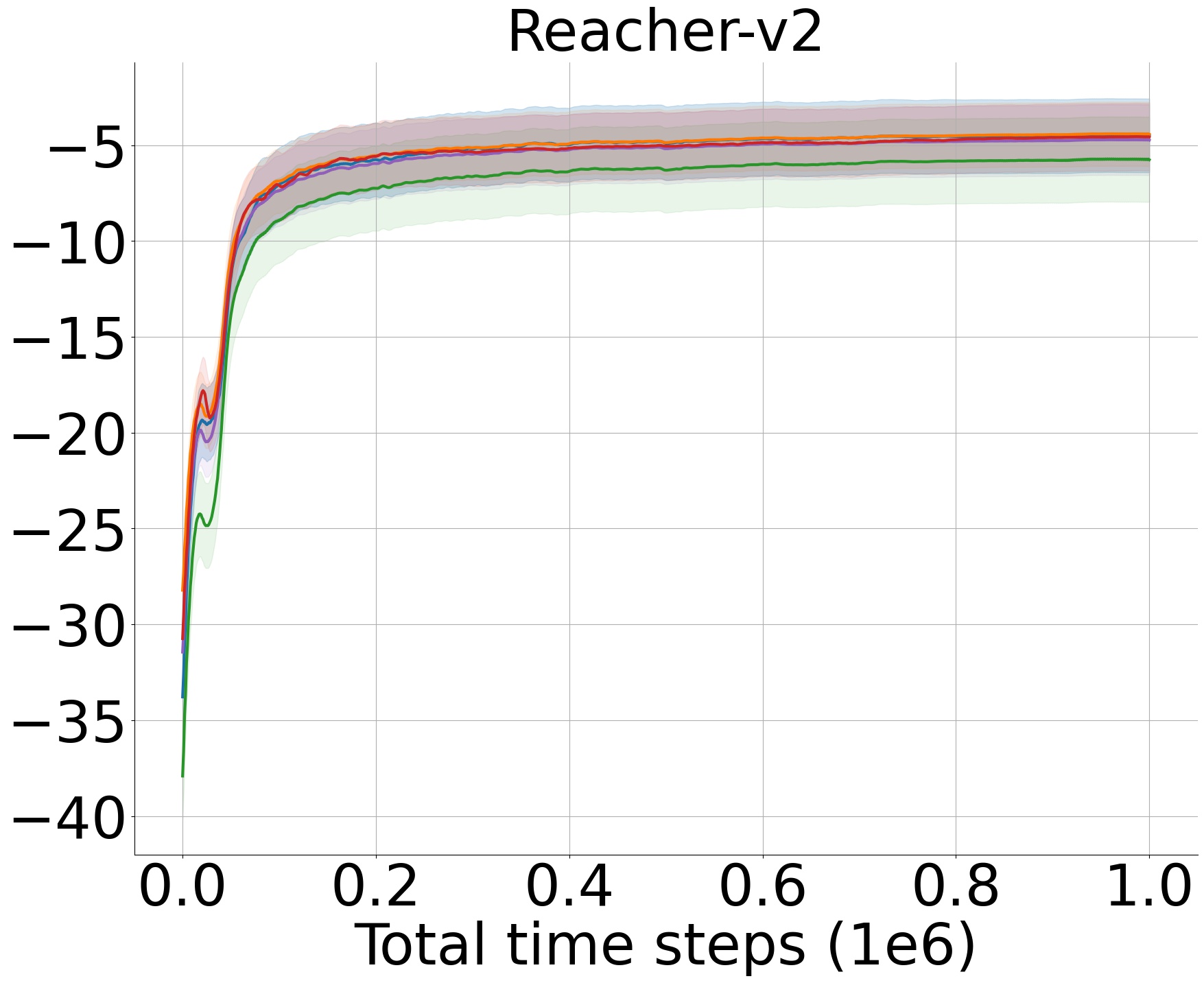}
	}
	\caption{Additional evaluation curves for the set of MuJoCo continuous control tasks under the PPO algorithm. The shaded region represents half a standard deviation of the average evaluation return over 10 random seeds. A sliding window of size 5 smoothes curves for visual clarity.}
\end{figure*}

% Additional Evaluation Results - DDPG
\begin{figure*}[!hbt]
    \centering
    \begin{align*}
        &\text{{\blue} DISCOVER-DDPG ($\lambda$ = 0.3)}  &&\text{{\orange} Coherent-DDPG ($\beta$ = 0.01)} &&\text{{\purple} NoisyNet-DDPG} \\ &\text{{\green} PSNE-DDPG} &&\text{{\red} Gaussian-DDPG} &&\text{{\brown} Greedy-DDPG}
    \end{align*}
	\subfigure{
		\includegraphics[width=2.0in, keepaspectratio]{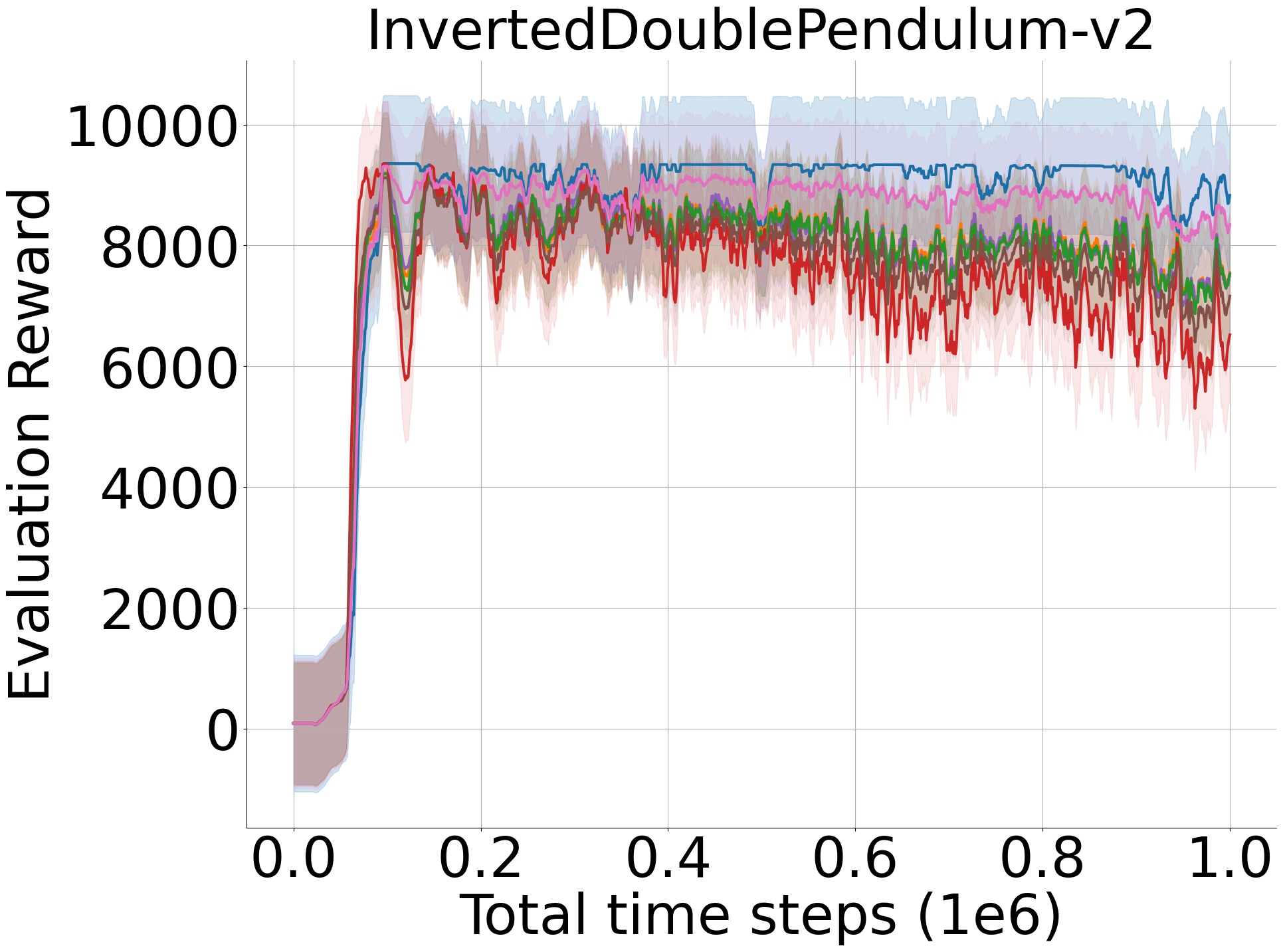}
		\includegraphics[width=2.0in, keepaspectratio]{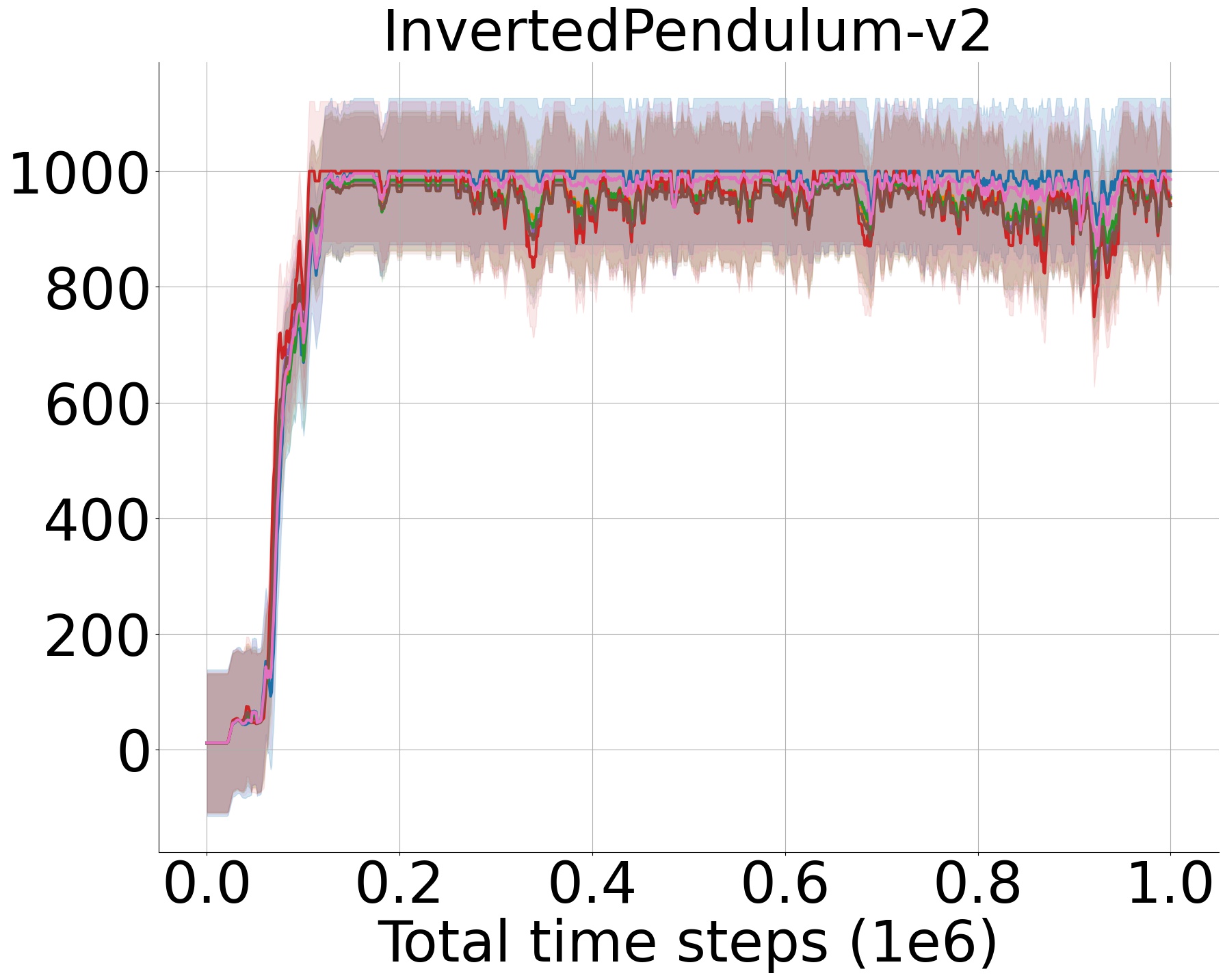}
		\includegraphics[width=2.0in, keepaspectratio]{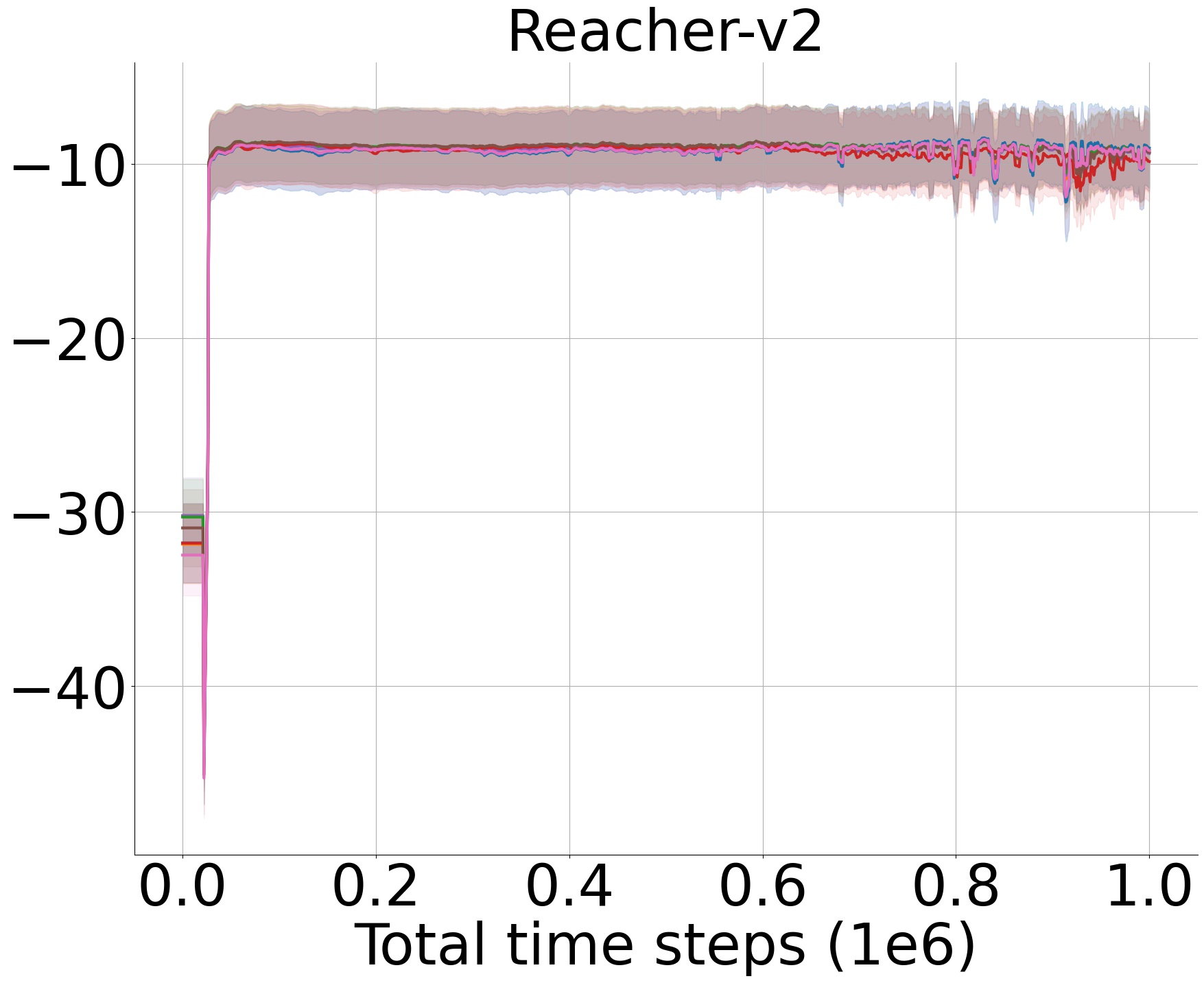}
	}
	\caption{Additional evaluation curves for the set of MuJoCo continuous control tasks under the DDPG algorithm. The shaded region represents half a standard deviation of the average evaluation return over 10 random seeds. A sliding window of size 5 smoothes curves for visual clarity.}
\end{figure*}

% Additional Evaluation Results - SAC
\begin{figure*}[!hbt]
    \centering
    \begin{align*}
        &\text{{\blue} DISCOVER-SAC ($\lambda$ = 0.3)}  &&\text{{\orange} Coherent-SAC ($\beta$ = 0.01)} &&\text{{\purple} NoisyNet-SAC} \\ &\text{{\green} PSNE-SAC} &&\text{{\red} Gaussian-SAC} &&\text{{\brown} Greedy-SAC}
    \end{align*}
	\subfigure{
		\includegraphics[width=2.0in, keepaspectratio]{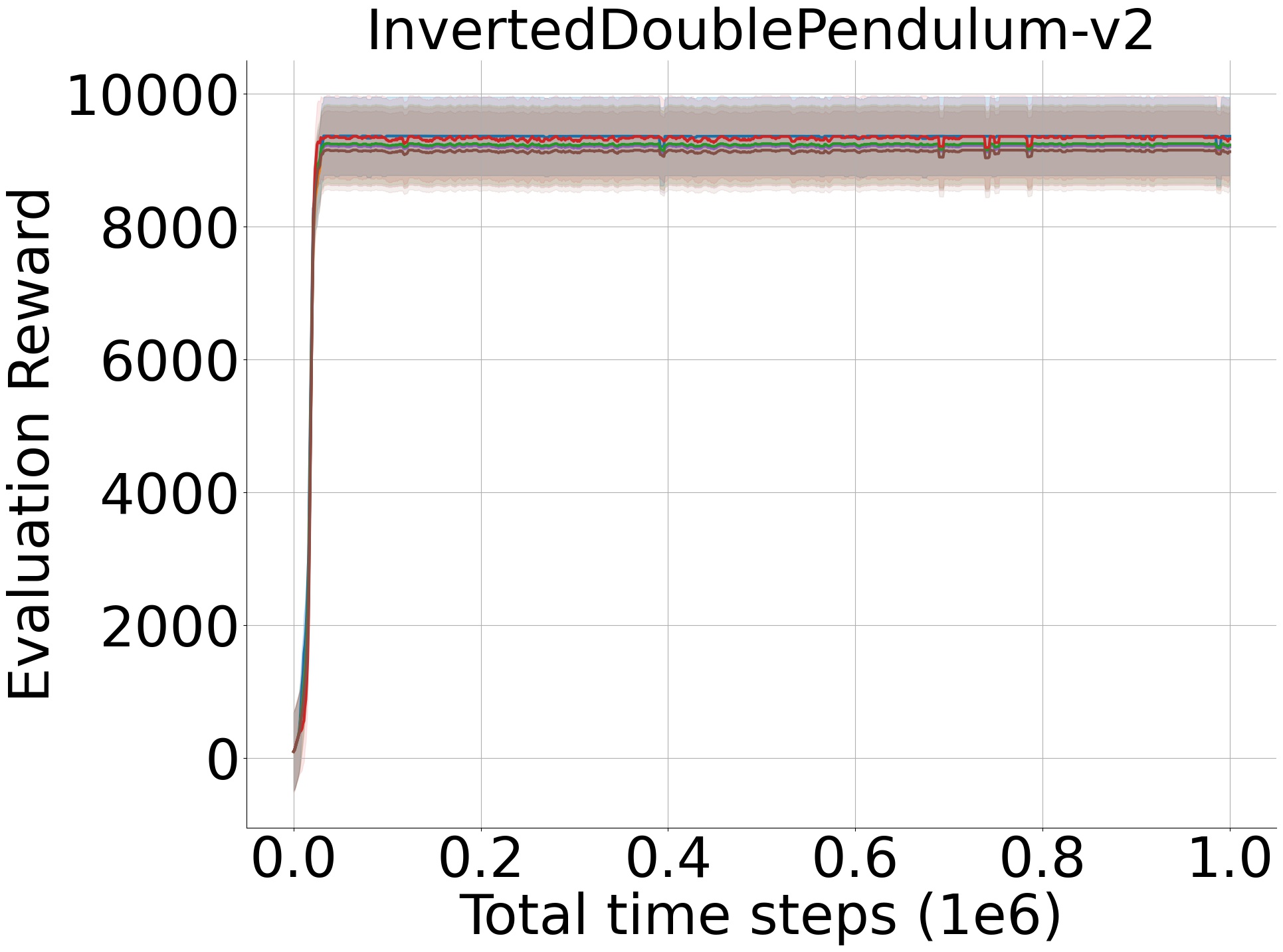}
		\includegraphics[width=2.0in, keepaspectratio]{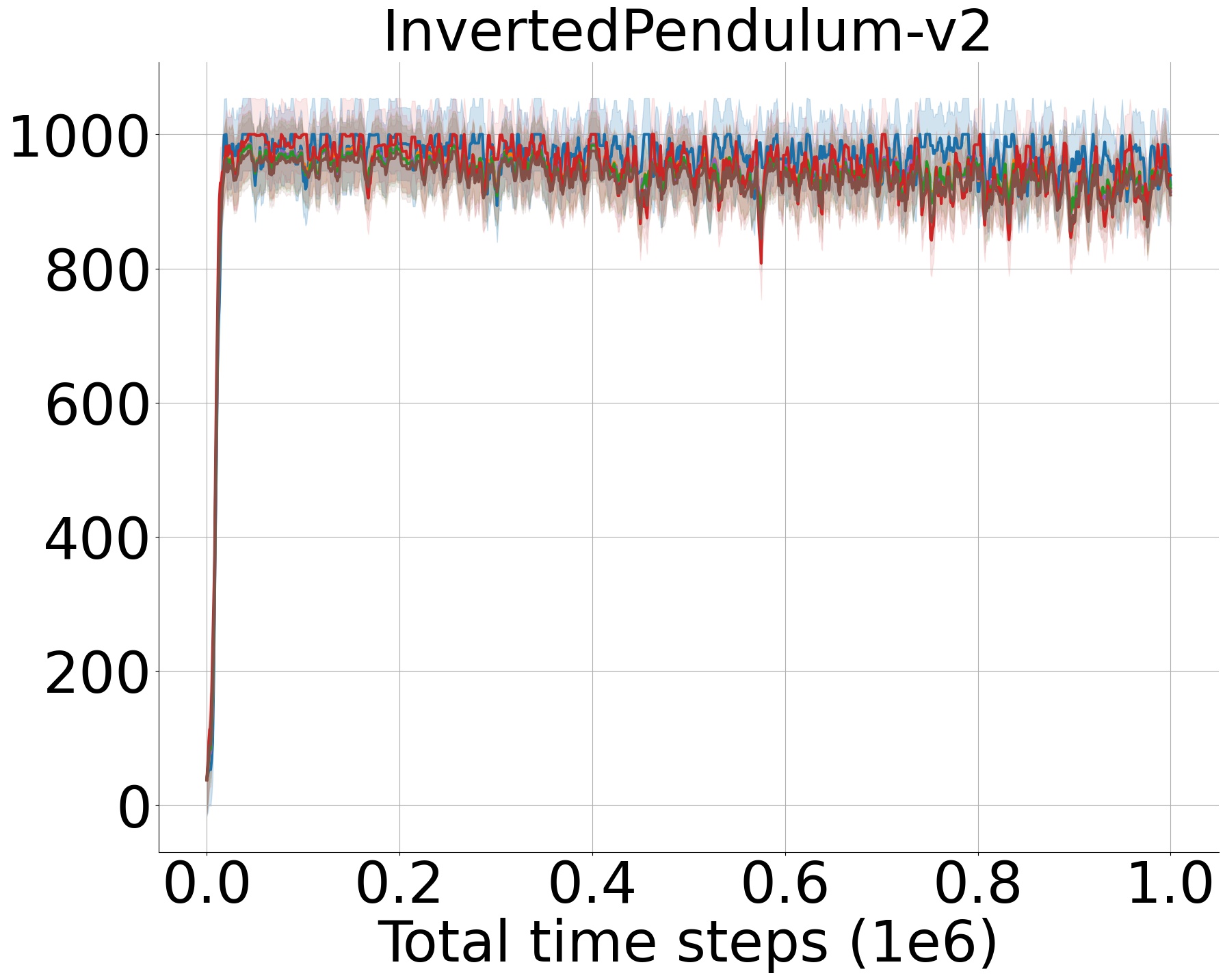}
		\includegraphics[width=2.0in, keepaspectratio]{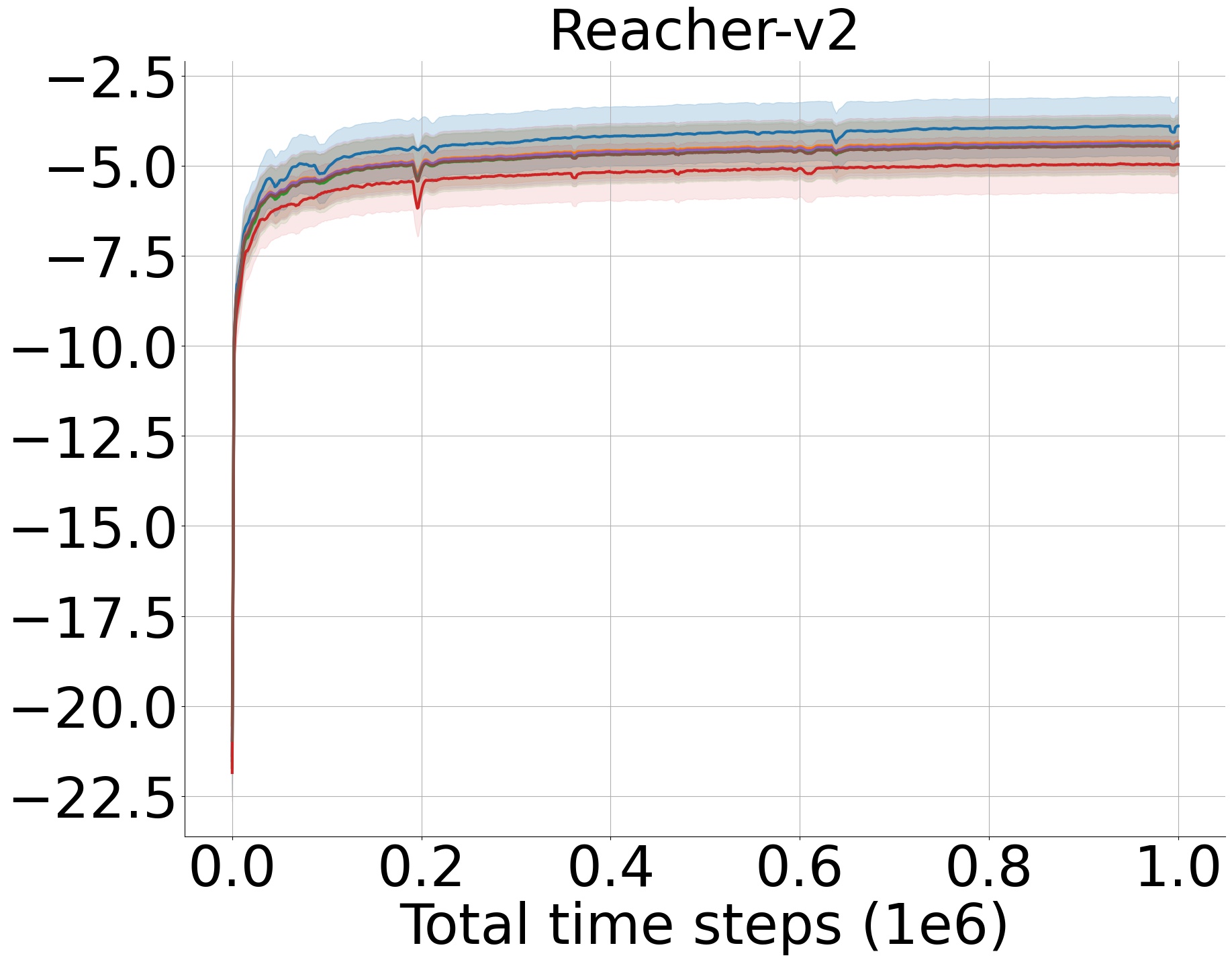}
	}
	\caption{Additional evaluation curves for the set of MuJoCo continuous control tasks under the SAC algorithm. The shaded region represents half a standard deviation of the average evaluation return over 10 random seeds. A sliding window of size 5 smoothes curves for visual clarity.}
\end{figure*}

% Additional Evaluation Results - TD3
\begin{figure*}[!hbt]
    \centering
    \begin{align*}
        &\text{{\blue} DISCOVER-TD3 ($\lambda$ = 0.3)}  &&\text{{\orange} Coherent-TD3 ($\beta$ = 0.01)} &&\text{{\purple} NoisyNet-TD3} \\ &\text{{\green} PSNE-TD3} &&\text{{\red} Gaussian-TD3} &&\text{{\brown} Greedy-TD3}
    \end{align*}
	\subfigure{
		\includegraphics[width=2.0in, keepaspectratio]{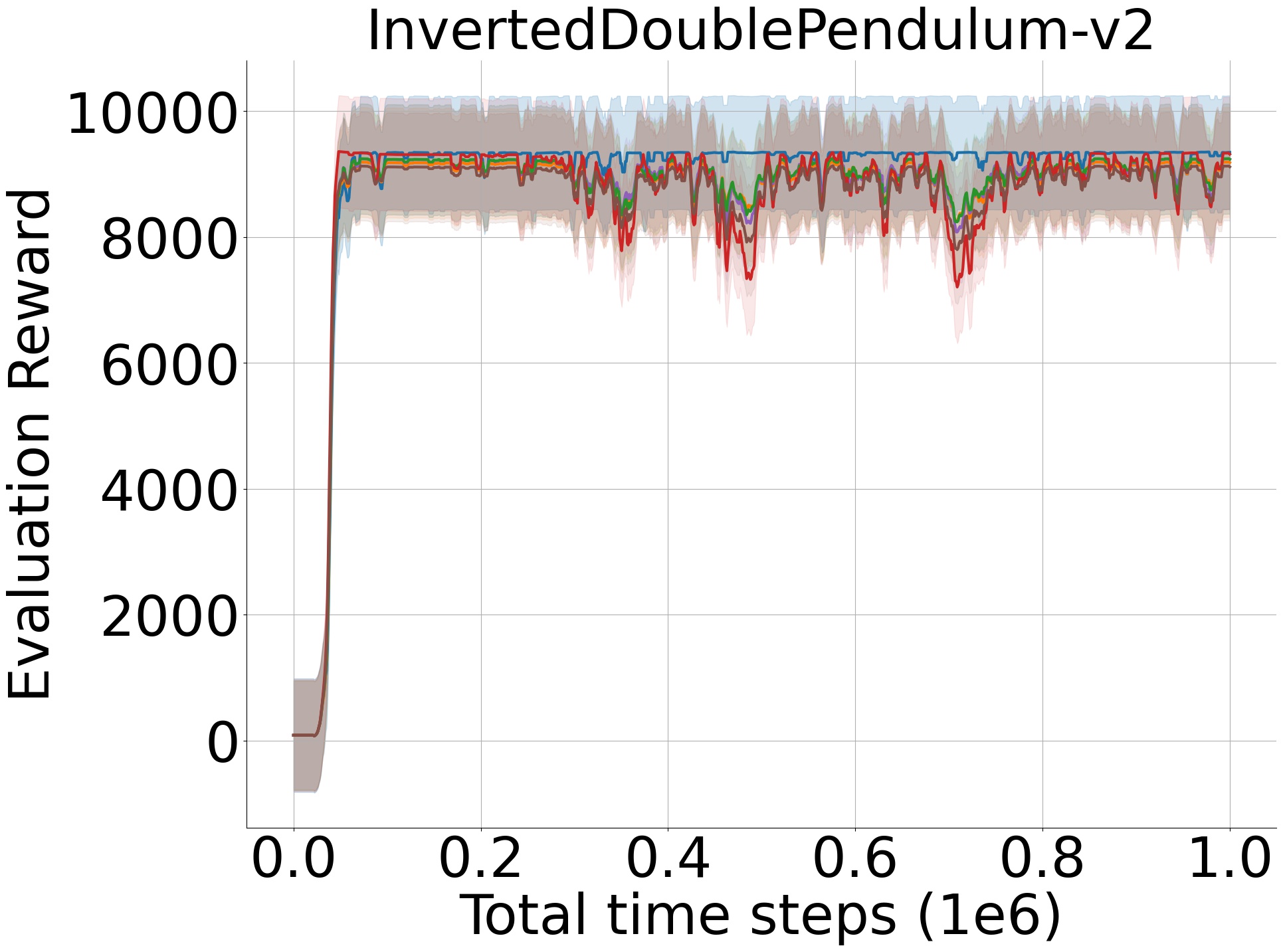}
		\includegraphics[width=2.0in, keepaspectratio]{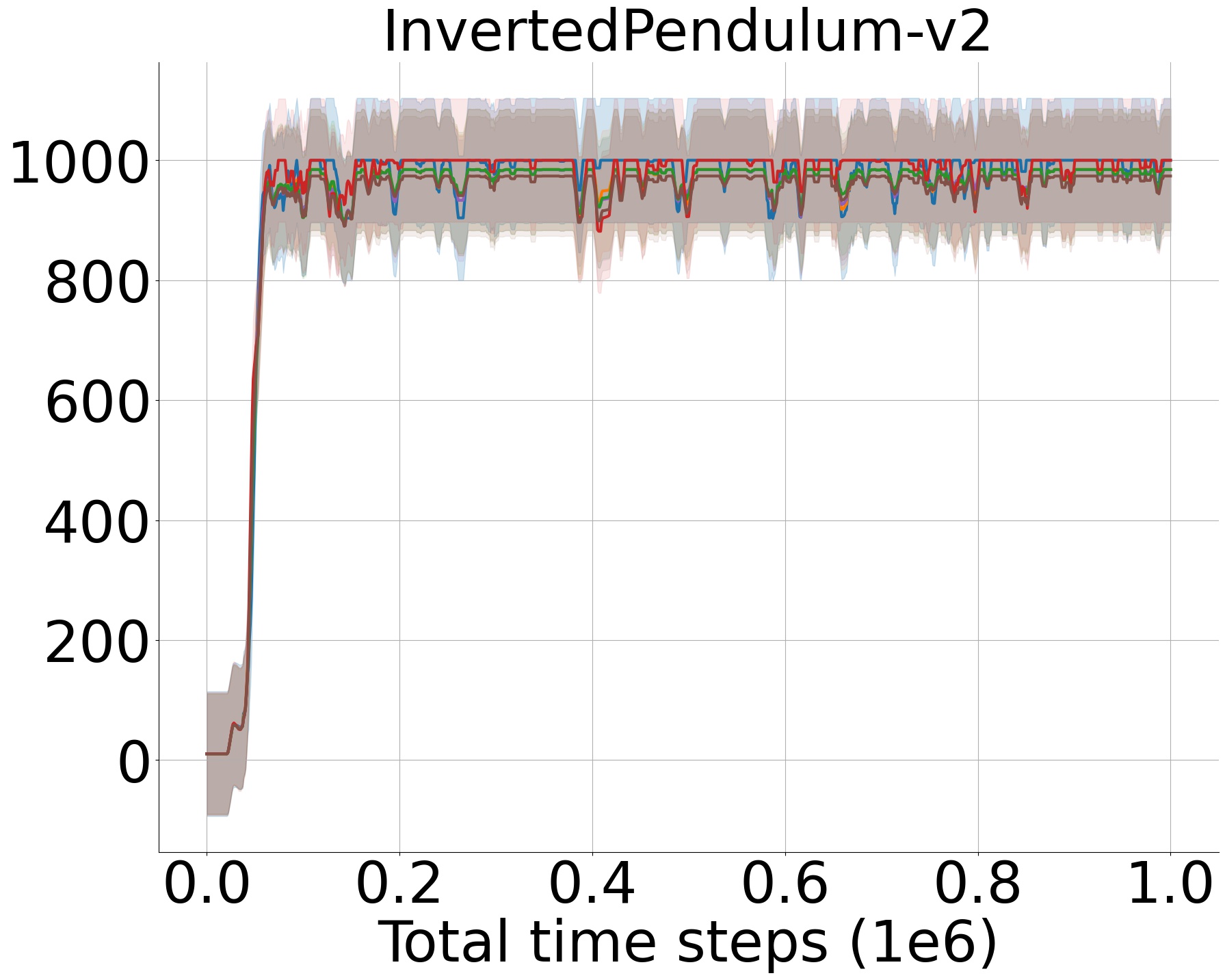}
		\includegraphics[width=2.0in, keepaspectratio]{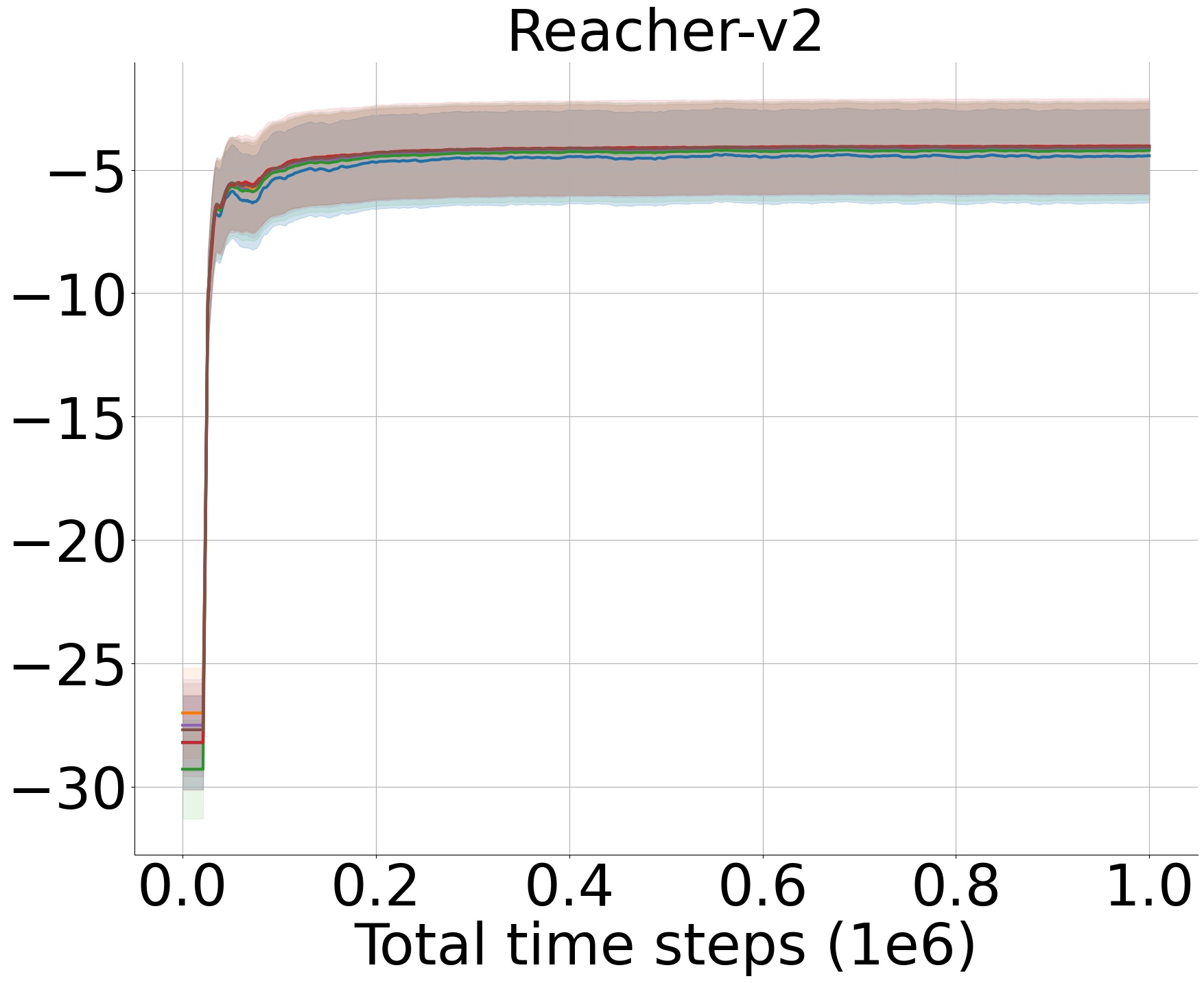}
	}
	\caption{Additional evaluation curves for the set of MuJoCo continuous control tasks under the TD3 algorithm. The shaded region represents half a standard deviation of the average evaluation return over 10 random seeds. A sliding window of size 5 smoothes curves for visual clarity.}
\end{figure*}

\clearpage

\subsection{Learning Curves for the Ablation Studies}\label{app:ab_study_learning_curves}

\subsubsection{Exploration Direction Regularization}
\begin{figure*}[!hbt]
    \centering
    \begin{align*}
        &\text{{\blue} $\lambda = 0.0$} &&\text{{\orange} $\lambda = 0.1$} &&\text{{\green} $\lambda = 0.3$} \\
        &\text{{\red} $\lambda = 0.6$} &&\text{{\purple} $\lambda = 0.9$} && \text{{\brown} $\lambda = 1.0$}
    \end{align*}
	\subfigure{
		\includegraphics[width=1.5in, keepaspectratio]{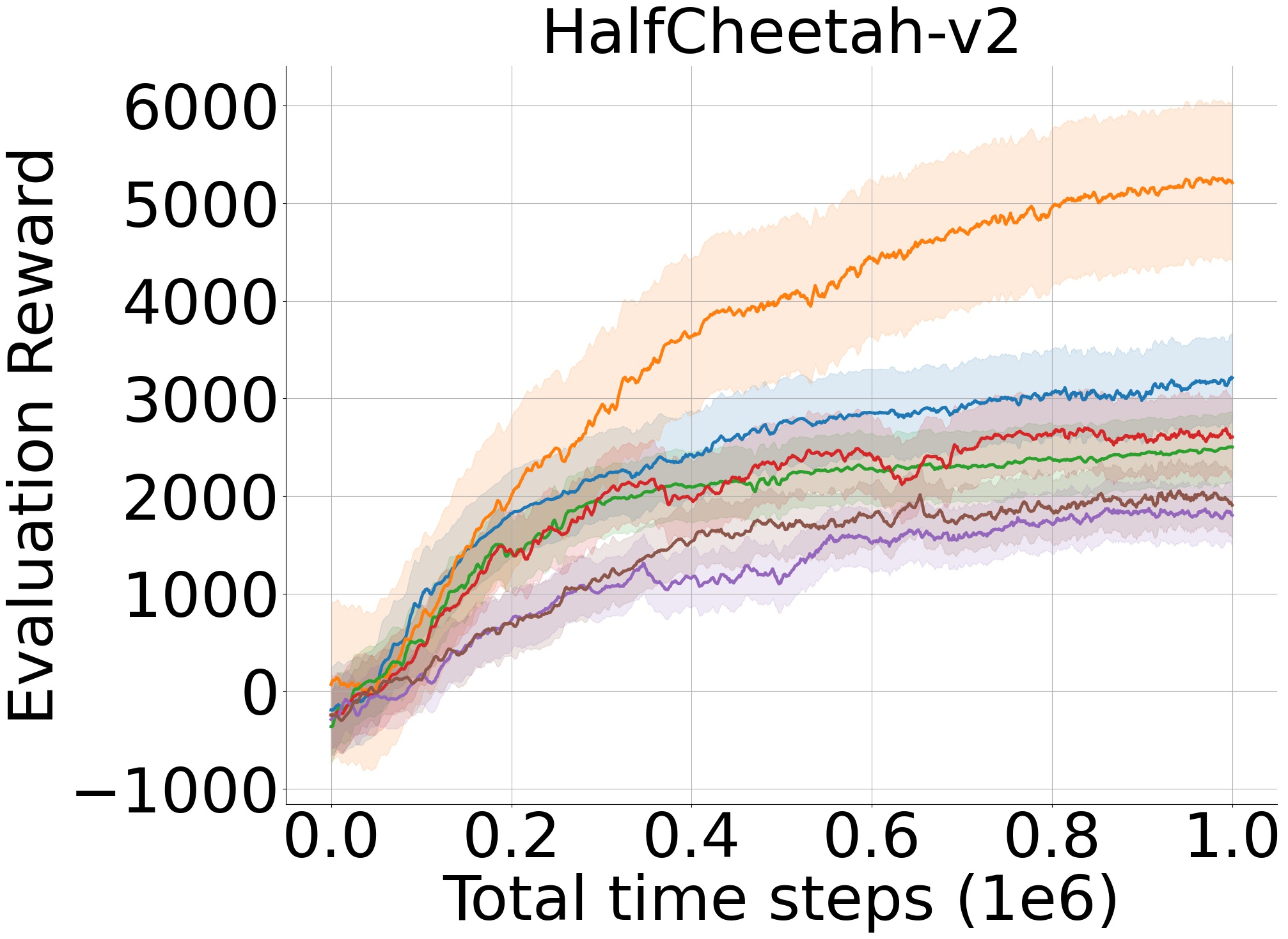}
		\includegraphics[width=1.5in, keepaspectratio]{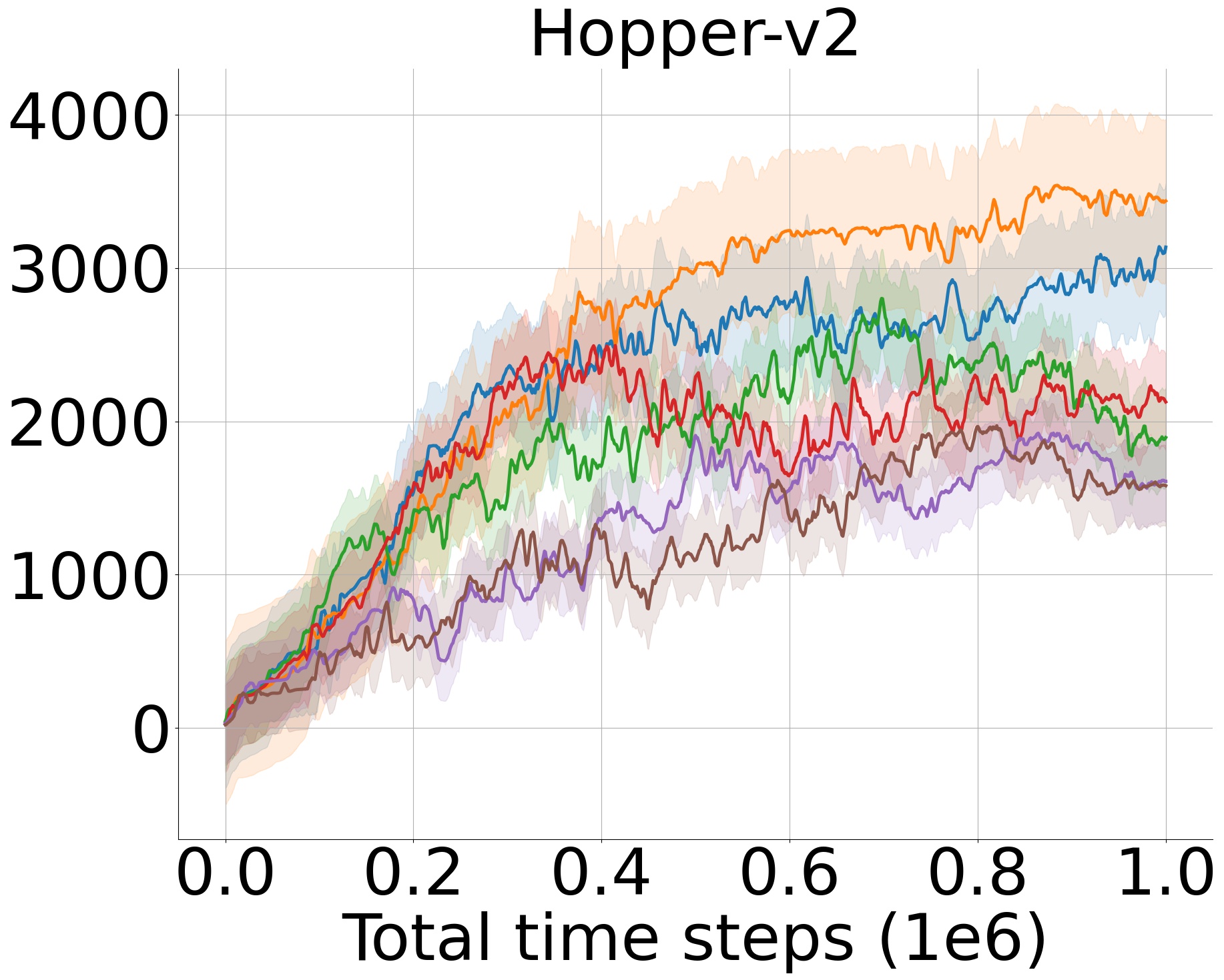}
		\includegraphics[width=1.5in, keepaspectratio]{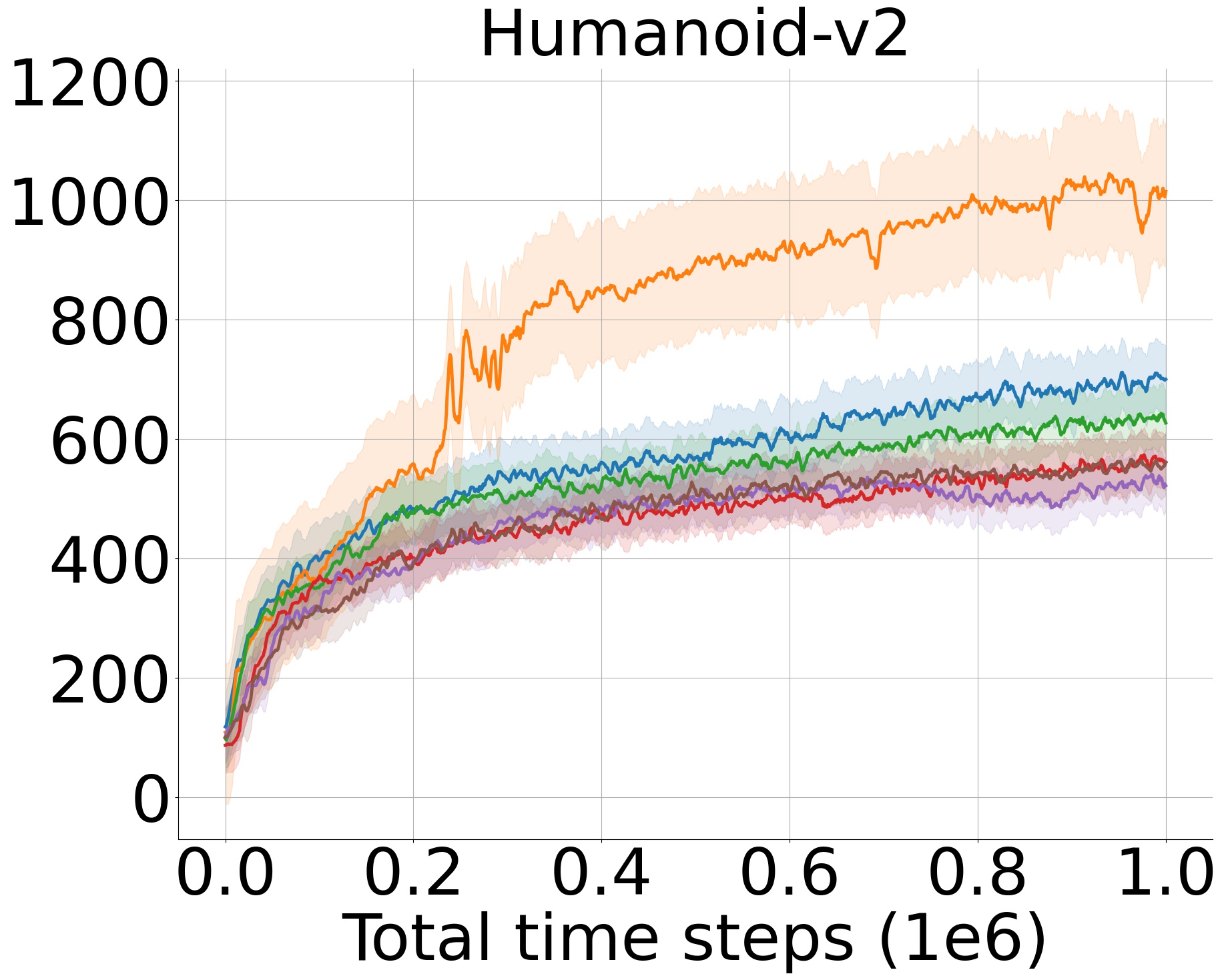}
		\includegraphics[width=1.5in, keepaspectratio]{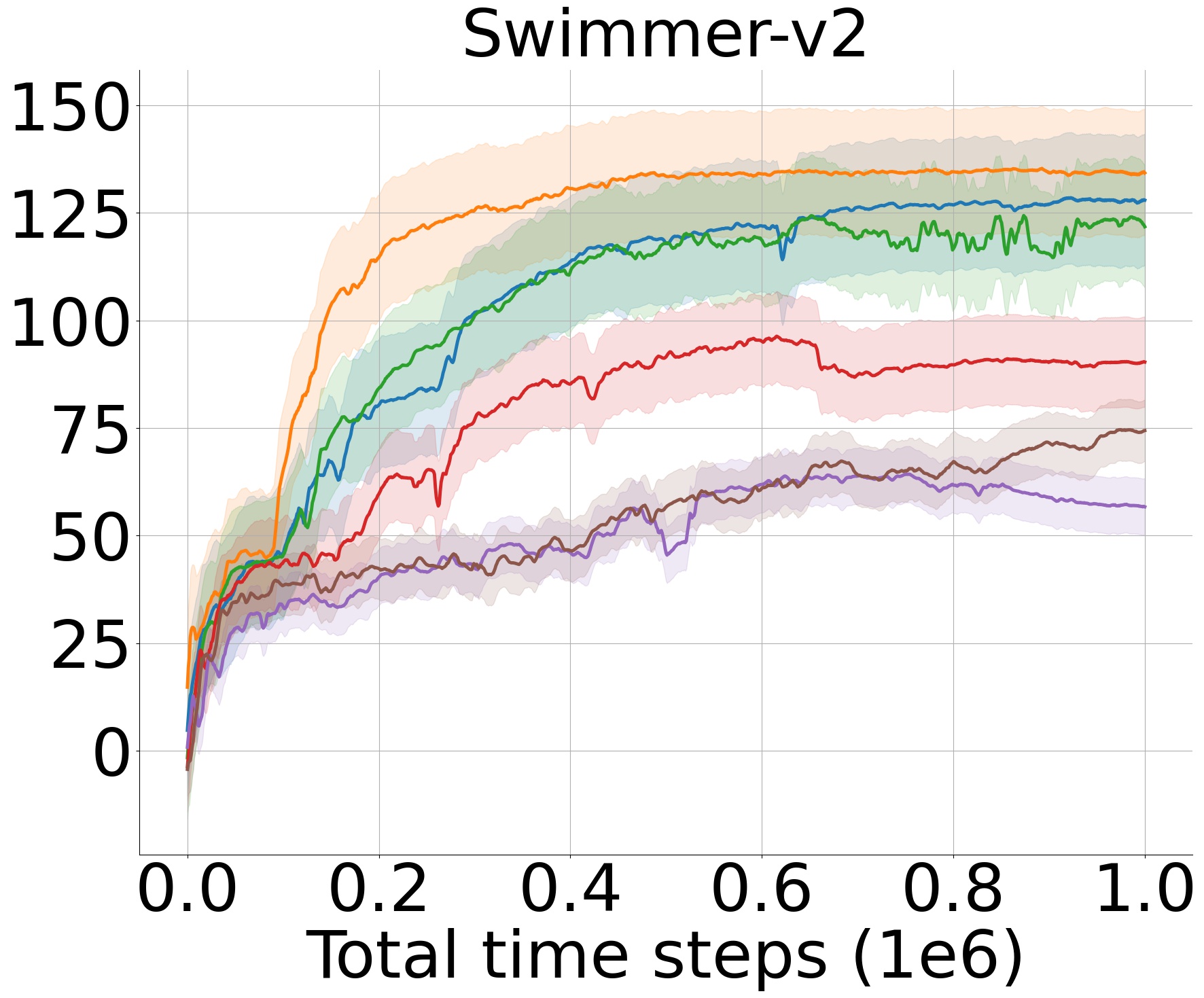}
	} 
	\caption{Evaluation curves for the set of MuJoCo and Box2D continuous control tasks under the DISCOVER-PPO algorithm when $\lambda = \{0.0, 0.1, 0.3, 0.6, 0.9, 1.0\}$. The shaded region represents half a standard deviation of the average evaluation return over 10 random seeds. A sliding window of size 5 smoothes curves for visual clarity.}
	\label{fig:exploration_tuning_results_on}
\end{figure*}

\begin{figure*}[!hbt]
    \centering
    \begin{align*}
        &\text{{\blue} $\lambda = 0.0$} &&\text{{\orange} $\lambda = 0.1$} &&\text{{\green} $\lambda = 0.3$} \\
        &\text{{\red} $\lambda = 0.6$} &&\text{{\purple} $\lambda = 0.9$} && \text{{\brown} $\lambda = 1.0$}
    \end{align*}
	\subfigure{
		\includegraphics[width=1.5in, keepaspectratio]{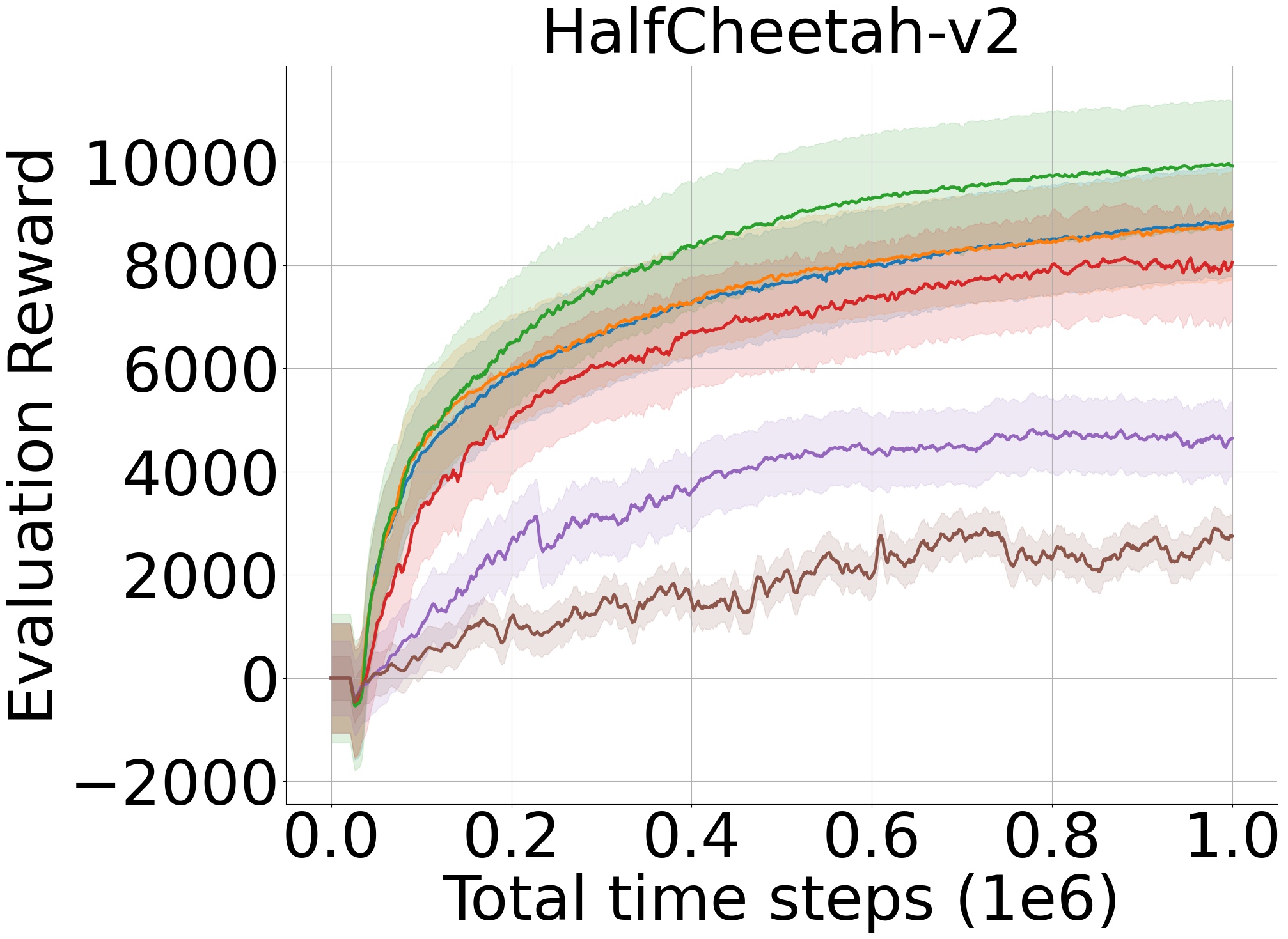}
		\includegraphics[width=1.5in, keepaspectratio]{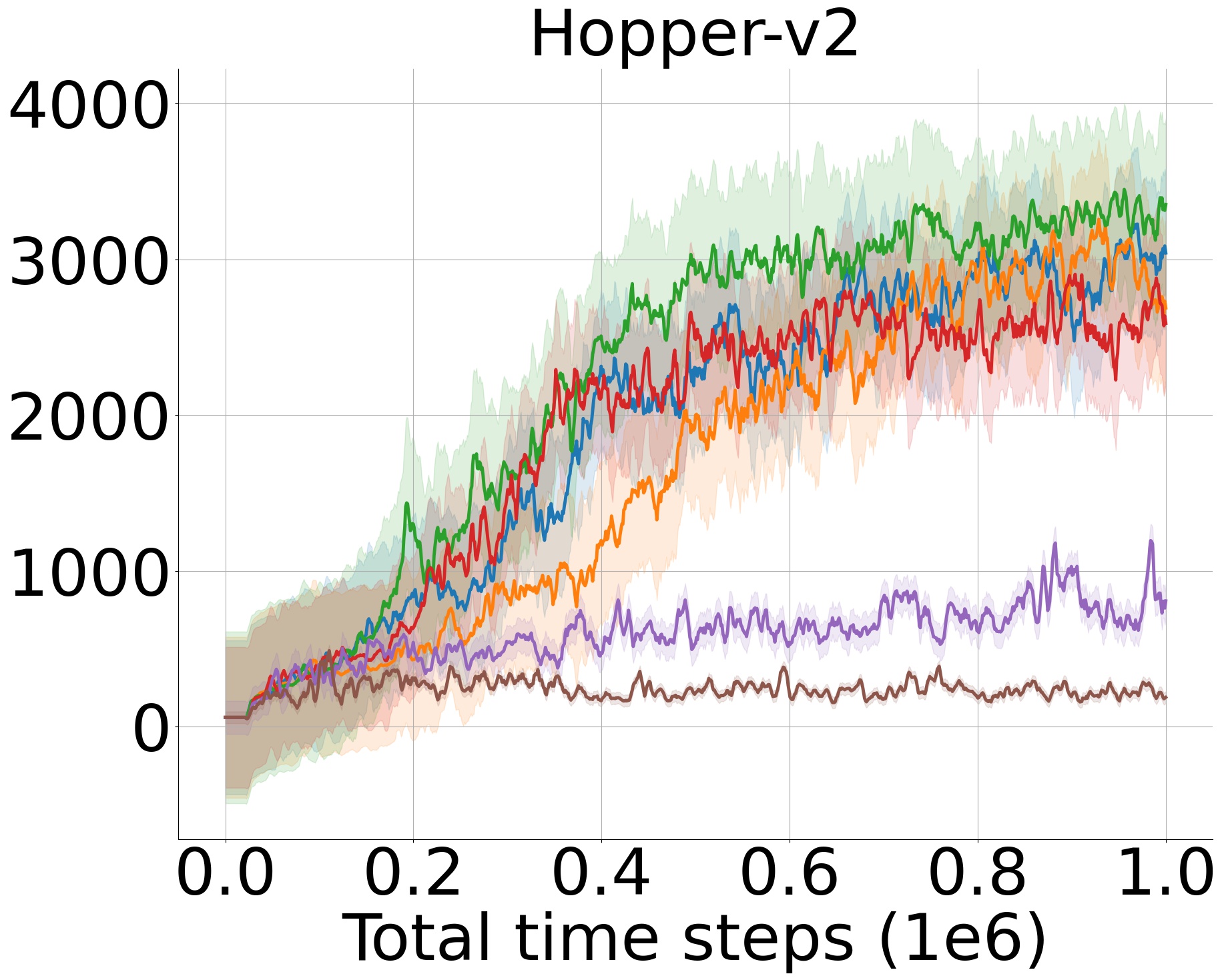}
		\includegraphics[width=1.5in, keepaspectratio]{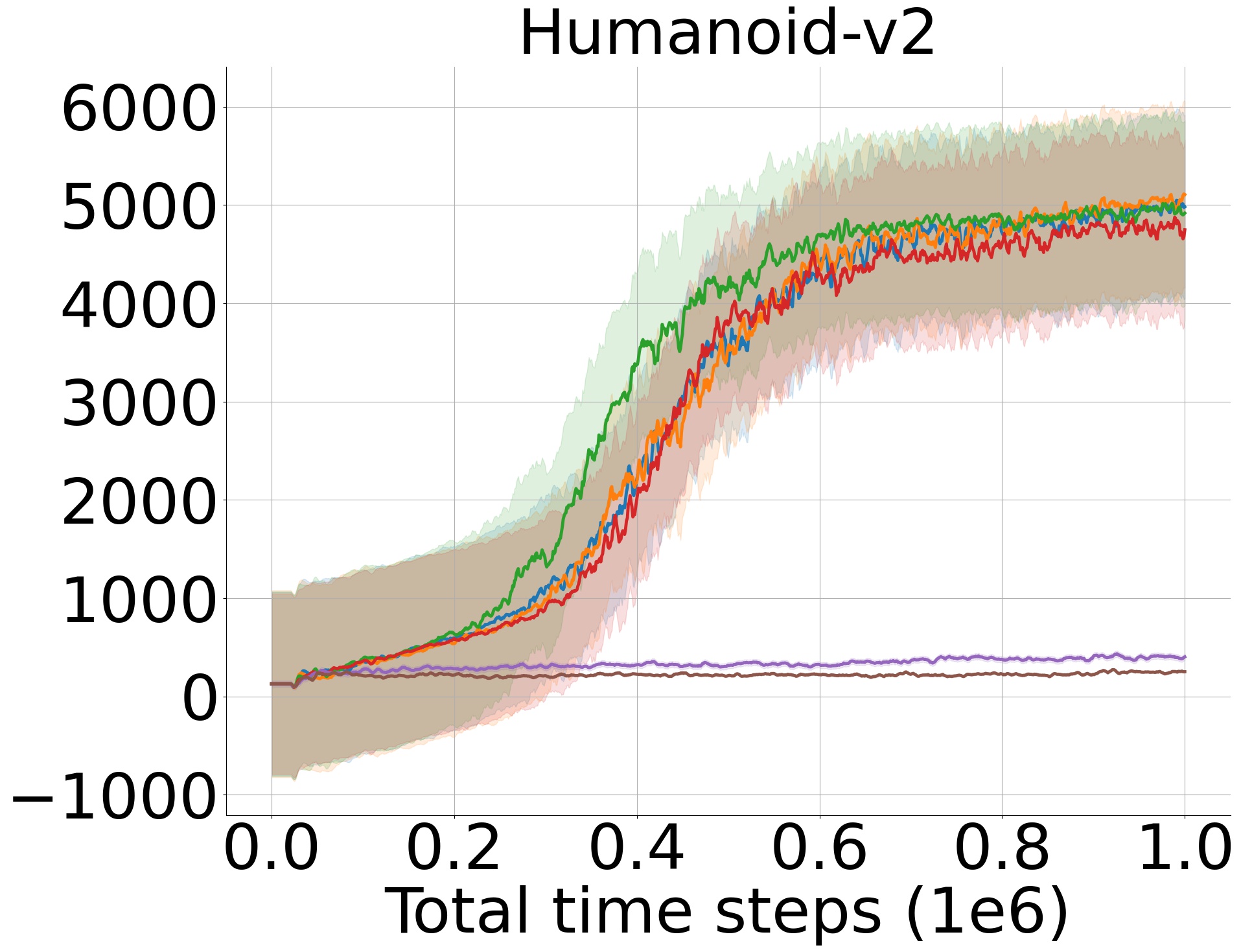}
		\includegraphics[width=1.5in, keepaspectratio]{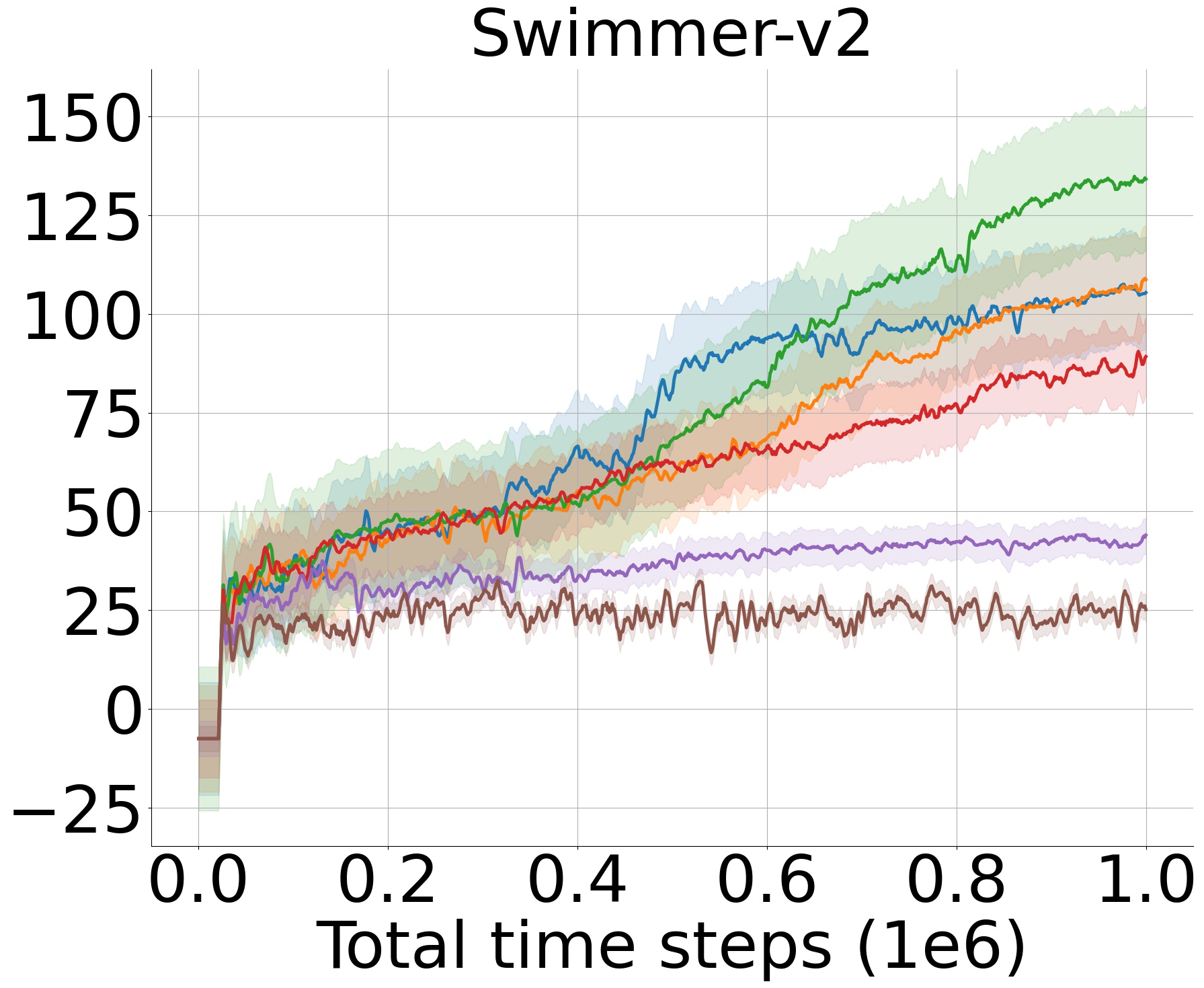}
	} 
	\caption{Evaluation curves for the set of MuJoCo and Box2D continuous control tasks under the DISCOVER-TD3 algorithm when $\lambda = \{0.0, 0.1, 0.3, 0.6, 0.9, 1.0\}$. The shaded region represents half a standard deviation of the average evaluation return over 10 random seeds. A sliding window of size 5 smoothes curves for visual clarity.}
	\label{fig:exploration_tuning_results_off}
\end{figure*}

From Figure and \ref{fig:exploration_tuning_results_on} and \ref{fig:exploration_tuning_results_off}, we observe that the increasing degree of exploration detrimentally degrades the performance of the baseline algorithms. This is an expected result as highly perturbed actions and state distributions prevent agents from effectively learning from its mistakes. Inversely, insufficient exploration obtains a notable but suboptimal performance. In fact, greedy action selection, i.e., $\lambda = 0.0$, peforms second best after DISCOVER yet, converges to a suboptimal policy. 

The optimal exploration regularization value is found to be 0.1 and 0.3 for on- and off-policy settings. These values either enables a faster convergence to the optimal policy, e.g., Humanoid, or higher evaluation results, e.g., Swimmer, or both, e.g., HalfCheetah, Hopper. In addition, from Table \ref{table:ablation}, we find that the last 10 returns for $\lambda = 0.1$ is higher than $\lambda = 0.3$ in the Hopper-v2 environment under the off-policy setting. However, this is not visible in Figure \ref{fig:exploration_tuning_results_off} due to the sliding window on the evaluation results. As $\lambda = 0.3$ setting has a faster convergence than $\lambda = 0.1$, we observe a better overall performance in the plots. 

As explained, we perform our ablation studies in environments with different characteristics. Interestingly, we find that $\lambda = 0.3$ can generalize all the tested tasks. Hence, we infer that one can tune DISCOVER on a single physics dynamics and use the optimal value in different environments. 

% Exploration Direction Regularization Figures - PPO

% Exploration Direction Regularization Figures - TD3

\clearpage

\subsubsection{Ablation Study of DISCOVER}
\begin{figure*}[!hbt]
    \centering
    \begin{align*}
        &\text{{\blue} DISCOVER} &&\text{{\orange} w/o DPU} &&\text{{\purple} w/o TN} &&\text{{\green} w/o TSR}
    \end{align*}
	\subfigure{
		\includegraphics[width=1.5in, keepaspectratio]{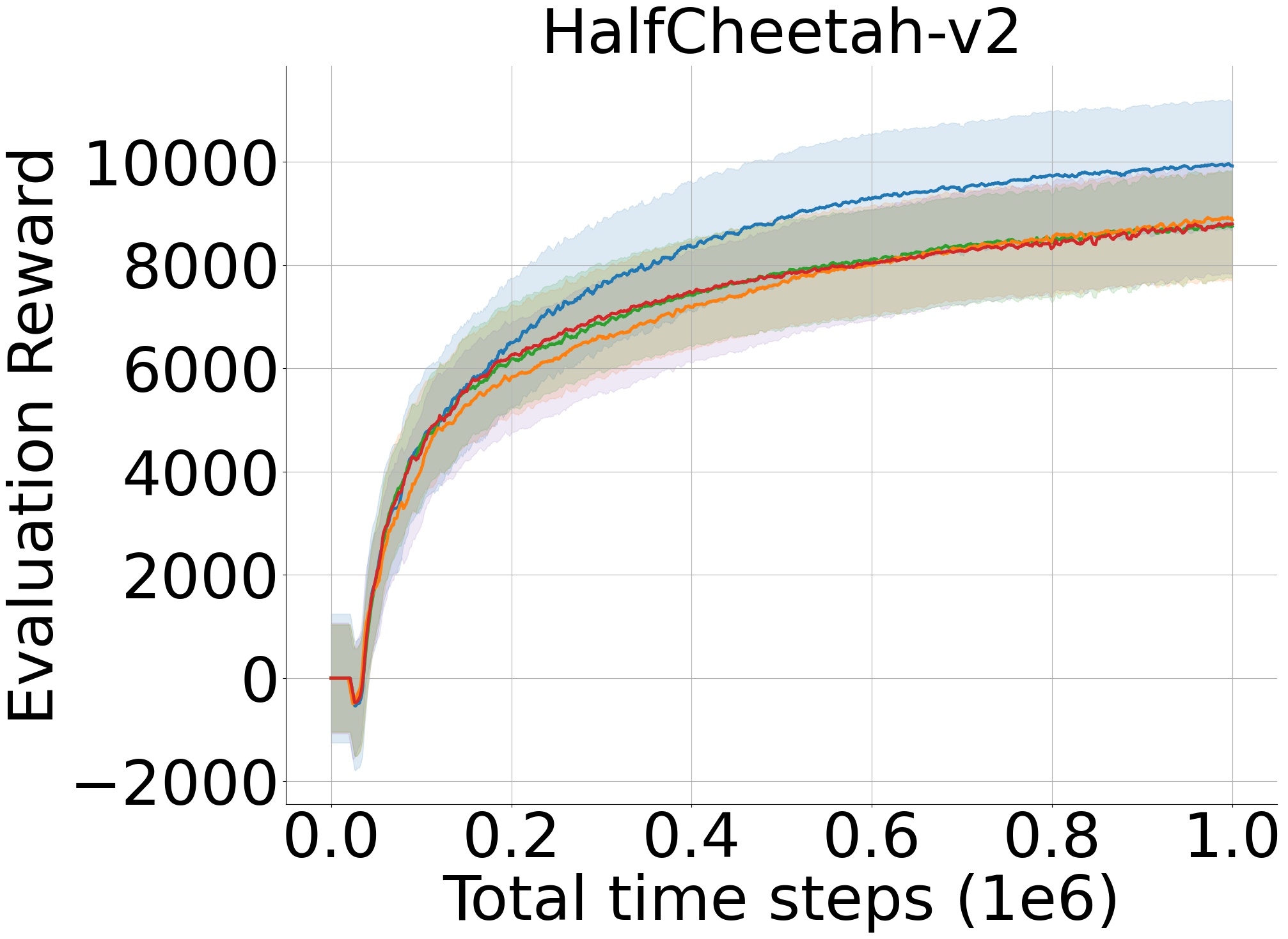}
		\includegraphics[width=1.5in, keepaspectratio]{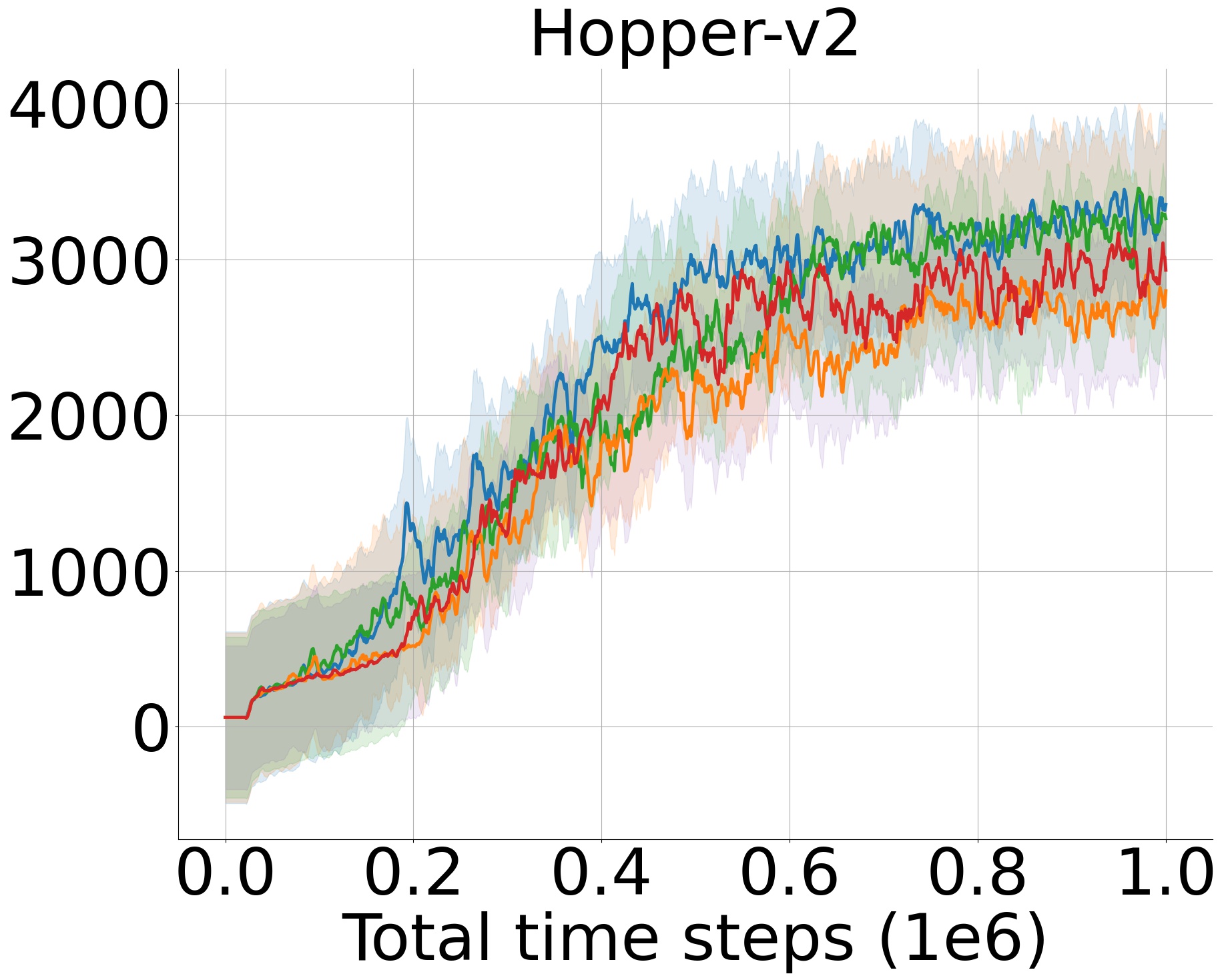}
		\includegraphics[width=1.5in, keepaspectratio]{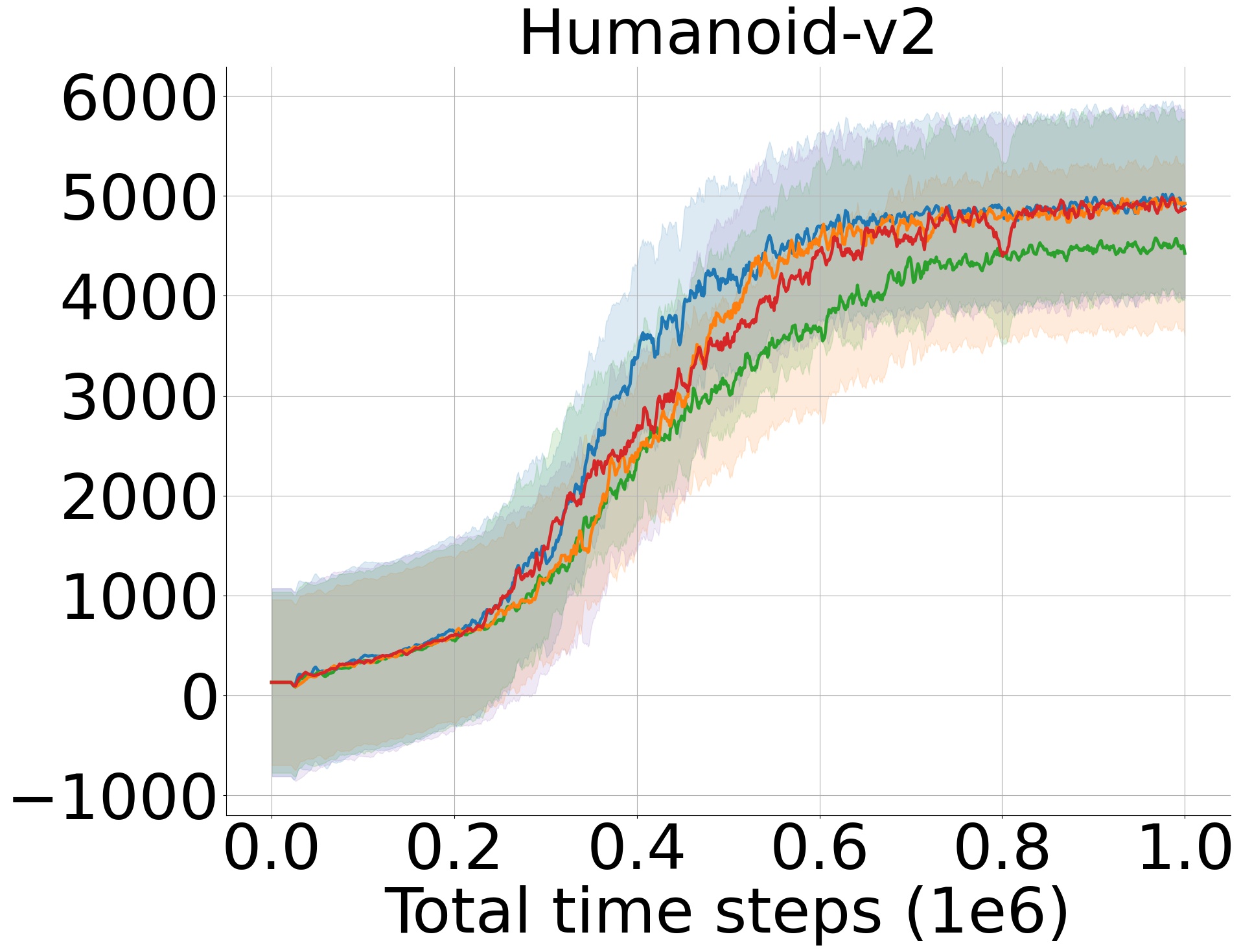}
		\includegraphics[width=1.5in, keepaspectratio]{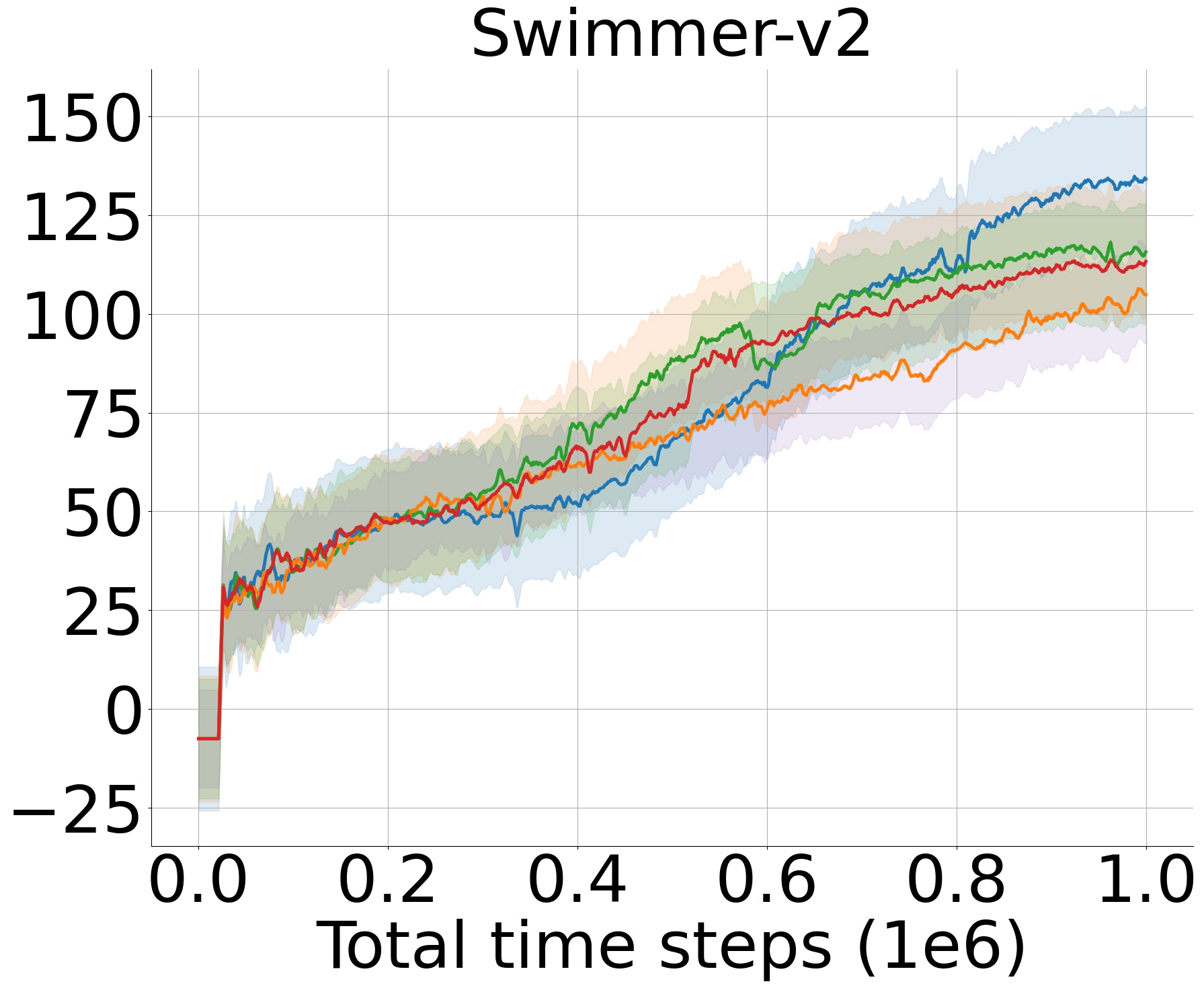}
	} 
	\caption{Evaluation curves for the set of MuJoCo continuous control tasks under the DISCOVER-TD3 algorithm when each of the DISCOVER components is removed. The shaded region represents half a standard deviation of the average evaluation return over 10 random seeds. A sliding window of size 5 smoothes curves for visual clarity.}
	\label{fig:ablation_studies}
\end{figure*}

Overall, Figure \ref{fig:ablation_studies} demonstrates that all settings exhibit a similar performance except for HalfCheetah as the complete algorithm converges faster to significantly higher evaluation results. This is because the HalfCheetah environment vastly requires off-policy samples to be solved, as highlighted in \citep{deep_rl_that_matters}. Therefore, the complete algorithm can further benefit from off-policy learning, demonstrated in the HalfCheetah environment.

In the rest of the environments, the complete algorithm attains slightly faster convergence to higher returns, from which we conclude that the ensemble of each component is the most effective setting in improving the baseline's policy. Thus, the exploration policy of DISCOVER should mimic the baseline's policy framework to obtain the optimal performance.  

% Ablation Studies Figures

\end{appendices}

\end{document}